\documentclass[runningheads]{llncs}
\usepackage{eccv}
\usepackage{eccvabbrv}
\usepackage{orcidlink}

\usepackage{wrapfig}
\usepackage{graphicx}
\usepackage{booktabs}
\usepackage{multirow}
\usepackage{amsmath}
\usepackage{amsfonts}
\usepackage{hyperref}
\usepackage{algorithmic}
\usepackage{algorithm}
\usepackage[accsupp]{axessibility}
\usepackage[font=small,labelfont=bf]{caption}
\usepackage[toc,page,header]{appendix}
\usepackage{minitoc}
\renewcommand \thepart{}
\renewcommand \partname{}

\begin{document}
\doparttoc %
\faketableofcontents %

\title{View Selection for 3D Captioning via Diffusion Ranking} 

\author{Tiange Luo\inst{1} \and Justin Johnson\inst{1}$^{,\dagger}$ \and Honglak Lee\inst{1,2}$^{,\dagger}$}

\authorrunning{Tiange Luo et al.}

\institute{University of Michigan \and
LG AI Research\\
\href{https://huggingface.co/datasets/tiange/Cap3D}{\textcolor{purple}{https://huggingface.co/datasets/tiange/Cap3D}}}

\maketitle

\begin{abstract}

Scalable annotation approaches are crucial for constructing extensive 3D-text datasets, facilitating a broader range of applications. However, existing methods sometimes lead to the generation of hallucinated captions, compromising caption quality. This paper explores the issue of hallucination in 3D object captioning, with a focus on Cap3D~\cite{luo2023scalable} method, which renders 3D objects into 2D views for captioning using pre-trained models. We pinpoint a major challenge: certain rendered views of 3D objects are accidental~\cite{freeman1994generic}, deviating from the training data of standard image captioning models and causing hallucinations. To tackle this, we present DiffuRank, a method that leverages a pre-trained text-to-3D model to assess the alignment between 3D objects and their 2D rendered views, where the view with high alignment closely represent the object's characteristics. By ranking all rendered views and feeding the top-ranked ones into GPT4-Vision, we enhance the accuracy and detail of captions, enabling the correction of 200k captions in the Cap3D dataset and extending it to 1.5 million captions across the entire Objaverse dataset and a portion of the Objaverse-XL high-quality subset. Additionally, our dataset includes point clouds with 16,384 colored points and 20 rendered images per caption, along with intrinsic and extrinsic camera parameters, depth data, and masks. Beyond datasets, we showcase the adaptability of DiffuRank by applying it to pre-trained text-to-image models for visual question answering tasks, where it outperforms the CLIP model.
\end{abstract}

{\let\thefootnote\relax\footnotetext{$^\dagger$ equal advising}}

\section{Introduction}
\label{sec:intro}

Recent advancements in generative models have shown remarkable performance in both image~\cite{saharia2022photorealistic, betker2023improving} and video~\cite{videoworldsimulators2024} domains, driven by the availability of extensive captioned datasets. Despite these successes, extending generative modeling to 3D domains has been challenging due to the scarcity of high-quality 3D-text pairs. This gap has been partially bridged by Cap3D~\cite{luo2023scalable}, which generates captions for 3D objects by rendering them into 2D images and employing image-based captioning models, further refined by Large Language Models (LLMs) to synthesize captions. Cap3D has contributed 660k captions for the Objaverse dataset~\cite{deitke2022objaverse}, facilitating developments in Text-to-3D~\cite{yariv2023mosaic, li2023instant3d}, Image-to-3D~\cite{xu2023dmv3d, zhao2023efficientdreamer}, robot simulator~\cite{wang2023robogen} and learning~\cite{qi2024shapellm}, and the pre-training of 3D LLMs~\cite{xu2023pointllm, zhou2023regionblip, panagopoulou2023x}.

\begin{figure}[t]
    \centering
    \includegraphics[width=1\textwidth]{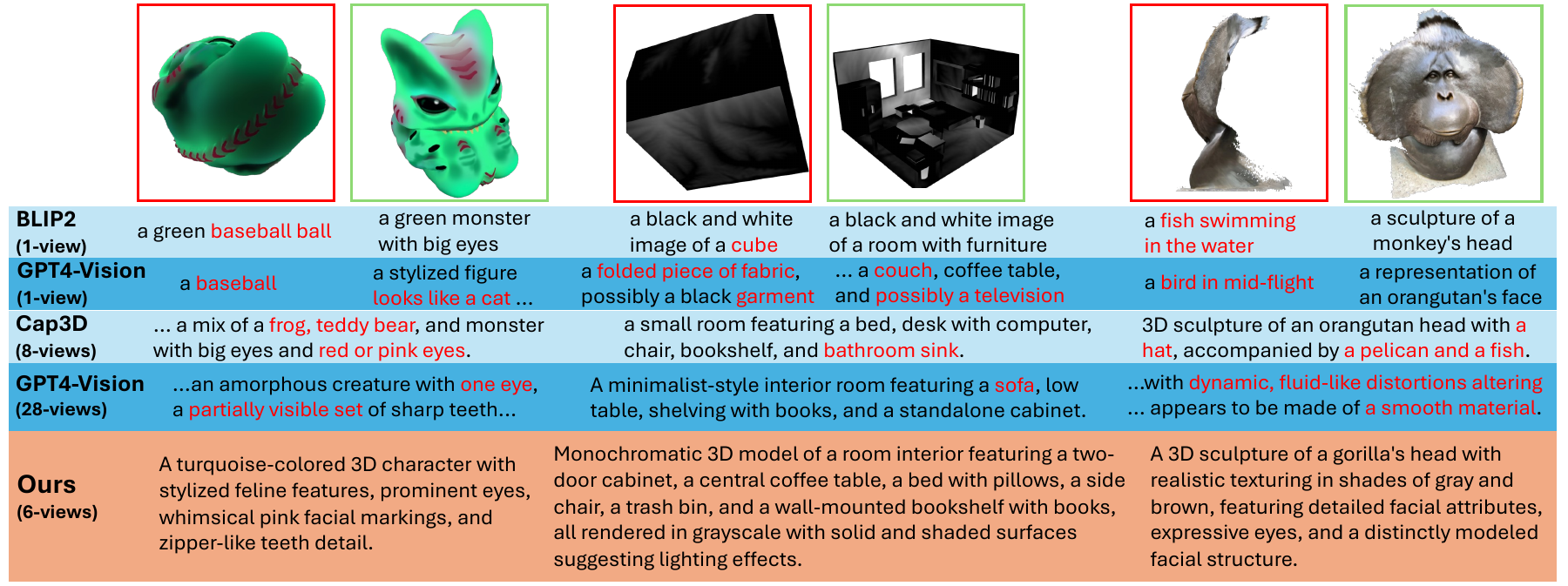}
    \caption{DiffuRank enhances caption accuracy and reduces hallucinations by prioritizing key rendered views (green box), contrasting the accidental views~\cite{freeman1994generic} (red box) that cause errors. Surprisingly, using fewer views (6 vs. 28) not only saves computational resources but also may yield more accurate and detailed outcomes (the middle example) by countering the uncertainty caused by excessive views.}
    \label{fig:teaser}
\end{figure}

Despite the utility of Cap3D, our analysis reveals that a significant portion of Cap3D captions contain inaccurate or hallucinated details, potentially compromising model training~\cite{tang2023volumediffusion}. Upon inspection, we identified one of the root causes as \emph{accidental views} -- special camera positions where coincidental feature alignments make the projection uninformative or misleading for recognition. Examples include the first ``baseball" image, which actually depicts a green creature, and the third ``cube" image, which is in fact an indoor design, both shown in Figure~\ref{fig:teaser}. Because Cap3D inherits Objaverse’s default object orientation, its cameras sit on a fixed horizontal ring, a heuristic that inevitably samples such accidental views. These accidental renderings are hard even for humans, and current captioners~\cite{li2023blip} struggle to describe them faithfully. Consequently, when these challenging views are included, even advanced captioning models like GPT4-Vision~\cite{achiam2023gpt} may generate erroneous information, as illustrated in Figure~\ref{fig:teaser}.

\begin{figure}[t]
    \centering
    \includegraphics[width=\textwidth]{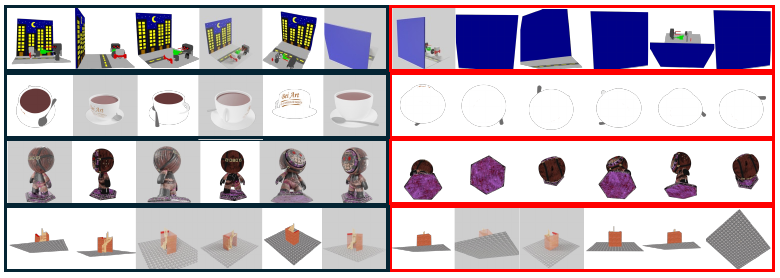}
    \caption{The left row features the top-6 views as ranked by DiffuRank, while the right row displays the bottom-6. Comparative analysis shows that the top-6 views generally uncover more characteristics of the object compared to the bottom-6. This finding underscores DiffuRank's capability to identify views that more accurately represent the features of the 3D object. More randomly sampled results are included in Appendix~B.5.}
    \label{fig:diffurank_effects}
\end{figure}

To address this, we introduce DiffuRank, an approach for ranking rendered views with pre-trained diffusion models. By leveraging a pre-trained text-to-3D diffusion model~\cite{jun2023shap}, DiffuRank evaluates the alignment between the captions of each view and the corresponding 3D object's information. The underlying premise is that captions generated from rendered views that closely match the object's 3D information will exhibit a higher alignment, indicating these views are more representative of the object. Consequently, DiffuRank promotes the preferable views (Figure~\ref{fig:diffurank_effects}) for captioning as those that better reflect the true essence of the 3D objects, leading to more accurate and truthful captions. 

Specifically, we first employ image-based captioning models to caption all candidate rendered views. We then utilize multiple iterations of a text-to-3D diffusion model objective to estimate an average alignment score, using each caption as a conditioning input for the same underlying 3D object feature, Gaussian noise, and timestamps. The intuition here is that captions better aligned with the 3D object feature provide more effective guidance during the diffusion process, resulting in higher alignment scores. This score effectively quantifies how well each caption aligns with the corresponding 3D object feature. Since the captions are generated from rendered views via captioning models trained through maximum likelihood estimation, the resulting alignment scores inherently reflect the alignment between the rendered views and the original 3D object. Higher scores indicate stronger alignment, implying fewer possibilities of being accidental views.

Following this scoring procedure, we rank the views based on their alignment scores and select the top-N rendered views for further caption generation using GPT4-Vision. Human evaluations demonstrate that captions produced using DiffuRank combined with GPT4-Vision exhibit significantly higher quality and fewer inaccuracies compared to captions generated by Cap3D. Additionally, our method consistently yields richer, more detailed captions and fewer hallucinations when employing just 6 strategically selected rendered views, as opposed to using GPT4-Vision directly on all 28 views or those selected by default object orientations.

Further, we extend DiffuRank to the 2D domain, demonstrating its effectiveness in the challenging Visual Question Answering task~\cite{tong2024eyes} when combined with text-to-2D diffusion models~\cite{rombach2022high}, and surpassing the zero-shot performance of CLIP~\cite{radford2021learning}.

Our contributions are as follows:
\begin{itemize}
    \item We propose DiffuRank to address the issue of accidental views in 3D object renderings. In principle, DiffuRank estimates the alignment score between different modalities. Therefore, we extend DiffuRank to the 2D domain and demonstrate that, with the assistance of a pre-trained text-to-2D diffusion model~\cite{rombach2022high}, DiffuRank outperforms CLIP on the VQA task~\cite{tong2024eyes}.
    \item We identify and alleviate the systematic hallucinations in Cap3D captions, revising approximately 200k entries with the help of DiffuRank and GPT4-Vision. The corrected captions consistently improve the finetuned performance of text-to-3D models (Point·E, Shap·E); note that Shap·E models fine-tuned with Cap3D captions show decreased performance.
    \item We extend the Cap3D caption dataset~\cite{luo2023scalable} from 660k to 1.5M across the entire Objaverse~\cite{deitke2022objaverse} and a portion of the Objavere-XL high-quality subset~\cite{deitke2023objaverse}. Each caption is complemented with point clouds containing 16,384 colorful points and 20 rendered images, including camera, depth, and MatAlpha details. All data is released under the ODC-By 1.0 license and is available at \href{https://huggingface.co/datasets/tiange/Cap3D}{\textcolor{purple}{https://huggingface.co/datasets/tiange/Cap3D}}.
    
\end{itemize}

\section{Related Work}
\label{sec:related_work}
\subsection{3D-Text}
Recent advancements introduced by Objaverse have significantly enriched the field of 3D object research. By integrating a comprehensive set of 3D objects with descriptive captions from Cap3D, a wide array of 3D applications has been enabled. These include Text-to-3D methods~\cite{yariv2023mosaic, li2023instant3d, he2023t, li2023sweetdreamer, mercier2024hexagen3d}, Image-to-3D conversion techniques~\cite{xu2023dmv3d, zhao2023efficientdreamer}, enhancements in robot learning~\cite{wang2023robogen, qi2024shapellm}, the pre-training of 3D language models~\cite{xu2023pointllm, zhou2023regionblip, qi2023gpt4point,liu2024uni3d,chen2023ll3da}, and the development of language models capable of processing diverse modalities~\cite{han2023onellm,panagopoulou2023x,chen2024model}.

Despite these advancements, we identified issues with hallucination contents in the captions provided by Cap3D. This discovery aligns with findings from concurrent research~\cite{tang2023volumediffusion, liu2023pi3d, kabra2023evaluating}, pinpointing inaccuracies in Cap3D captions. Our investigation reveals that the root cause of these inaccuracies is attributed to accidental rendered views, which lead to failures in captioning models. These failures are exacerbated as text summarization models (GPT4) are unable to rectify these errors. To address this challenge, we introduce DiffuRank that selects rendered views capturing the essential characteristics of 3D objects. Furthermore, we utilize the recent advancements in vision-language models, specifically GPT4-Vision, to provide holistic captions for 3D objects. We release our dataset under ODC-By 1.0 license to enable research and commercial usage, and hope facilitate related 3D-Text research~\cite{jain2021zero, poole2022dreamfusion, lin2022magic3d, sanghi2022clip, zhu2023hifa, wang2023score, chen2023fantasia3d, lorraine2023att3d, li2023instant3d, yi2023gaussiandreamer, li2023sweetdreamer, luo2023neural, ding2023text, chen2023control3d, michel2022text2mesh, wei2023taps3d, chen2023text2tex, nichol2022point, liu2023zero, liu2023one, melas2023realfusion, tang2023make, shi2023mvdream, xu2023dmv3d, chen2023cascade, shi2023mvdream}. 

\subsection{Diffusion Model}
Our proposed DiffuRank leverages denoising diffusion objective \cite{sohl2015deep, song2019generative, ho2020denoising} to model the alignment between the input and output modalities. By using pre-trained text-to-3D~\cite{nichol2022point, jun2023shap} and text-to-2D ~\cite{saharia2022photorealistic, betker2023improving, peebles2023scalable} diffusion models, we can model the alignment between given 3D object/image for a set of possible captions (text descriptions) as detailed in Section~\ref{sec:method:diffurank}. In our listed algorithm~\ref{alg:diffurank}, we adopt the objects $L_{3D}=E_{x_0 \sim q\left(x_0\right), \epsilon \sim \mathcal{N}(0, \mathbf{I}), t \sim U[1, T]}\left\|x_\theta\left(x_t, t\right)-x_0\right\|_2^2$ as used in Shap·E~\cite{jun2023shap}, where $x_0$ is data sampled from data distribution $q(x_0)$, $\epsilon$ is Gaussian noise, and t is timestamp. We also adopt the alternative but equivalent objective, $L_{\text {2D}}=E_{x_0 \sim q\left(x_0\right), \epsilon \sim \mathcal{N}(0, \mathbf{I}), t \sim U[1, T]}\left\|\epsilon-\epsilon_\theta\left(x_t, t\right)\right\|_2^2$, when we adopt the text-to-2D model, stable-diffusion, in Section~\ref{sec:exp:diffurank}. 

DiffuRank is related to score sampling distillation proposed in \cite{poole2022dreamfusion}, while we do not compute gradients but sampling loss to accumulate scores estimation for ranking. Our findings also relate to works which leverage pre-trained diffusion models for downstream tasks, such as image classification~\cite{mukhopadhyay2023diffusion,li2023your}, semantic segmentation~\cite{zhao2023unleashing}, visual grounding~\cite{liu2023vgdiffzero}, depth prediction~\cite{saxena2023monocular,zhang2024tale}, and other low-level computer vision tasks~\cite{du2023generative}.

When applying our method to the 2D domain, we discovered that our algorithm aligns closely with the insights of the approach presented in \cite{li2023your}. Consequently, our method can be considered an expansion of the findings from \cite{li2023your}, extending its applicability from 2D classification to broader domains and tasks,
including the use of a pre-trained text-to-3D diffusion model and a 2D-image-based captioning model to estimate the alignment between 3D objects and their 2D rendered views, as well as the application of a pre-trained text-to-2D diffusion model to solve Visual Question Answering tasks.

\begin{figure}[t]
    \centering
    \includegraphics[width=\textwidth]{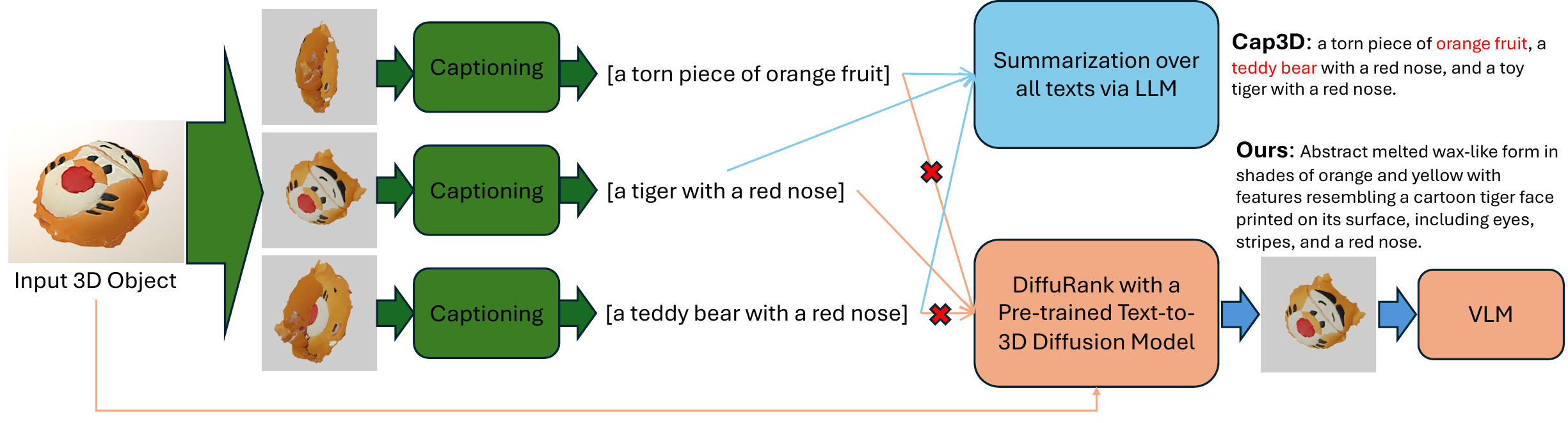}
    \caption{Methods overview. Both Cap3D and our method render input 3D objects into multiple views for caption generation (green steps). However, while Cap3D consolidates these captions into a final description (blue steps), our method employs a pre-trained text-to-3D diffusion model to identify views that better match the input object's characteristics. These selected views are then processed by a Vision-Language Model (VLM) for captioning (orange steps).}
    \label{fig:method}
\end{figure}

\section{Method}
In this section, we analyze the issues with accidental rendered views leading to hallucinations in Cap3D captions, motivating our proposed DiffuRank, a approach for selecting informative rendered views with 3D priors learned from a diffusion model. We then detail DiffuRank's formulation and describe our novel 3D captioning framework that integrates GPT4-Vision.

\subsection{Issues in Cap3D}

Firstly, we revisit the Cap3D pipeline, which unfolds across four stages. Initially, it renders a set of 2D views for each 3D object. Subsequently, image captioning is applied to generate preliminary descriptions (5 captions for each image). Then, the CLIP model is utilized in the third stage to select the best-aligned caption for each image, filtering out inaccuracies. The process culminates with an LLM synthesizing captions from various perspectives into a comprehensive caption.

However, the captioning of rendered views (the combined second and third stages) for given 3D objects can falter with accidental views, producing captions that diverge significantly from the actual 3D object. In the worst-case scenarios, each rendering view might correspond to an incorrect object, leading to compounded errors when these captions are summarized by GPT4. One example is shown in Figure~\ref{fig:method}. Since GPT4 operates solely on text, it cannot correct these inaccuracies, resulting in captions riddled with hallucinated details.

Due to the versatility of 3D object geometries, determining which rendered views best reflect a 3D object's characteristics is non-trivial. While measuring the geometric properties of 3D objects and computing their principal directions is feasible, positioning the camera orthogonally, as shown in the bottom-left example of Figure~\ref{fig:diffurank_effects}, is often suboptimal. Hence, we propose DiffuRank, which learns 3D priors from data to filter preferable rendered views by leveraging a pre-trained text-to-3D model. Our experiments demonstrate that DiffuRank efficiently enhances caption quality and reduces hallucinations with fewer renderings compared to using all available views.

\subsection{DiffuRank Formulation}
\label{sec:method:diffurank}

DiffuRank leverages a pre-trained text-to-3D diffusion model $D_{\text{text-to-3D}}$ to rank rendered views based on their alignment with captions and the corresponding 3D information.

Formally, given a 3D object $\mathcal{O}$, a set of candidate captions $\{c_i\}$, and a pre-trained diffusion model $D_{\text{text-to-3D}}$, the training objective of this diffusion model is to predict the 3D object $\mathcal{O}$ conditioned on a text description $c$. This corresponds to modeling the score function $\nabla_{\mathcal{O}, c} \log p(\mathcal{O}|c)$ of the data distribution $p(\mathcal{O}|c)$. Specifically, the diffusion model aims to minimize the objective:
\[
\mathcal{L}_{c} = \|D_{\text{text-to-3D}}(\mathcal{O}_t | c) - \mathcal{O}_0\|,
\]
where the noised input is defined as $\mathcal{O}_t = \sqrt{\bar{\alpha}_t}\,\mathcal{O}_0 + \sqrt{1-\bar{\alpha}_t}\,\epsilon$, for timestamp $t$, randomly sampled Gaussian noise $\epsilon \sim \mathcal{N}(0, I)$, and noise scheduling parameters $\bar{\alpha}_t$ defined as in~\cite{ho2020denoising}. Our intuition here is straightforward: a caption closely aligned with the 3D object's characteristics (e.g., structure, colors, textures) should assist the diffusion model in accurately predicting $\mathcal{O}_0$ from the same noised input $\mathcal{O}_t$, thus resulting in a lower score matching loss. By sampling multiple sets of $\{t_j,\epsilon_j\}$ for each caption $c_i$, we measure the alignment score $\text{Cor}(\mathcal{O}, c_i)$ between the 3D object and captions as the average loss over these samples.

Initially, we generate candidate captions for $\mathcal{O}$ by rendering it into multiple views $\{I_i\}$ and generating captions $\{c_i^j\}$ using an image captioning model $D_{\text{cap}}$. This captioning procedure aims to maximize the joint likelihood $p(c_i^j, I_i)$ of the generated captions $c_i^j$ and the rendered image $I_i$. Consequently, we estimate the alignment between the 3D object and all captions of a given rendering as $\text{Cor}(\mathcal{O}, \mathbb{E}_j[c_i^j])$, which is proportional to the alignment between the 3D object and the rendering itself:
\[
\text{Cor}(\mathcal{O}, \mathbb{E}_j[p(c_i^j, I_i)]) \propto \text{Cor}(\mathcal{O}, I_i).
\]
We summarize the complete pipeline in Algorithm~\ref{alg:diffurank}.

Specifically, we adopted shap-E as the text-to-3D diffusion model in our paper, and the above $\mathcal{O}_i$ should be $E_{encoder}(\mathcal{O}_i)$, where $E_{encoder}$ is the encoder (transmitter in \cite{jun2023shap}) to extract feature embeddings from given 3D object. 

Furthermore, DiffuRank's application is not confined to 3D captioning; because it is a general framework for measuring the alignment between two modalities received and output by a diffusion model. It can be seamlessly extended to other domains, such as 2D images. In section~\ref{sec:exp:diffurank}, we show an example where we apply DiffuRank to perform 2D VQA and beat CLIP model~\cite{radford2021learning}. 

\begin{algorithm}[t]
\caption{DiffuRank for modeling the alignments between 3D object and its rendered views}
\label{alg:diffurank}
\begin{algorithmic}
\REQUIRE Given 3D object $\mathcal{O}$, pre-trained text-to-3D model $D_{\text{text-to-3D}}$, captioning model $D_{\text{cap}}$
\STATE \# 1. rendered views $\{I_i\}_{i=1,\cdots,M}$ for $\mathcal{O}$ with rendering program (e.g., Blender).
\STATE \# 2. Generate candidate captions for $\mathcal{O}$.
\FOR{each view ${I_i}$ of $\mathcal{O}$}
    \STATE Generate captions $\{c_i^j\}_{j=1,\cdots, N}$ with captioning model $D_{\text{cap}}$.
\ENDFOR

\STATE \# 3. Compute average alignment scores
\FOR{$k \gets 1$ to $\text{num\_samples}$}
    \STATE Sample timestamp $t_k \sim \text{Uniform}(0, 1)$.
    \STATE Sample noise $\epsilon_k \sim \mathcal{N}(0, I)$.
\ENDFOR

\FOR{each rendering view $I_i$}
    \FOR{$k \gets 1$ to $\text{num\_samples}$}
        \STATE Compute noised input $\mathcal{O}_{t_k} = \sqrt{\bar{\alpha}_{t_k}} \mathcal{O}_0 + \sqrt{1-\bar{\alpha}_{t_k}} \epsilon_k$.
        \FOR{$j \gets 1$ to $N$}
            \STATE Compute loss $\mathcal{L}_{c_i^j,k} = \|D_{\text{text-to-3D}}(\mathcal{O}_{t_k} | c_i^j) - \mathcal{O}_0\|$.
        \ENDFOR
    \ENDFOR
    \STATE Compute average loss for all captions of $I_i$, $Cor({I_i},\mathcal{O}) = -\mathbb{E}_{j,k} \mathcal{L}_{c_i^j, k}$.
\ENDFOR

\RETURN Top-P($\{Cor(I_i, \mathcal{O})\}_{i=1,\cdots, M}$)
\end{algorithmic}
\end{algorithm}

\subsection{New 3D Captioning Framework}
\label{sec:method:new_framework}
With the proposed DiffuRank, we establish a new 3D captioning pipeline, as shown in Figure~\ref{fig:method}. For given 3D object, we render it into 28 images, which are then captioned into 5 descriptions using an image-based captioning model. Following captioning, DiffuRank ranks the rendered views using a pre-trained text-to-3D model. This ranking enables the selection of the Top-6 rendered views for processing by a vision-language model, resulting in holistic captions that describe structure, form, color, and texture, with enhanced accuracy and detail.

\begin{figure}[t]
    \centering
        \includegraphics[width=\textwidth]{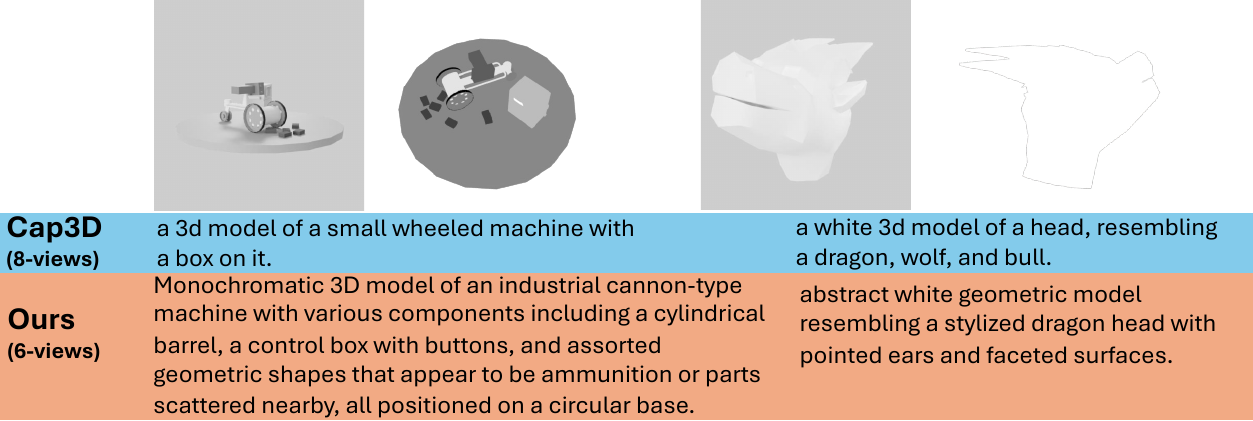}
        \caption{We utilized both grey background + ray-tracing render engine (left images) and transparent background + real-time render engine (right images) for rendering, discovering that the effectiveness of each varies. We noticed DiffuRank can select the views with the appropriate rendering that highlight object features.}
        \label{fig:qual_comp_1}
\end{figure}

To elaborate, our methodology integrates two distinct rendering strategies, as illustrated in Figure~\ref{fig:qual_comp_1}. The first strategy, derived from Cap3D~\cite{luo2023scalable}, renders objects into 8 views against a uniform grey background, arranged horizontally around the object’s default orientation, with Blender ray-tracing render engine `CYCLES'. Concurrently, we apply a second technique from Shap·E~\cite{jun2023shap}, where 20 views are generated through randomized sampling after object normalization, set against a transparent background, with Blender real-time engine `EEVEE'. These 20 views, created following the Shap·E methodology, are instrumental in forming Shap·E latent codes, i.e. $E_{encoder}(\mathcal{O}_i)$ in Section~\ref{sec:method:diffurank}. Altogether, this approach results in 28 distinct views for each object. Additionally, as grey and transparent backgrounds may accentuate or obscure details variably across objects, we observed that DiffuRank adeptly selects the views with the proper background that most effectively highlight object features, without manual intervention. Some examples are included in Appendix~B.

Following this, the captioning model, BLIP2~\cite{li2023blip}, is employed to generate five captions for each view. These captions, alongside the pre-trained text-to-3D diffusion model, Shap·E~\cite{jun2023shap}, and the previously derived 3D latent code $E_{encoder}(\mathcal{O}_i)$, undergo analysis in the DiffuRank process, as detailed in Algorithm~\ref{alg:diffurank}. Subsequent to DiffuRank, the six views that demonstrate the highest alignment scores are chosen to input into GPT4-Vision for caption generation.

\section{Dataset}
\label{sec:dataset}
This section details our process for correcting the Cap3D captions, expanding the dataset with high-quality 3D objects from Objaverse-XL, and ethical filtering. More detailed hyper-parameters and comparisons are included in Appendix~B. 

\subsection{Correction of Cap3D Captions}
As Cap3D contains a lot of good quality captions, our first objective is to identify erroneous Cap3D captions, which might contain incorrect information or hallucinations. We tried three strategies as outlines the below. 

\begin{figure}[t]
    \centering
    \includegraphics[width=1\textwidth]{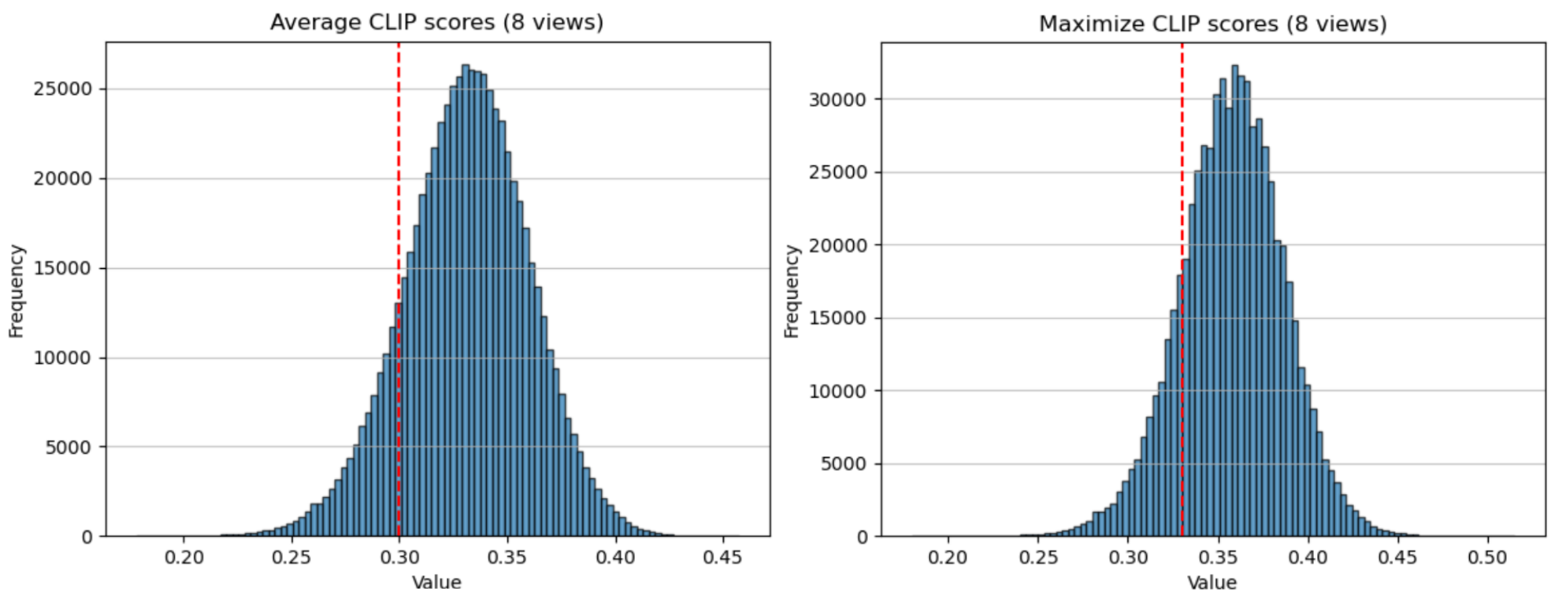}
  \caption{Mean and Max clip score distribution for Cap3D captions and their 8 rendering images. Selected thresholds are the two red dash lines via our annotated validation set.}
  \label{fig:clip}
\end{figure}

\textbf{Image-Text Alignment Method:} We discovered that utilizing the maximum and average CLIP scores effectively filters out inaccurate captions. Most of erroneous captions, like those depicted in Figure~\ref{fig:teaser}, described improbable combinations of objects (e.g., ``a mix of a frog, teddy bear, and monster" or ``an orangutan accompanied by a pelican and a fish") in scenarios where only one entity was present in the given 3D object. Such discrepancies arise when different views of the same 3D object receive varied entity captions from BLIP2, which GPT4 then erroneously combines, shown in Figure~\ref{fig:method}. To detect this kind of case, we computed both the average and maximum CLIP scores between the final caption and all eight rendered views used in Cap3D. A validation set of $\sim 7k$ objects with inaccurate captions was annotated and used to determine two thresholds (mean \& max as shown in Figure~\ref{fig:clip}),  with the goal of encompassing all objects in this set. We then use the two selected thresholds to filter out $\sim 167k$ possible issued objects out of a total of $660k$.

\textbf{Image-Based Method:} Approximately $10k$ renderings in Cap3D dataset were identified as having all-grey images, likely due to rendering issues within the Cap3D process. We addressed this by re-rendering these objects and updating their captions with descriptions generated by our method (Section~\ref{sec:method:new_framework}). 

\textbf{Text-Based Method:} Attempting to identify errors solely based on captions proved challenging due to the diverse and complex nature of objects within Objaverse, making it difficult to detect hallucinations based on text alone. This complexity arises because some 3D objects genuinely comprise multiple or unusual components. Despite this, we developed a technique for identifying the misuse of terms related to ``image" and ``rendering", as these are directly associated with the rendering process rather than the 3D objects themselves. Through this method, we identified approximately 23,000 objects requiring correction.

\subsection{Dataset Expansion and Ethical Filtering}
Our expansion includes adopting the remaining objects of Objaverse, where Cap3D did not include, and high-quality 3D objects from Objaverse-XL's curated subset (Section 4.1 of \cite{deitke2023objaverse}, selected through human evaluation and heuristics from a pool of 10 million objects. This extension enhances the diversity and quality of our dataset. 

\begin{figure}[h]
    \centering
    \begin{minipage}{0.65\textwidth}
        \centering
        \includegraphics[width=1\textwidth]{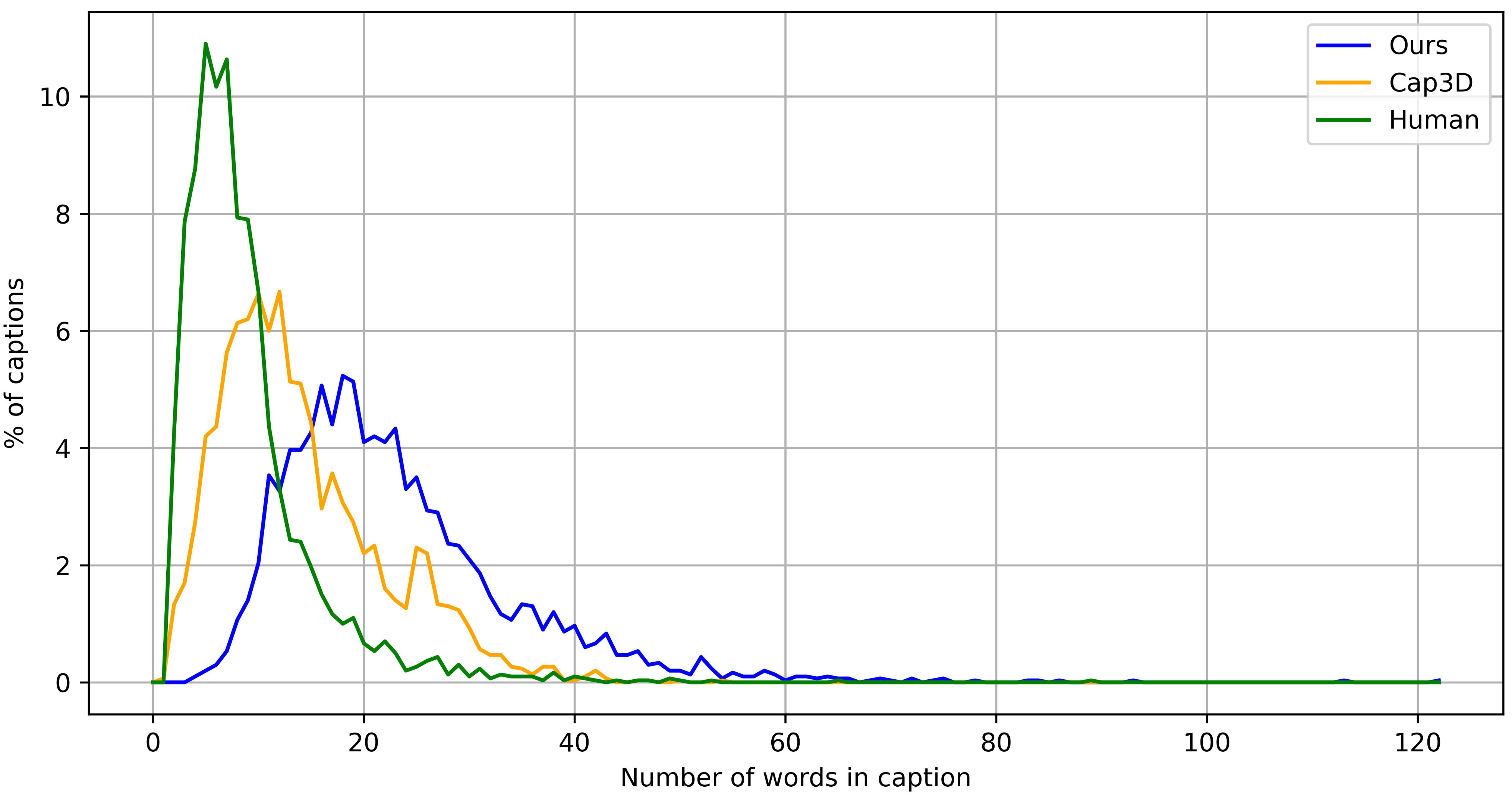}
        \caption{Number of words in caption.  }
        \label{fig:word_len_comp}
    \end{minipage}%
    \hspace{2mm}
    \scalebox{0.8}{
    \begin{minipage}{0.35\textwidth}
        \centering
        \begin{tabular}{lrrr}
\toprule
                          & Human & Cap3D & Ours \\ \midrule
Unigrams         & 2,876          & 2,767          & \textbf{5,600}         \\
Bigrams          & 11,374         & 12,293         & \textbf{29,521}        \\
Trigrams         & 16,535         & 23,062         & \textbf{52,457}        \\ \bottomrule
\end{tabular}
        \caption{Number of n-grams for captions generated by different methods.}
        \label{fig:n_gram}
    \end{minipage}
    }
\end{figure}

Moreover, we apply ethical filtering to both the rendered images and generated captions to remove potentially NSFW content and identifiable human faces, following Cap3D's protocol. We also leverage GPT4-Vision's internal detection capabilities for identifying images with potential ethical issues. It returns `content\_policy\_violation' once their model detection the image possibly against their safety policy. These comprehensive measures have allowed us to detect a list of $\sim 35k$ objects. 

We compared caption length and n-grams~\cite{bird2009natural} of captions among Human, Cap3D, and our captions in a 5k common set. As shown in Figure~\ref{fig:word_len_comp}, our captions usually contain longer length indicating more details than Cap3D and human-authored captions. Table~\ref{fig:n_gram} demonstrates we have the largest vocabulary size.

\section{Experiments}
In this section, we compare our captions against Cap3D captions and human-authored captions in terms of quality and hallucination degrees through human studies. We also ablate our methods to verify the effectiveness of the proposed DiffuRank. Then, we compare text-to-3D models finetuned on Cap3D and our updated Captions on the same set to measure the improvements of caption alignment at scale.  Finally, we further verify the effectiveness of our propose DiffuRank by examining it on a VQA task. For the sake of space, we list quantitative results here and include qualitative comparisons in Appendix~B and C.

\subsection{Captioning Evaluation}
\label{sec:exp:caption_evaluation}

\textbf{Settings.} We first evaluate the quality of captions generated by our method. Our captioning process involves selecting the top 6 captions out of a total of 28, as determined by DiffuPick, and then feeding these captions into GPT4-Vision (for further details, see Section~\ref{sec:dataset}). We evaluate the generated captions by comparing them to those produced by Cap3D, as well as to the human-authored captions that Cap3D provides. Our goal is to determine whether our method can produce captions of higher quality and with fewer inaccuracies or hallucinations.

Furthermore, we conduct ablation studies to assess the effectiveness of another component of our method, DiffuRank. We compare various approaches to highlight the benefits of DiffuRank: (1) \textbf{Allviews 28-views:} using all 28 rendered views as input to GPT4-Vision (details in Section~\ref{sec:method:new_framework}), (2) \textbf{Horizontal 6-views:} selecting 6 rendered views that place the camera horizontally across the object's default orientation, applying the same up and down positioning heuristics as Cap3D, and (3) \textbf{Bottom 6-views:} using the bottom-6 captions, defined as those with the worst alignment scores according to our DiffuRank algorithm (see Alg.~\ref{alg:diffurank}), as input to GPT4-Vision. Through these comparisons, we aim to demonstrate the impact of DiffuRank's selection process on the quality of the generated captions.

\textbf{Metrics.} Our primary evaluation method utilizes A/B testing with human judgment, where participants evaluate a pair of captions on a 1-5 scale, with 3 representing a neutral preference (i.e., tie). Our approach includes two distinct assessments: (a) evaluating which caption more accurately describes the object's type, appearance, and structure, and (b) determining which caption is less prone to presenting incorrect information or hallucinations. Each assessment involves over 10,000 ratings across 4,000 objects to ensure statistical reliability. We calculate and report the average scores and the frequency each option is preferred (i.e., excluding neutral (tie) responses). More human evaluation details are included in Appendix~B.7. Additionally, we follow Cap3D~\cite{luo2023scalable} and employ automated metrics, including CLIP score, measuring the cosine similarity between CLIP encodings and input images, and CLIP R percision~\cite{poole2022dreamfusion}, assessing the match between a rendered image and all potential texts.

\begin{table}[t]
\caption{\textbf{Objaverse Captions Evaluations}. All A/B testing represents captions from other methods vs. ours. We tested on 5k objects. }
\label{table:caption_evaluation}
\resizebox{\ifdim\width>\linewidth \linewidth \else \width \fi}{!}{
\begin{tabular}{lccc|ccc|cccc}
\toprule
\multirow{2}{*}{Method} & \multicolumn{3}{c}{Quality A/B test} & \multicolumn{3}{|c}{Hallucination A/B test} & \multicolumn{3}{|c}{CLIP} \\ 
 & Score(1-5) & Win \% & Lose \%  & Score(1-5) & Win \% & Lose \% & Score & R@1 & R@5 & R@10\\
\midrule
Human & 2.57 & 31.9 & 62.1 & 2.88 & 39.9 & 46.4 & 66.2& 8.9 & 21.0 &27.8 \\
Cap3D & 2.62 & 32.7 & 60.2 & 2.43 & 25.8 & 63.9 & 71.2& 20.5& 40.8 & 51.9 \\
Ours & - & - & - & - & - & - &\textbf{74.6}& \textbf{26.7} & \textbf{48.2} &\textbf{57.5}   \\
\midrule
\midrule
Allviews 28-views & 2.91 & 37.9 & 43.6 & 2.85 & 35.1 & 47.2 & 73.5 & 24.9	&46.7 & 55.7  \\
Horizontal 6-views & 2.84 & 35.2 & 44.5  & 2.90 & 36.2 & 40.9 &  73.8 & 25.8 & 46.7 & 55.9  \\
Bottom 6-views & 2.74 & 31.1 & 52.0 & 2.61 & 30.1 & 57.0 & 72.8 & 24.6& 45.1	&55.2 \\
\bottomrule
\end{tabular}
}
\end{table}

\textbf{Results.} The evaluation results, presented in Table~\ref{table:caption_evaluation}, highlight the effectiveness of our captioning approach. According to scores from human evaluators on quality and hallucination metrics, our captions feature more accurate details with fewer instances of hallucination, compared to Cap3D and human-authored captions. Supporting qualitative findings are detailed in Appendix~B.3, reinforcing these conclusions.

A comparison of our method, which selects the top-6 views, with alternatives—the bottom-6 views and horizontally placed 6-views—demonstrates the impact of DiffuRank on performance. Specifically, as depicted in Figure~\ref{fig:diffurank_effects}, bottom-6 views often relate less to the 3D object as they may capture only the back or bottom. This issue highlights the difficulties arising from Objaverse's random default orientation, positioning cameras `horizontally' does not always ensure they are actually horizontal. More qualitative compairsons between the three types of view selection are included in Appendix~B.5. Furthermore, DiffuRank does not consistently achieve optimal performance, as illustrated by the selection of the 6th image in the first row (referenced in Figure~\ref{fig:diffurank_effects}) captioned 'a blue laptop.' Enhancements could be achieved through using an improved text-to-3D diffusion models, a topic explored in detail in Appendix~E.

Furthermore, our approach outperforms the variant using 24 views, delivering captions with greater detail and fewer hallucinations (See qualitative comparisons at Appendix~B.4). Interestingly, providing a larger number of views (24) does not necessarily improve details; it appears to complicate the model's ability to access precise information due to the variance in detail across different perspectives. This observation contradicts expectations, suggesting an optimal balance of view selection is crucial for accurate 3D object captioning.

\begin{table}[t]
\caption{\textbf{Text-to-3D Finetuning experiments}.}
\label{tab:text_3d}
\begin{center}
\begin{tabular}{lccccc}
\toprule
& \multirow{2}{*}{FID$\downarrow$} & CLIP  & \multicolumn{3}{c}{CLIP R-Precision (2k)} \\
& & Score & R@1 & R@5 & R@10 \\
\midrule
Ground Truth Images & - & 81.6 & 32.7 &	55.1 & 64.3 \\
\midrule
Point·E & 36.1 & 61.5 & 3.4 & 10.4 & 15.3 \\
Point·E + Cap3D & 32.8 & 65 & 7.1 & 19.4 & 26.4 \\
Point·E + Ours (330k) & 32.4 & 66.2 & 8.1 & 20.3 & 28.5 \\
Point·E + Ours (825k) & \textbf{31.2} & \textbf{66.5} & \textbf{10.1} & \textbf{21.9} & \textbf{29.8} \\
\midrule
Shap·E (STF) & 37.2 & 68.8 & 12.7 & 29.0 & 37.9 \\
Shap·E (STF) + Cap3D & 35.5 & 68.2 & 11.9 & 28.8 & 37.4 \\
Shap·E (STF) + Ours (330k) & 35.6 & 69.4 & 13.4 & 29.7 & 39.3 \\
Shap·E (STF) + Ours (825k) & \textbf{34.3} & \textbf{69.8} & \textbf{14.9} & \textbf{33.7} & \textbf{42.8} \\
\midrule
Shap·E (NeRF) & 48.7 & 68.3 & 12.2 & 27.9 & 36.2 \\
Shap·E (NeRF) + Cap3D & 48.2 & 68.0 & 11.7 & 27.1 & 35.1 \\
Shap·E (NeRF) + Ours (330k) & 48 & 68.4 & 13.2 & 29.3 & 38.4 \\
Shap·E (NeRF) + Ours (825k) & \textbf{47.9} & \textbf{69.3} & \textbf{14.3} & \textbf{31.7} & \textbf{40.4} \\
\bottomrule
\end{tabular}
\end{center}
\end{table}

\subsection{Text-to-3D Generation with New Captions}
\label{sec:exp:text-to-3D}
\textbf{Settings.} This section we finetune Text-to-3D models to check if our updated captions can bring more improvements compared to Cap3D captions. For this purpose, we would mainly conduct experiments over point-E~\cite{nichol2022point} and shap-E~\cite{nichol2023shape} as they are used in Cap3D. We follow the same setting as Cap3D, including learning rate, batch size, optimizer, and steps. We adopted the same 330k training split and test split used in \cite{luo2023scalable}, and we have updated $72k$ captions in this 330k set ($\sim$20\%). Additionally, we scale our experiment up, and train models with 825k ($2.5\times 330k)$ data from our full 3D-text pairs. More details and qualitative results are included in Appendix~C.

\textbf{Metrics.}  We incorporated the use of CLIP Score and CLIP R-Precision \cite{poole2022dreamfusion, luo2023scalable} in our evaluation process. CLIP R-Precision involves ranking a rendered image among all text pairs within the test set based on the cosine similarity as measured by CLIP, then determining the precision based on accurate text-image matches. Given the availability of ground truth images, we employed the FID metric to compare the fidelity of 3D rendered images with these true images.

\textbf{Results.} 
Results are showcased in Table~\ref{tab:text_3d}. Considering we updated nearly 20\% captions of the 330k training set for Cap3D 3D-text pairs, we anticipated some improvement, albeit modest. However, the improvements exceeded our expectations. Our enhanced model (`model + Ours' with 330K data points) consistently outperformed both the `model + Cap3D' (with 330K data points) version and pre-trained Shap·E model. Surpassing the pre-trained Shap·E model is non-trivial, as the 'model + Cap3D' version generally showed declining performance when compared to the pre-trained Shap·E model, indicating that fine-tuning on Cap3D data actually harms the performance. The performance enhancement achieved by correcting 20\% of the data underscores the effectiveness of addressing misalignments in the 3D-text of Cap3D by locating the potential errors and refining with our new captioning approach. Furthermore, by expanding our dataset by 2.5 times, we have boosted performance across multiple metrics and models. Given that Shap·E model was trained on proprietary data, our findings suggest that our proposed 3D-text dataset could be a competitive open-source alternative.

\subsection{DiffuRank on VQA}
\label{sec:exp:diffurank}

\begin{figure}[ht]
    \centering
    \begin{minipage}{0.56\textwidth}
        \centering
        \begin{tabular}{lr}
            \toprule
            \textbf{Model} & \textbf{Accuracy (\%)} \\
            \midrule
            Human & 95.7 \\
            Gemini~\cite{team2023gemini} & 40.7 \\
            GPT4-Vision~\cite{gpt4} & 38.7 \\
            Ours & 30.7\\
            Random Guess & 25.0 \\
            LLaVA-1.5~\cite{liu2024visual} & 24.7 \\
            Bard & 19.0 \\
            Bing Chat & 17.3 \\
            InstructBLIP~\cite{dai2023instructblip} & 16.7 \\
            CLIP~\cite{radford2021learning} & 13.3\\
            mini-GPT4~\cite{zhu2023minigpt} & 12.7 \\
            LLaVA~\cite{liu2024visual} & 6.0 \\
            \bottomrule
        \end{tabular}
        \captionof{table}{Accuracy comparison among various VLMs, CLIP, and our method. }
            \label{tab:vqa}
    \end{minipage}%
    \hspace{1mm}
    \begin{minipage}{0.3\textwidth}
        \centering
        \includegraphics[width=\textwidth]{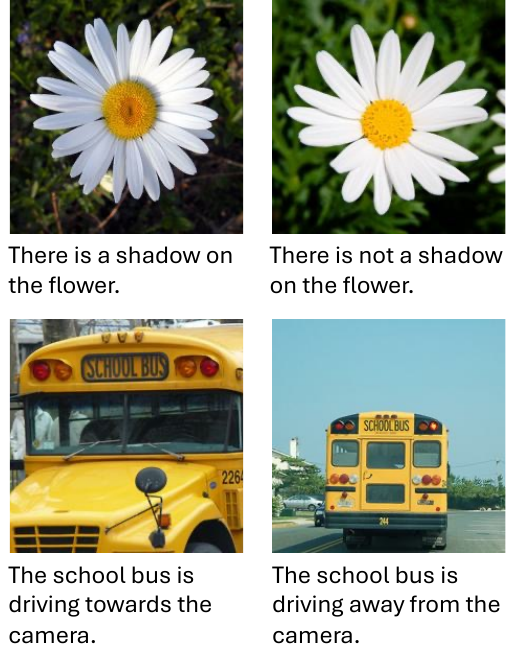}
        \caption{Each row represents a matched pair, and the accompanying text beneath it is the description.}
        \label{fig:vqa}
    \end{minipage}
\end{figure}

\textbf{Settings.} We extend our DiffuRank to solve Visual Question Answering task, with the help of a pre-trained text-to-2D diffusion model~\cite{rombach2022high}. We list our detailed settings and the updated algorithm in Appendix~D. We mainly compare to CLIP~\cite{radford2021learning} in terms of zero-shot VQA performance and test on the Multimodal Visual Patterns (MMVP) benchmark~\cite{tong2024eyes}, comprising nine fundamental visual patterns across 150 images pairs. Each pair of images (Figure~\ref{fig:vqa}), despite having clear visual distinctions, are perceived similarly by the CLIP model. Each pair is associated with a question that has two divergent answers. Numerous Vision-Language Models (VLMs) have been shown to underperform on this challenging benchmark.

Given that the task involves Visual Question Answering (VQA), neither our approach nor the CLIP model is inherently designed to generate textual responses directly. To address this, we employed GPT-4 to transform each question and its corresponding answers into declarative statements. Consequently, for each image pair, we obtained two distinct statements. For DiffuRank, we executed multiple iterations of alignment estimation for the statements corresponding to each image, selecting the statement with the highest alignment estimate as the correct answer/statement. For CLIP model, we determined the appropriate answer by calculating the cosine similarity between an image and each statement, choosing the statement with the greatest similarity as the response. We used ``ViT-B/32" CLIP here for evaluation.

\textbf{Metrics.} Our evaluation metrics are aligned with those proposed by \cite{tong2024eyes}. A model's response is deemed accurate only if it correctly identifies the appropriate statements for both images in a pair. Hence, if a model accurately selects the correct statement for only one image within the pair, its attempt is marked as incorrect. It is important to note that both DiffuRank and CLIP may occasionally select identical statements for different images within the same pair.

\textbf{Results.} Table~\ref{tab:vqa} shows the quantitative results which demonstrate DiffuRank significantly outperforms CLIP in the MMVP benchmark with the help of pre-trained stable diffusion model. Also, for the example pairs shown on the Figure~\ref{fig:vqa}, our method is able to select the correct corresponding image-statement pairs. In contrast, the CLIP model incorrectly selects There is not a shadow on the flower' and The school bus is driving towards the camera' for both images in each pair.

\section{Conclusion}
This paper help alleviate inaccuracies and hallucinations in Cap3D captions (a 3D-Text dataset for Objaverse), attributed to suboptimal render views based on default object orientations. We introduced DiffuRank to address this issue, a method that ranks rendered views by their alignment with 3D object information using pre-trained text-to-3D diffusion models. Combining DiffuRank and GPT4, our new captioning approach improved caption quality, reduced inaccuracies, and enhanced detail richness with fewer views. Our efforts have not only improved the quality of existing Cap3D captions but also expanded the dataset to cover a total of 1M 3D-text pairs (whole Objaverse and a subset of Objaver-XL highquality set). We also extended DiffuRank's application to the 2D domain, demonstrating its effectiveness in Visual Question Answering tasks.

\section*{Acknowledgement}

This work has been made possible through the generous support of the ``Efficient and Scalable Text-to-3D Generation" grant from LG AI Research. We also thank the OpenAI Researcher Access Program for partially supporting our use of the GPT-4 API. We greatly appreciate Chris Rockwell for his invaluable technical support in caption evaluation, and Mohamed El Banani for his insightful feedback to our initial draft. Tiange thanks Minghua Liu and Jiaming Song for their insightful discussions back at NeurIPS 2023 in NOLA.

\bibliographystyle{splncs04}
\bibliography{main}

\clearpage
\appendix

\section*{Table of Contents}
\noindent \hyperlink{broaderimpact}{A\, Broader Impact}\\
\hyperlink{B}{B\, Dataset: more details \& results}\\
\hspace*{10mm}\hyperlink{B1}{B.1 Extra dataset details}\\
\hspace*{10mm}\hyperlink{B2}{B.2 Captions: overcome the failure cases in Cap3D}\\
\hspace*{10mm}\hyperlink{B3}{B.3 Captions: Ours vs. Cap3D vs. human-authored}\\
\hspace*{10mm}\hyperlink{B4}{B.4 Captions: Ours vs. ablated variants}\\
\hspace*{10mm}\hyperlink{B5}{B.5 DiffuRank: Ours vs. bottom 6-views vs. horizontal 6-views}\\
\hspace*{10mm}\hyperlink{B6}{B.6 Failure cases}\\
\hspace*{10mm}\hyperlink{B7}{B.7 Human evaluation details}\\
\hyperlink{C}{C\, Text-to-3D: more details \& results}\\
\hspace*{10mm}\hyperlink{C1}{C.1 Setting}\\
\hspace*{10mm}\hyperlink{C2}{C.2 Qualitative comparisons}\\
\hyperlink{D}{D\, DiffuRank on VQA}
\hyperlink{E}{E\, Future Work \& Limitations}

\hypertarget{broaderimpact}{\section{Broader Impact}}
\label{appen:broader_impact}
By enhancing the accuracy and richness of captions for 3D objects, this work facilitates advancements in 3D modeling and prompote related applications in educational tools, interactive learning environments, and assistive technologies, making digital content more accessible and informative. Moreover, by addressing inaccuracies and hallucinations in captions which could be used in AI-content generations, our work underscores the pursuit of more reliable and trustworthy AI systems. During the process, we undertaken with a commitment to ethical considerations to filter out potential ethical issued 3D objects. We recognize the wide-reaching effects of our work on society and maintain that it chiefly offers positive contributions towards the progress of generative modeling and its implementation in diverse fields.

\hypertarget{B}{\section{Dataset: more details \& results}}
\label{appen:dataset}
\hypertarget{B1}{\subsection{Extra dataset details}}
In Section~\ref{sec:dataset}, we addressed approximately $200k$ caption corrections for the Cap3D dataset, significantly reducing its hallucinations. Our efforts also expand the dataset to include over 1 million 3D-text pairs, encapsulating the entirety of the Objaverse~\cite{deitke2022objaverse} and portions of the Objaverse-XL high-quality set~\cite{deitke2023objaverse}. The objects with updated captions are cataloged in a CSV file within the supplementary material, accessible via ``uid" or cryptographic hash values ("sha256"). These identifiers correspond to the ones provided in the Objaverse and Objaverse-XL datasets.

As mentioned in the Introduction, we are excited to also provide access to rendered images associated with each object. These images include detailed camera information (both intrinsic fov and extrinsic RT matrix), depth map, and MatAlpha, in addition to point clouds that complement the textual captions. Alongside these resources, we are releasing the source code for our DiffuRank methodology, which facilitates the replication of our findings. The distribution also includes pre-trained models, further aiding in the exploration and utilization of our dataset. This comprehensive package aims to empower researchers in our community. They will be released under ODC-By 1.0 license.

Our GPT4-Vision prompt is defined as ``Renderings show different angles of the same set of 3D objects. Concisely describe 3D object (distinct features, objects, structures, material, color, etc) as a caption" accompanied by six image tokens. On average, the context encompasses approximately 1,867 tokens, while the average number of tokens generated stands at approximately 26.72. Notably, we employed the "GPT-4-1106-vision-preview" model for this study.

As described in Section~\ref{sec:method:new_framework}, given a 3D object, we generate 28 views using two distinct rendering methods~\cite{luo2023scalable, jun2023shap}. For each view, we generate 5 captions with BLIP2. Subsequently, we apply the DiffuRank algorithm (Algorithm~\ref{alg:diffurank}) to evaluate the alignment of the 28 renderings relative to the input 3D object by doing inference ovew 140 captions and the 3D object. Ultimately, we select the best 6 views for further caption generation using GPT4-Vision. 

For the ray-tracking render engine, we used Blender render engine `CYCLES' with samples 16. Additionally, we adopted `OPTIX' denoiser for the cycle engine. For the real-time render engine, we used Blender render engine `EEVEE' with `taa\_render\_samples' 1.

\hypertarget{B2}{\subsection{Captions: overcome the failure cases in Cap3D}}
The Cap3D captions we used to compare throughout the whole paper are from their \href{https://huggingface.co/datasets/tiange/Cap3D/blob/main/Cap3D_automated_Objaverse.csv}{dataset page}. Specically, the version described in their \href{https://arxiv.org/pdf/2306.07279.pdf}{paper}.

\begin{figure}[h]
    \centering
    \includegraphics[width=\textwidth]{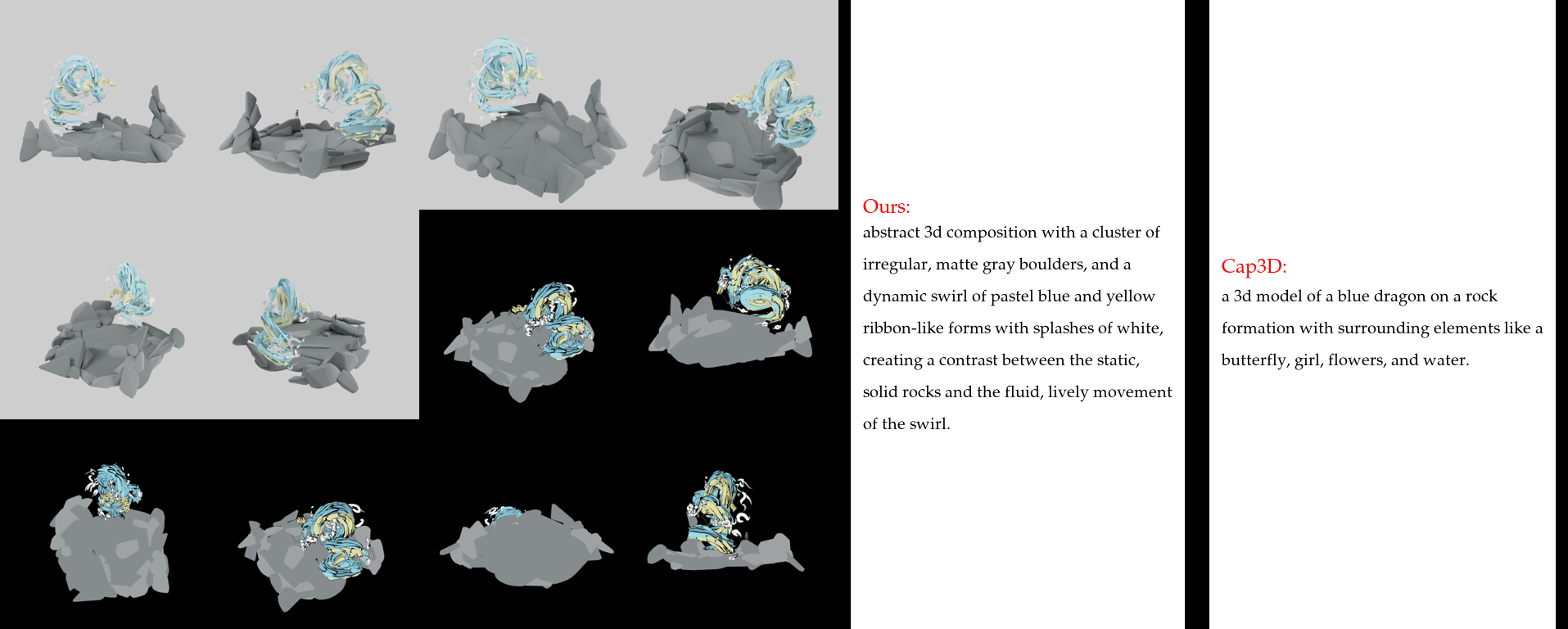}
    \caption{Comparisons between our captions and Cap3D captions.}
    \label{fig:cmp_1}
\end{figure}

\begin{figure}[h]
    \centering
    \includegraphics[width=\textwidth]{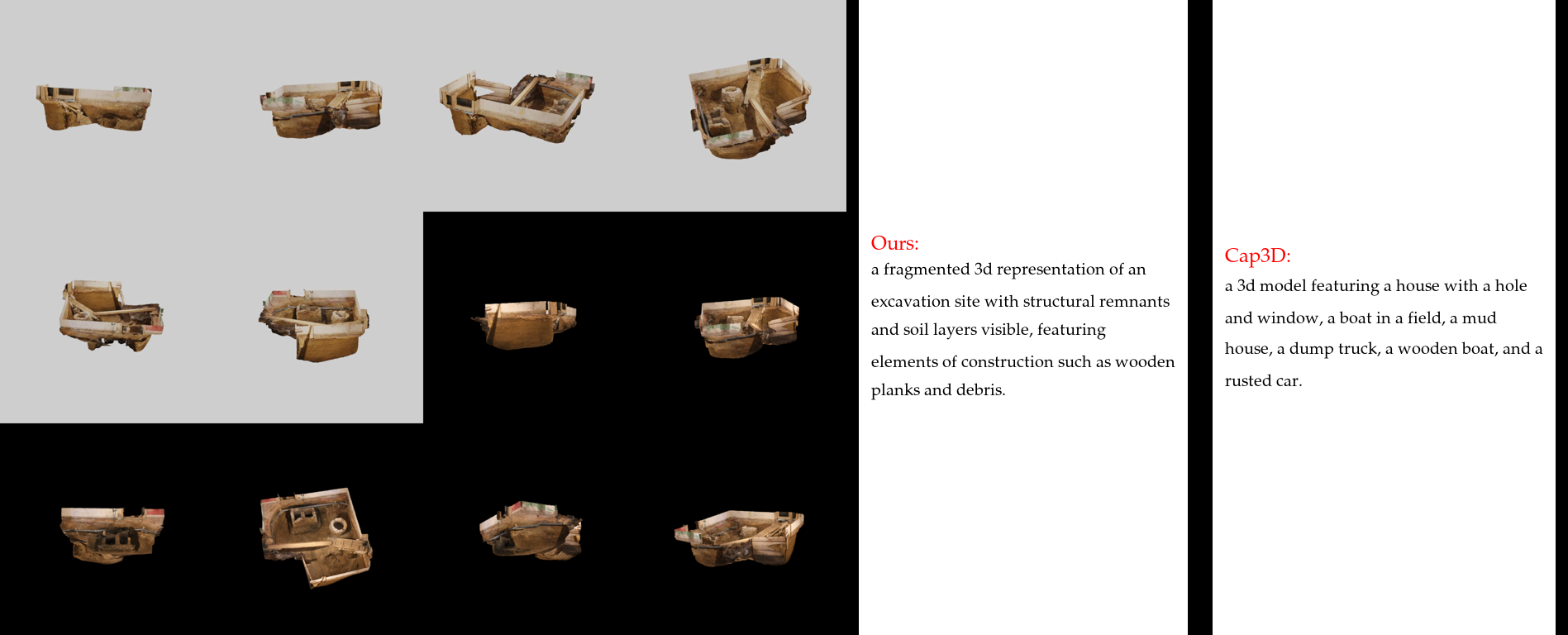}
    \caption{Comparisons between our captions with Cap3D captions.}
    \label{fig:cmp_2}
\end{figure}

Here, we provide direction comparisons with the failure cases mentioned in their paper ``Limitations and Failure Cases". Our captions have obviously eliminated lots of hallucinations, such as `butterfly' and `flowers' in Figure~\ref{fig:cmp_1}, and `dump truck' in Figure~\ref{fig:cmp_2}. 

\clearpage
\hypertarget{B3}{\subsection{Captions: Ours vs. Cap3D vs. human-authored}}

\label{appen:datasets:comparisons}
We present a variety of qualitative comparisons: those generated by our model, those produced by Cap3D, and captions written by humans, all of which were selected through random sampling. The below qualitative results show the captions generated by our method usually contain more details and less hallucinations. 

\begin{figure}[h]
    \centering
    \includegraphics[width=\textwidth]{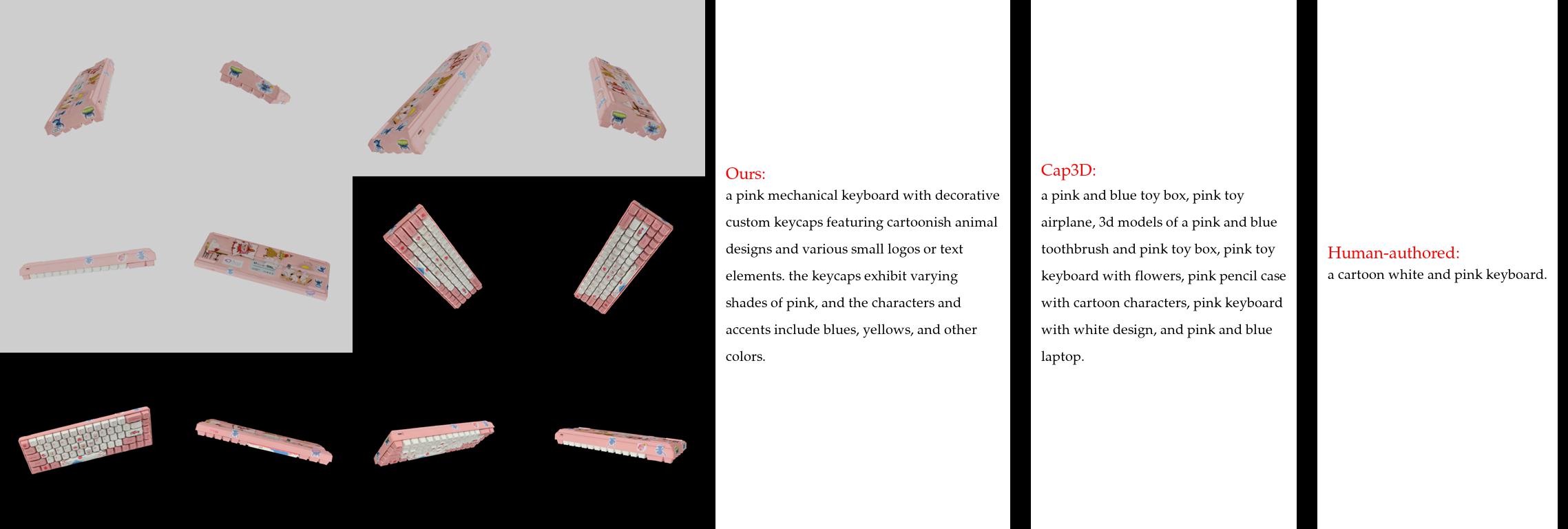}
    \caption{We compare captions through random sampling, including those generated by our method, by Cap3D, and those authored by humans.}
    \label{fig:appen_main_cmp_01}
\end{figure}

\begin{figure}[h]
    \centering
    \includegraphics[width=\textwidth]{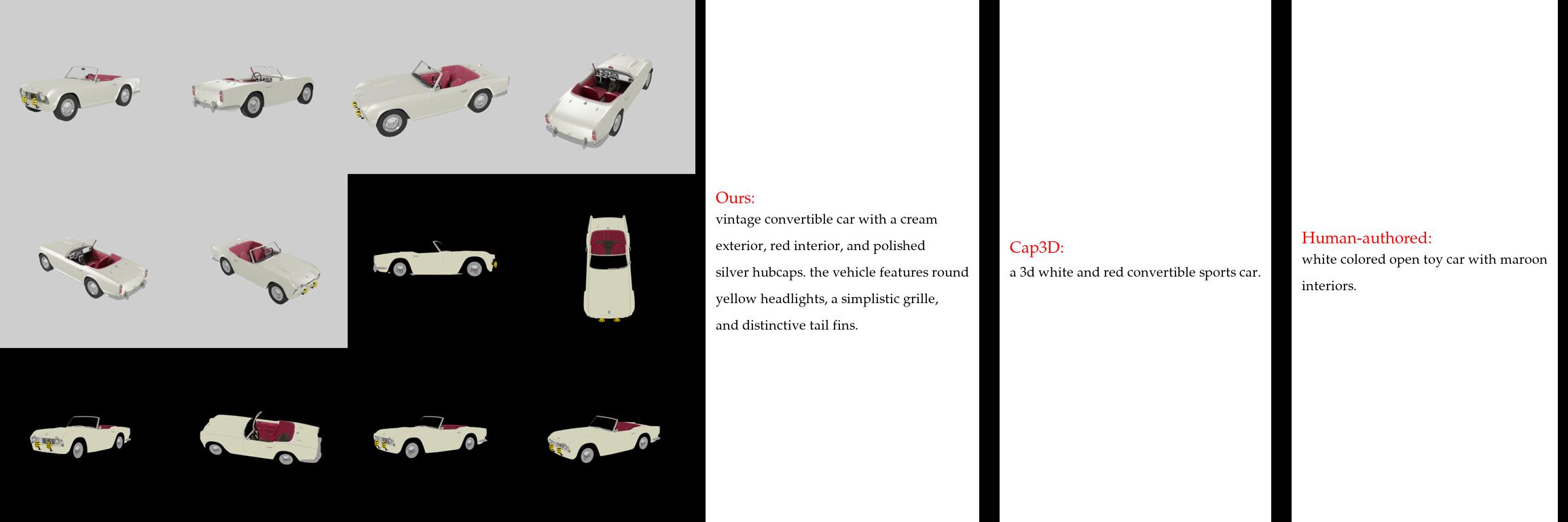}
    \caption{We compare captions through random sampling, including those generated by our method, by Cap3D, and those authored by humans.}
    \label{fig:appen_main_cmp_02}
\end{figure}

\begin{figure}[h]
    \centering
    \includegraphics[width=\textwidth]{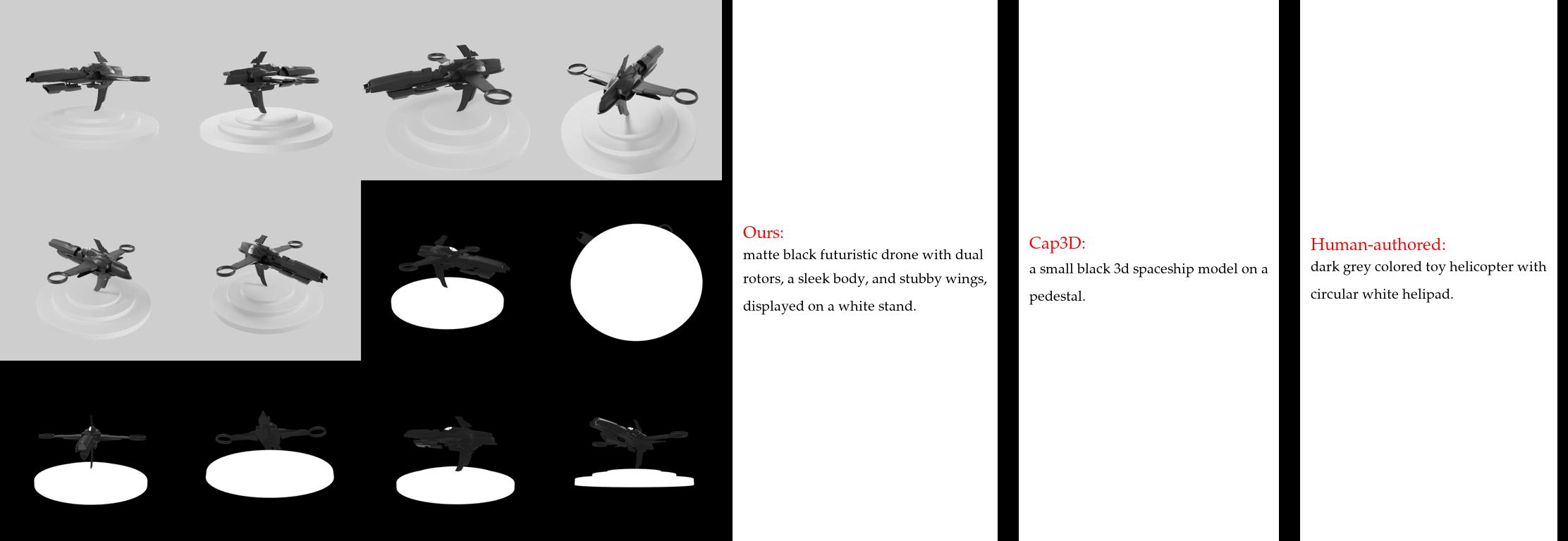}
    \caption{We compare captions through random sampling, including those generated by our method, by Cap3D, and those authored by humans.}
    \label{fig:appen_main_cmp_03}
\end{figure}

\begin{figure}
    \centering
    \includegraphics[width=\textwidth]{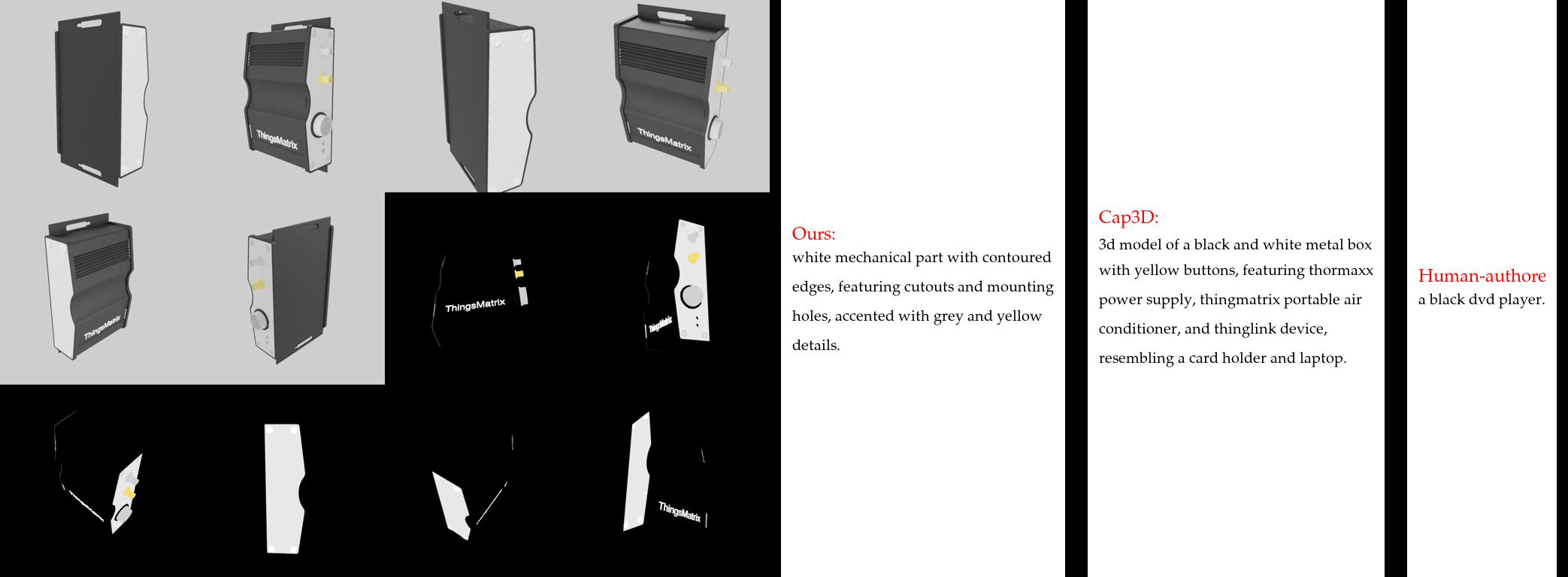}
    \caption{We compare captions through random sampling, including those generated by our method, by Cap3D, and those authored by humans.}
    \label{fig:appen_main_cmp_04}
\end{figure}

\begin{figure}
    \centering
    \includegraphics[width=\textwidth]{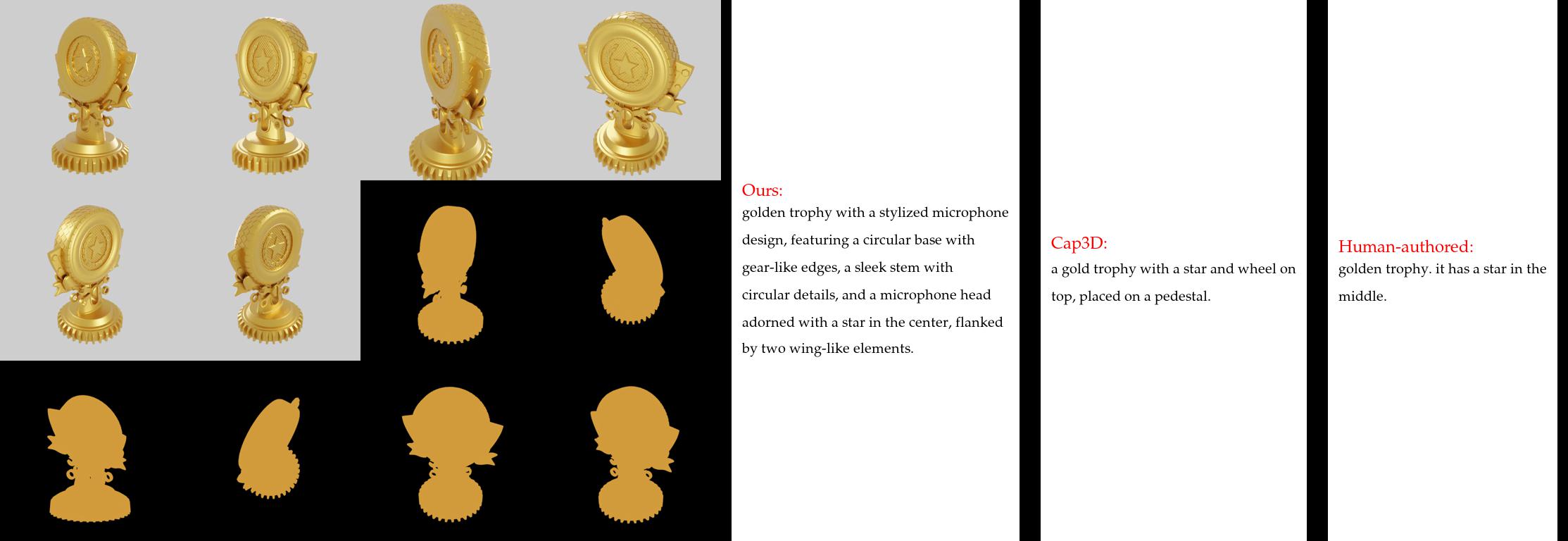}
    \caption{We compare captions through random sampling, including those generated by our method, by Cap3D, and those authored by humans.}
    \label{fig:appen_main_cmp_05}
\end{figure}

\begin{figure}
    \centering
    \includegraphics[width=\textwidth]{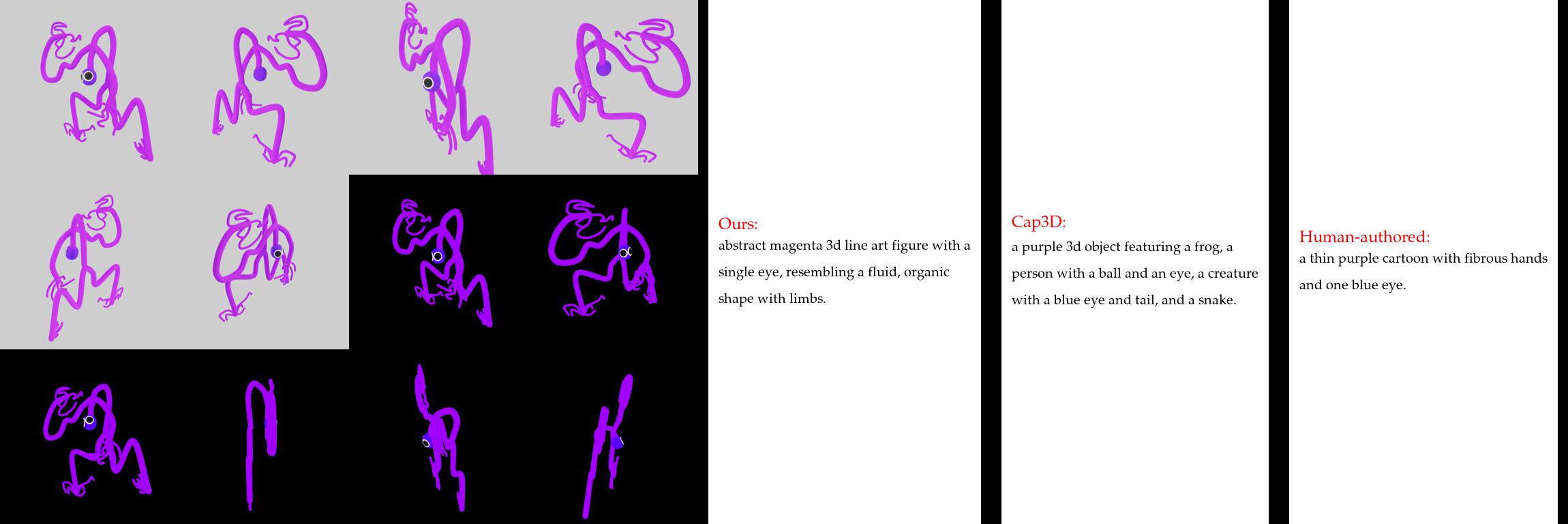}
    \caption{We compare captions through random sampling, including those generated by our method, by Cap3D, and those authored by humans.}
    \label{fig:appen_main_cmp_06}
\end{figure}

\begin{figure}
    \centering
    \includegraphics[width=\textwidth]{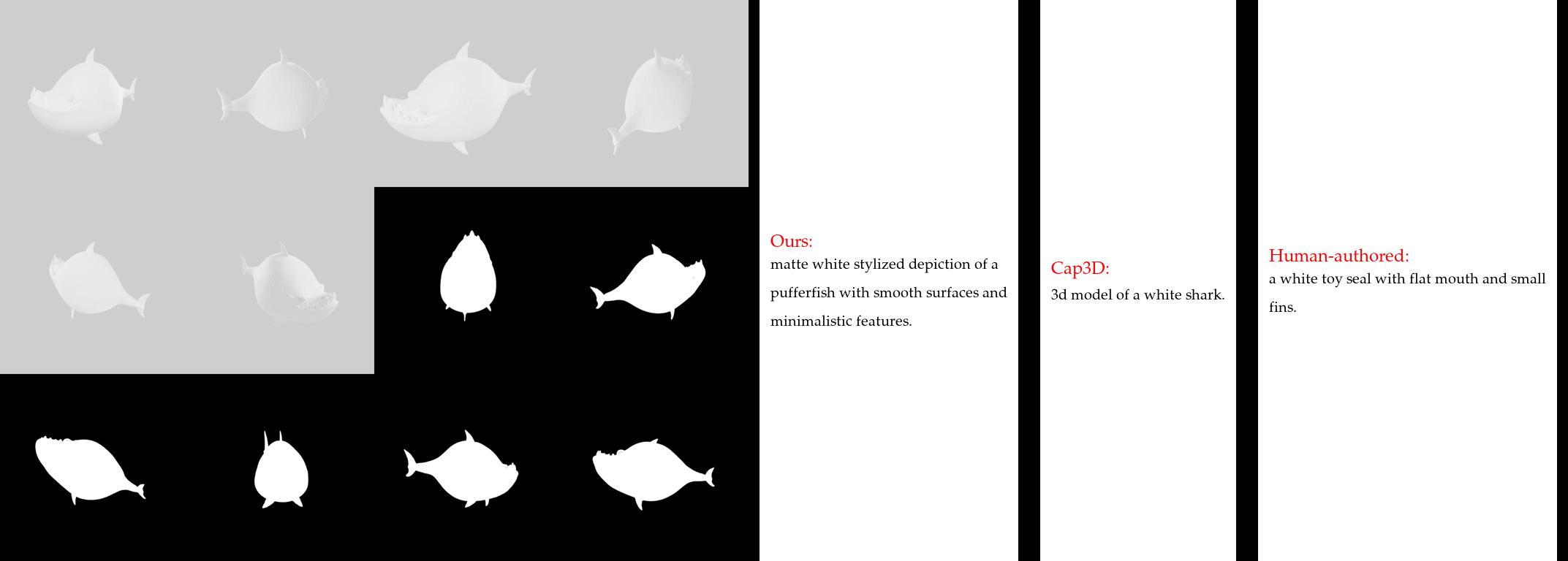}
    \caption{We compare captions through random sampling, including those generated by our method, by Cap3D, and those authored by humans.}
    \label{fig:appen_main_cmp_07}
\end{figure}

\begin{figure}
    \centering
    \includegraphics[width=\textwidth]{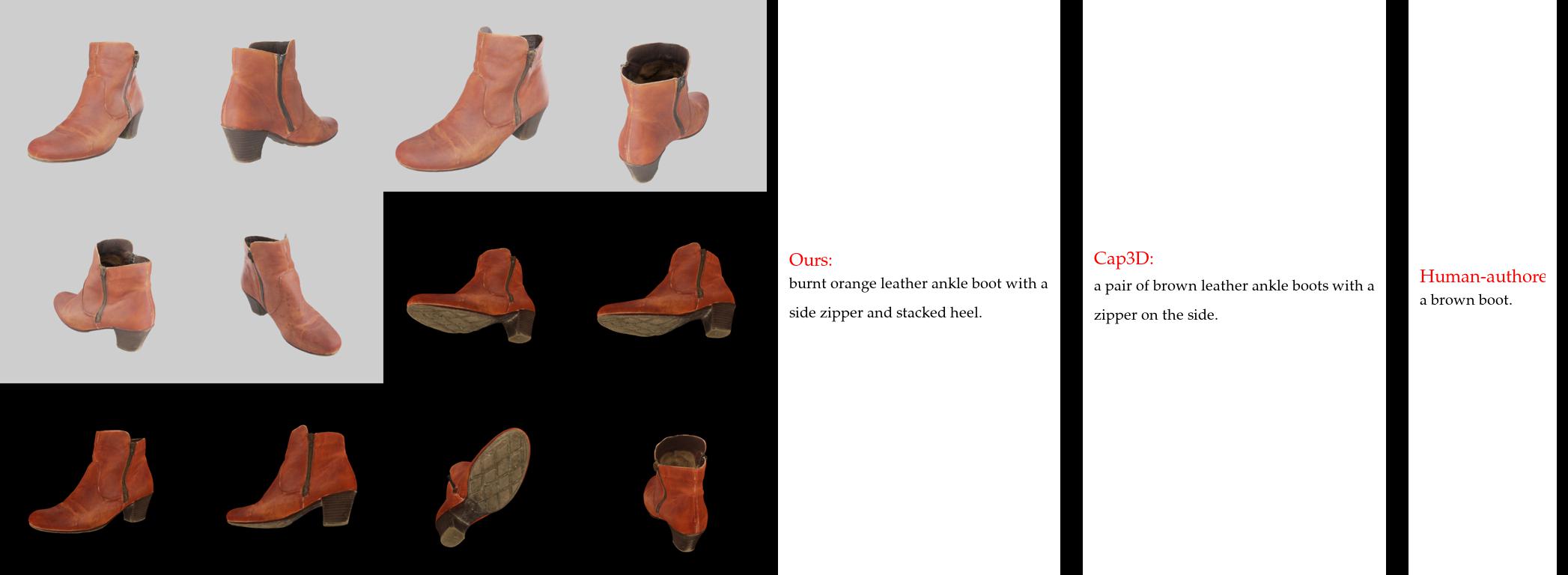}
    \caption{We compare captions through random sampling, including those generated by our method, by Cap3D, and those authored by humans.}
    \label{fig:appen_main_cmp_08}
\end{figure}

\begin{figure}
    \centering
    \includegraphics[width=\textwidth]{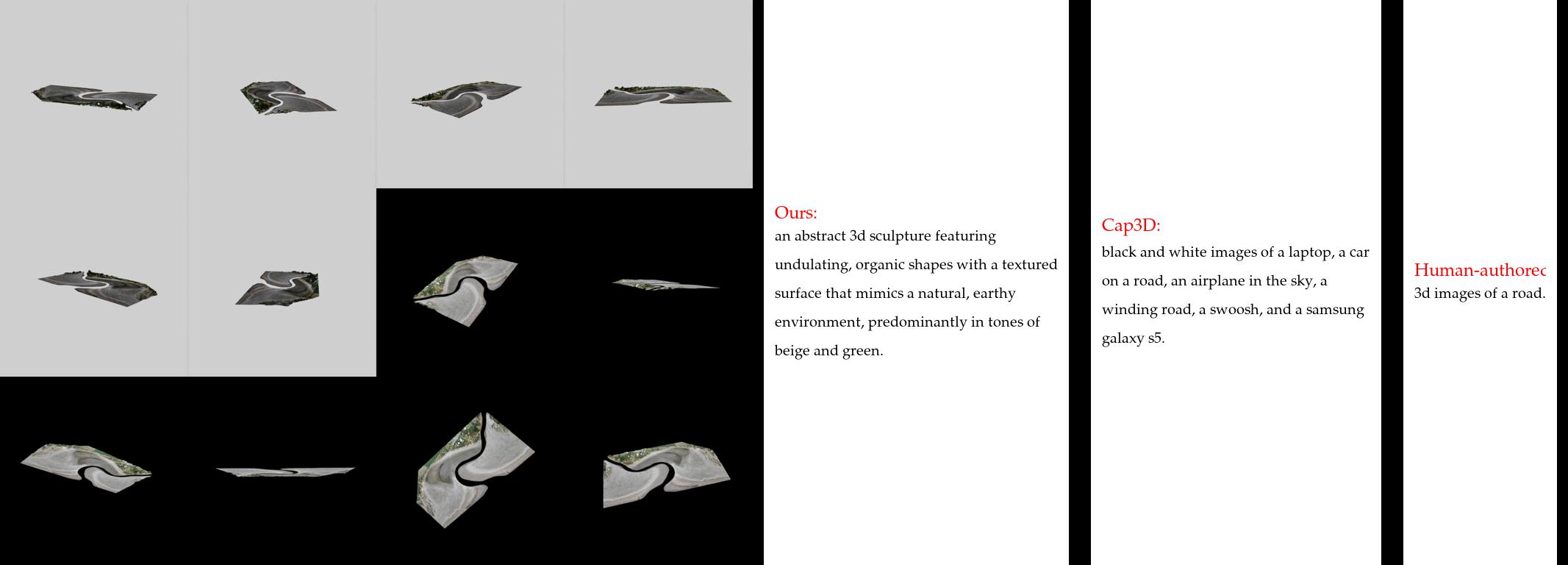}
    \caption{We compare captions through random sampling, including those generated by our method, by Cap3D, and those authored by humans.}
    \label{fig:appen_main_cmp_09}
\end{figure}

\begin{figure}
    \centering
    \includegraphics[width=\textwidth]{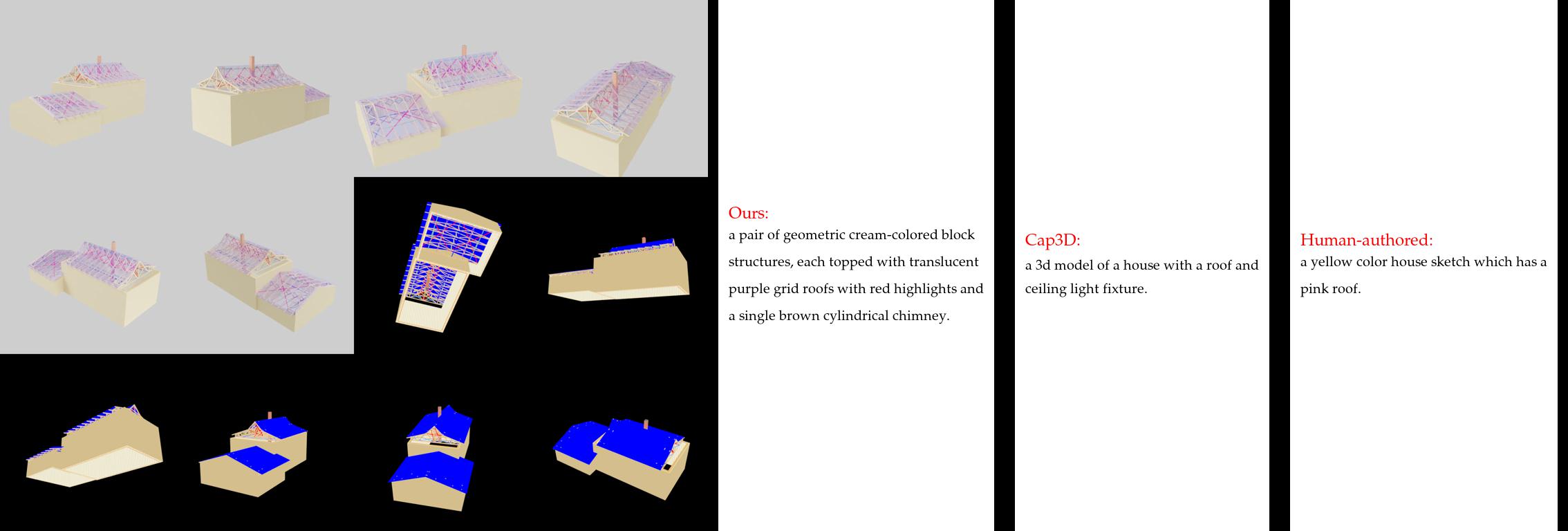}
    \caption{We compare captions through random sampling, including those generated by our method, by Cap3D, and those authored by humans.}
    \label{fig:appen_main_cmp_10}
\end{figure}

\begin{figure}
    \centering
    \includegraphics[width=\textwidth]{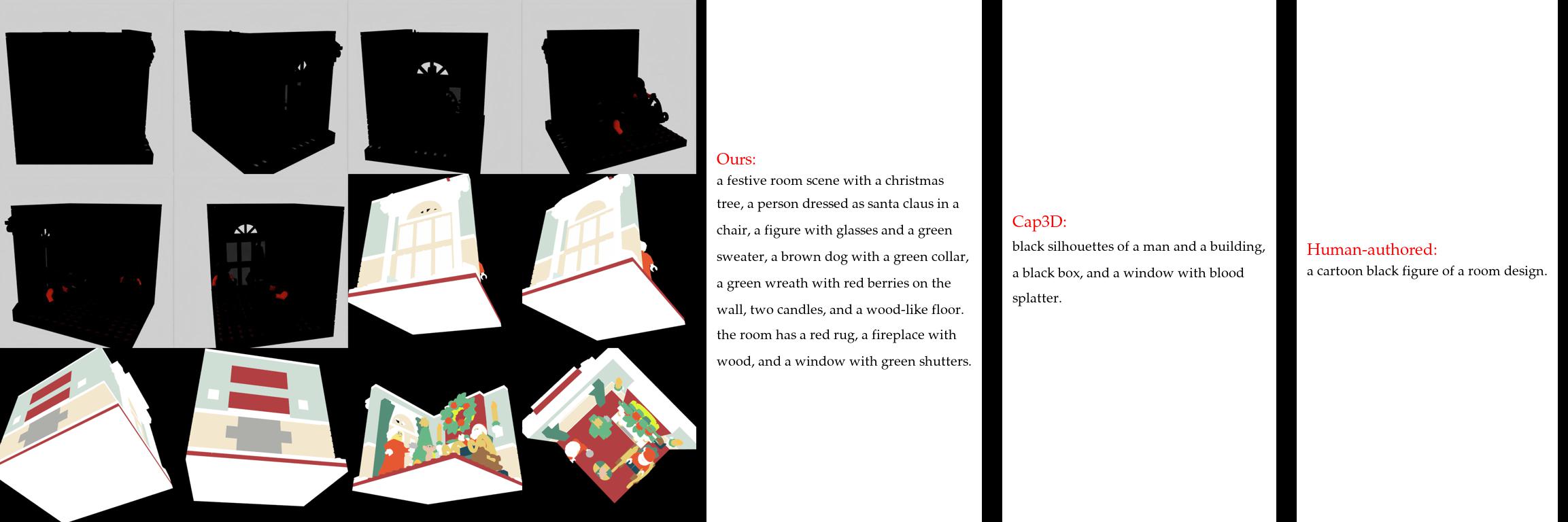}
    \caption{We compare captions through random sampling, including those generated by our method, by Cap3D, and those authored by humans.}
    \label{fig:appen_main_cmp_11}
\end{figure}

\begin{figure}
    \centering
    \includegraphics[width=\textwidth]{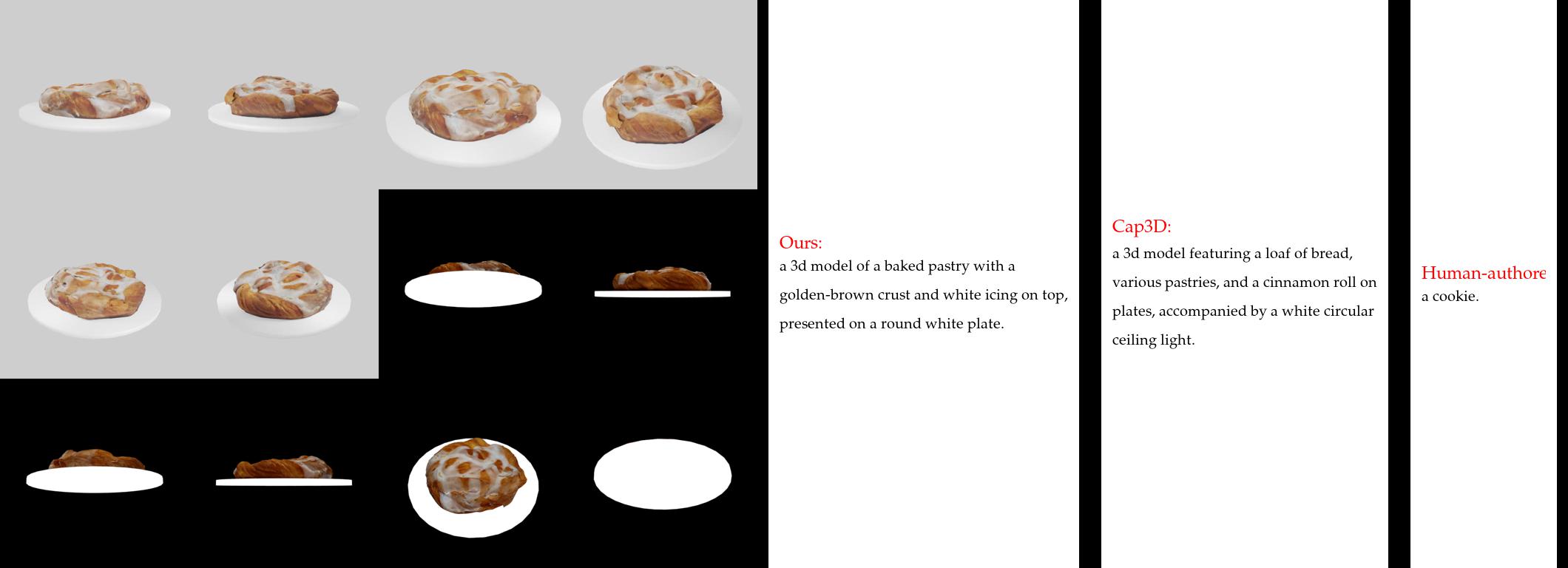}
    \caption{We compare captions through random sampling, including those generated by our method, by Cap3D, and those authored by humans.}
    \label{fig:appen_main_cmp_12}
\end{figure}

\begin{figure}
    \centering
    \includegraphics[width=\textwidth]{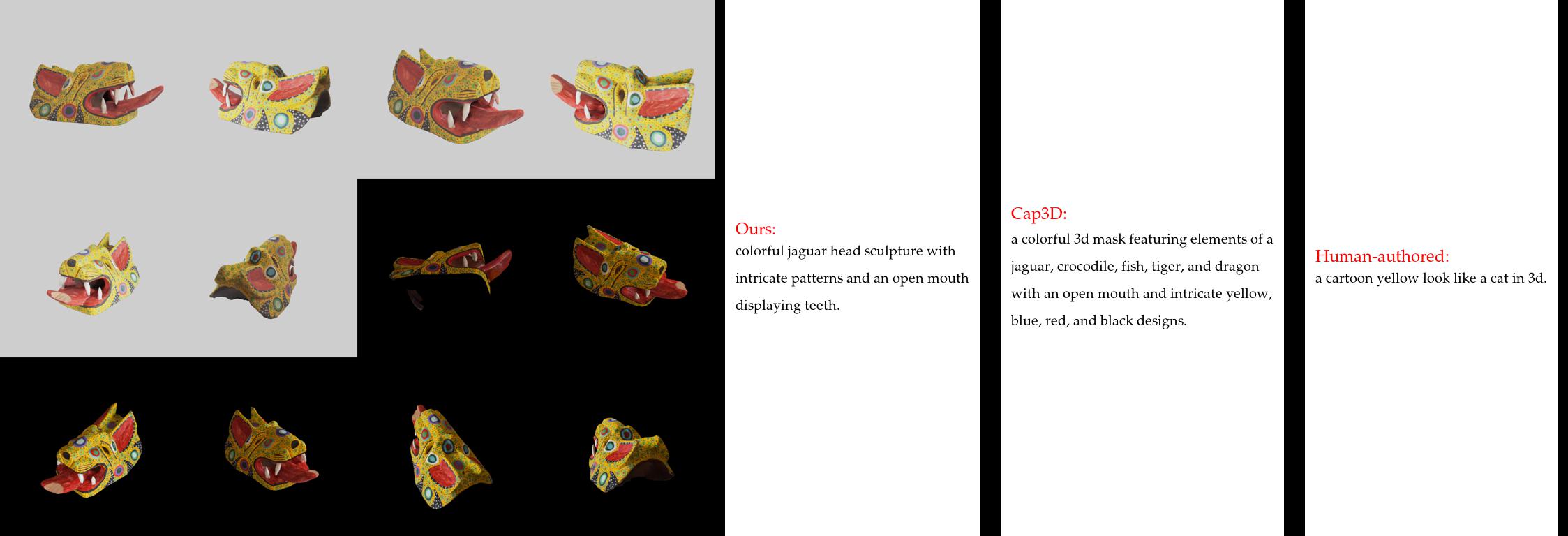}
    \caption{We compare captions through random sampling, including those generated by our method, by Cap3D, and those authored by humans.}
    \label{fig:appen_main_cmp_13}
\end{figure}

\begin{figure}
    \centering
    \includegraphics[width=\textwidth]{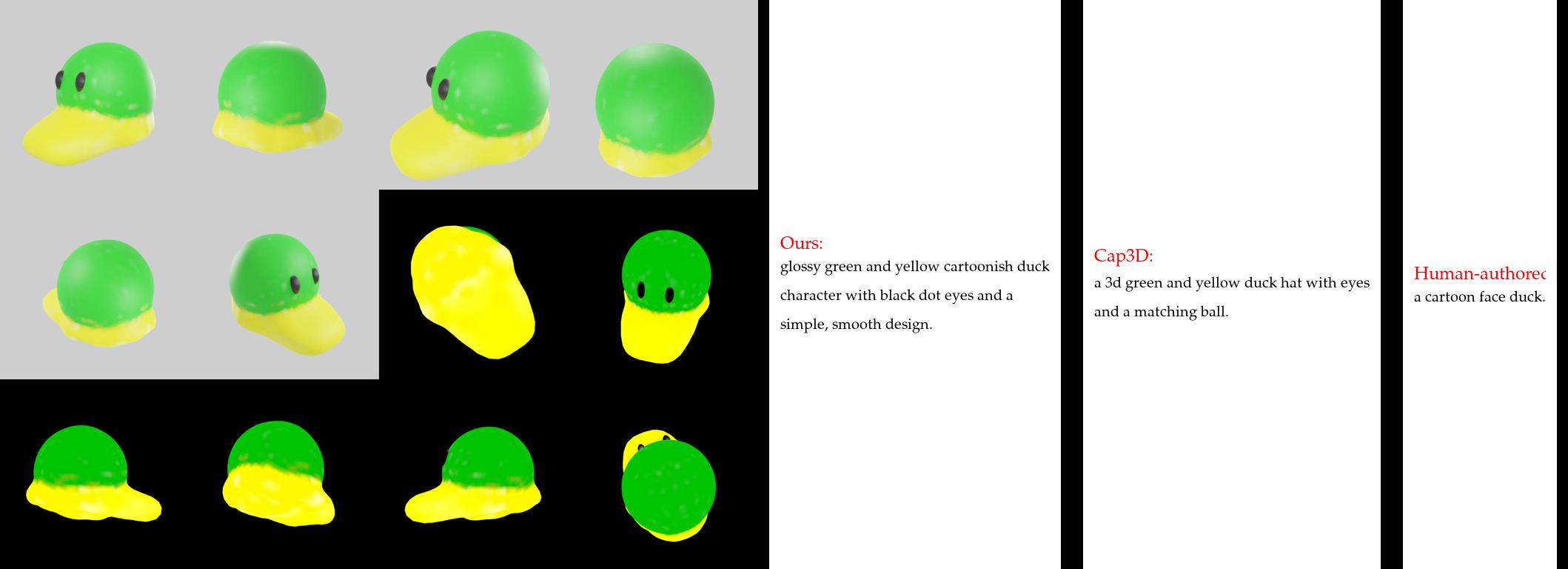}
    \caption{We compare captions through random sampling, including those generated by our method, by Cap3D, and those authored by humans.}
    \label{fig:appen_main_cmp_14}
\end{figure}

\begin{figure}
    \centering
    \includegraphics[width=\textwidth]{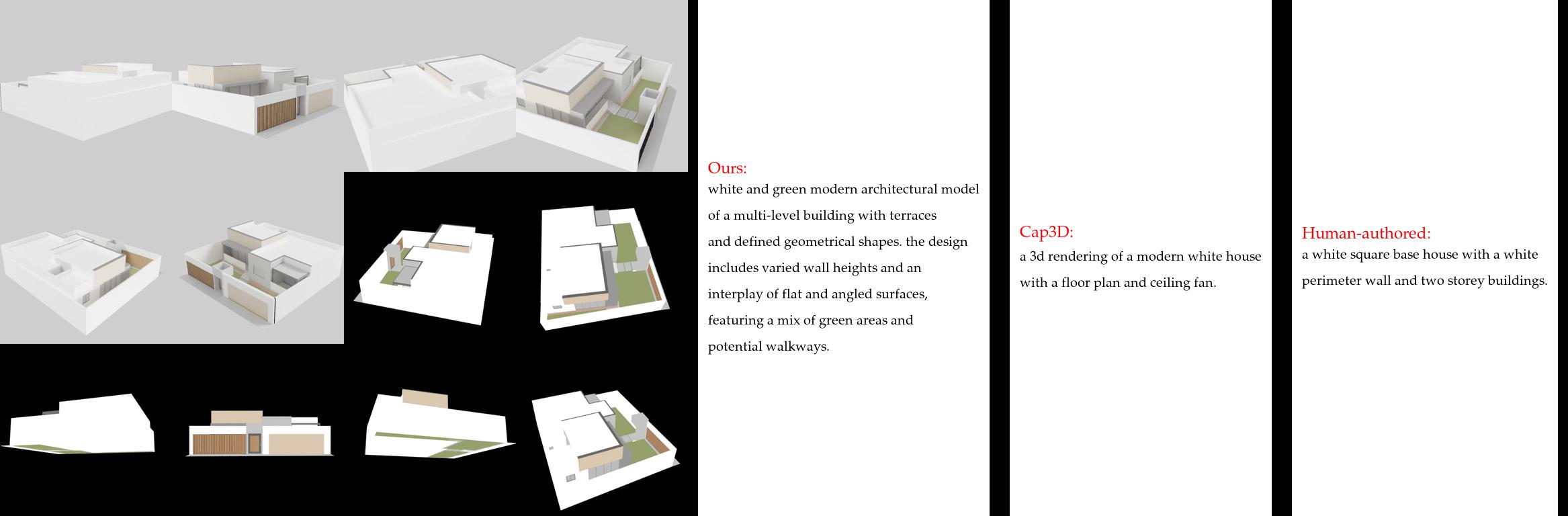}
    \caption{We compare captions through random sampling, including those generated by our method, by Cap3D, and those authored by humans.}
    \label{fig:appen_main_cmp_15}
\end{figure}

\begin{figure}
    \centering
    \includegraphics[width=\textwidth]{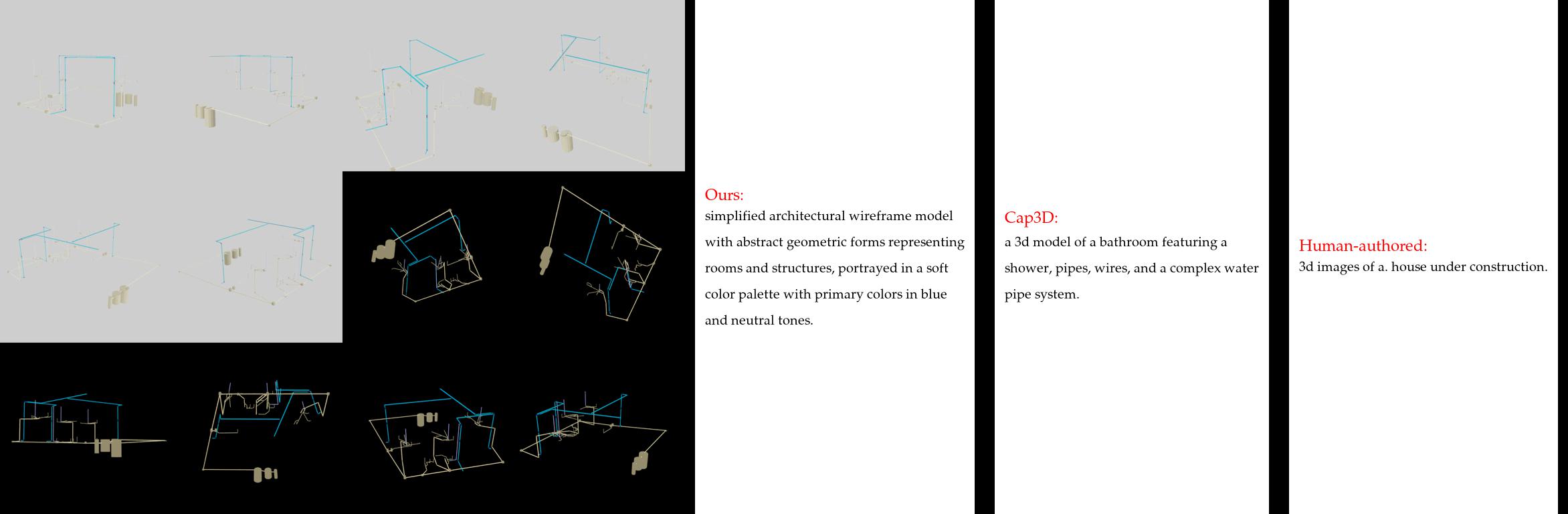}
    \caption{We compare captions through random sampling, including those generated by our method, by Cap3D, and those authored by humans.}
    \label{fig:appen_main_cmp_16}
\end{figure}

\begin{figure}
    \centering
    \includegraphics[width=\textwidth]{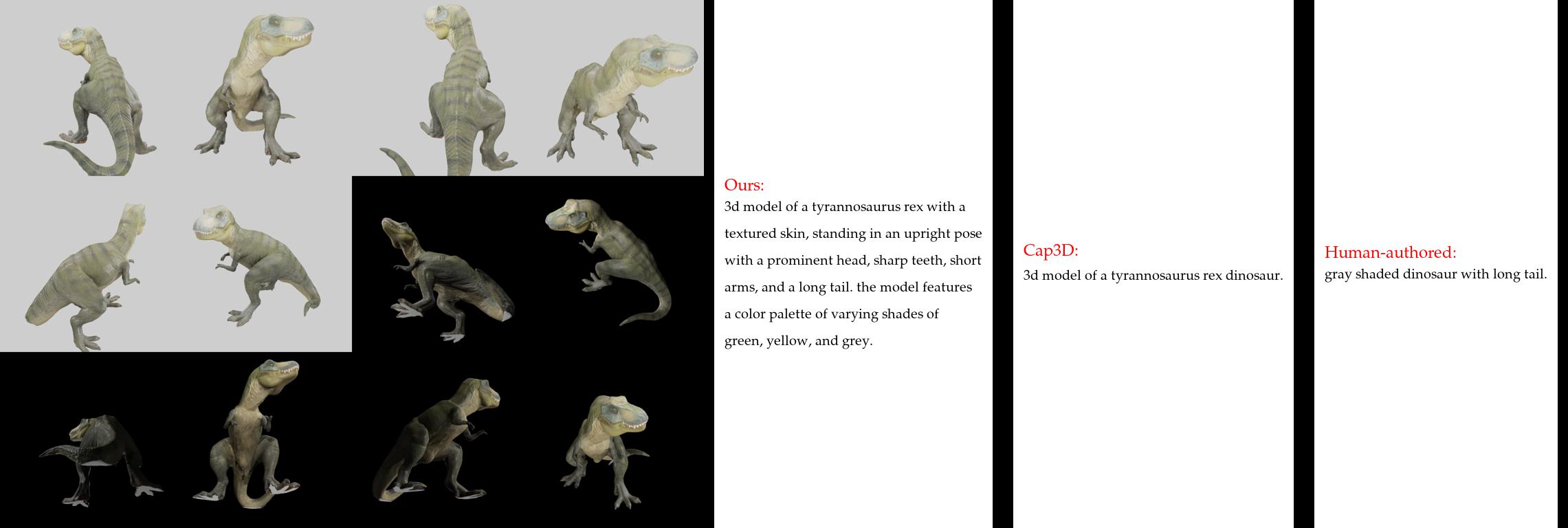}
    \caption{We compare captions through random sampling, including those generated by our method, by Cap3D, and those authored by humans.}
    \label{fig:appen_main_cmp_17}
\end{figure}

\begin{figure}
    \centering
    \includegraphics[width=\textwidth]{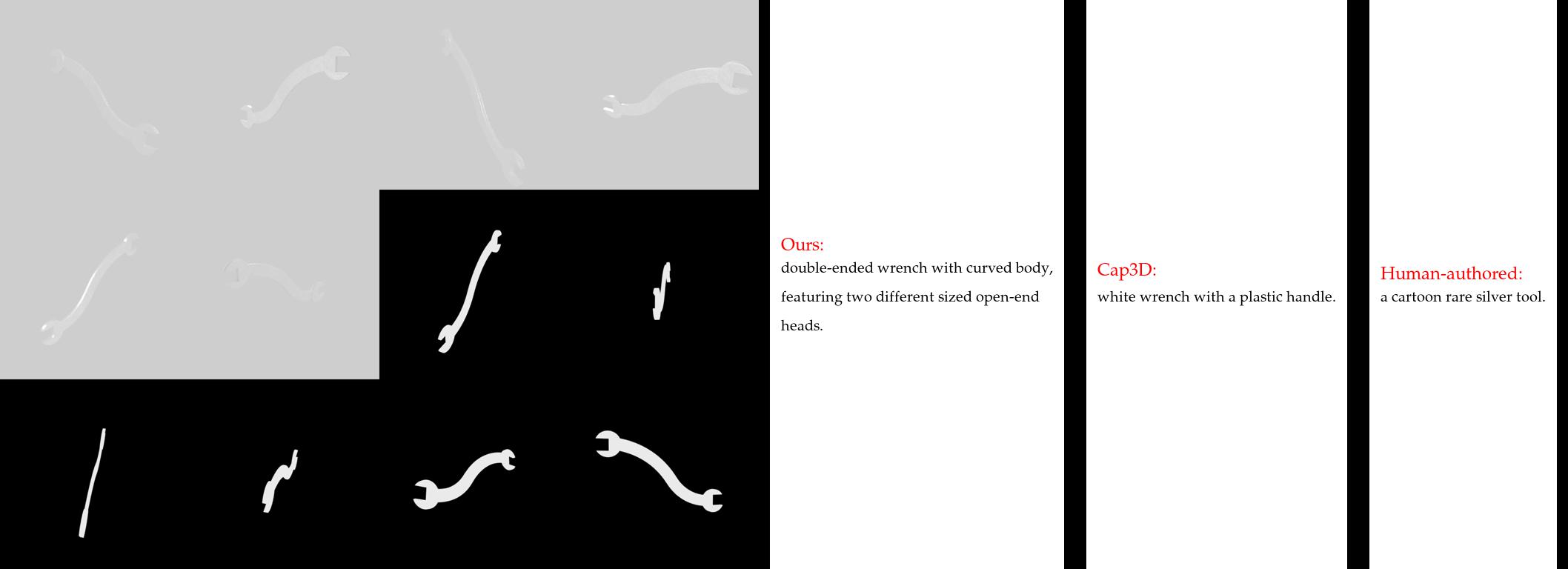}
    \caption{We compare captions through random sampling, including those generated by our method, by Cap3D, and those authored by humans.}
    \label{fig:appen_main_cmp_18}
\end{figure}

\begin{figure}
    \centering
    \includegraphics[width=\textwidth]{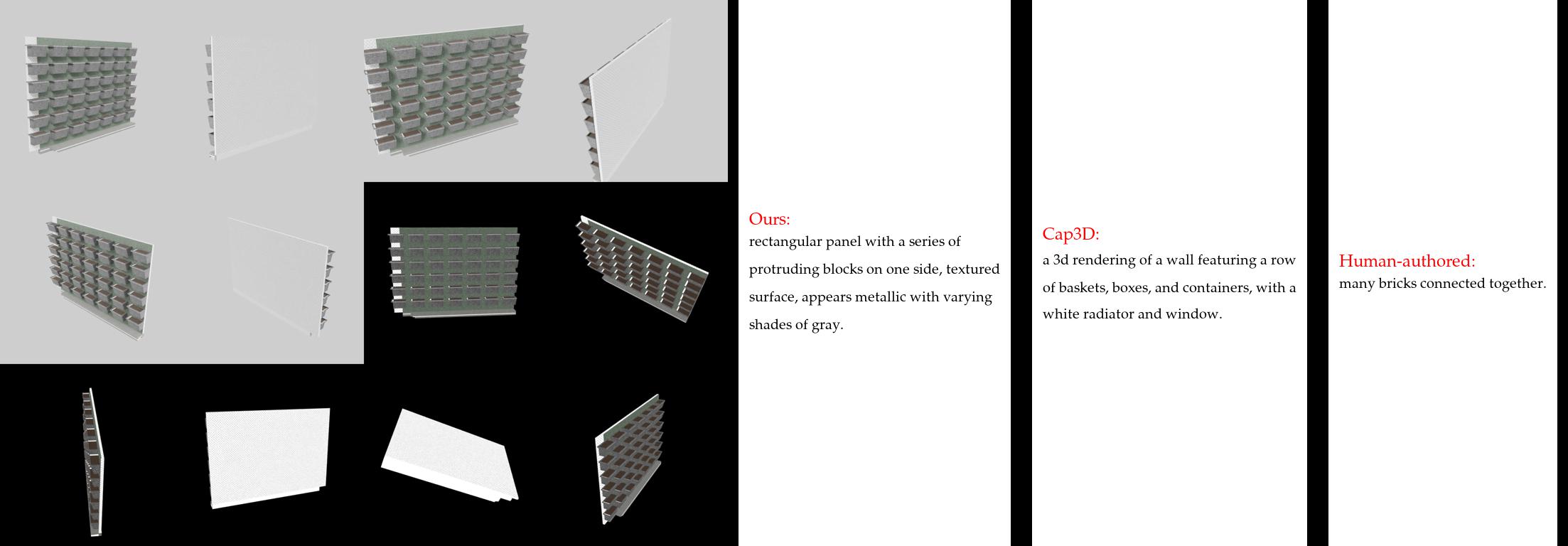}
    \caption{We compare captions through random sampling, including those generated by our method, by Cap3D, and those authored by humans.}
    \label{fig:appen_main_cmp_19}
\end{figure}

\begin{figure}
    \centering
    \includegraphics[width=\textwidth]{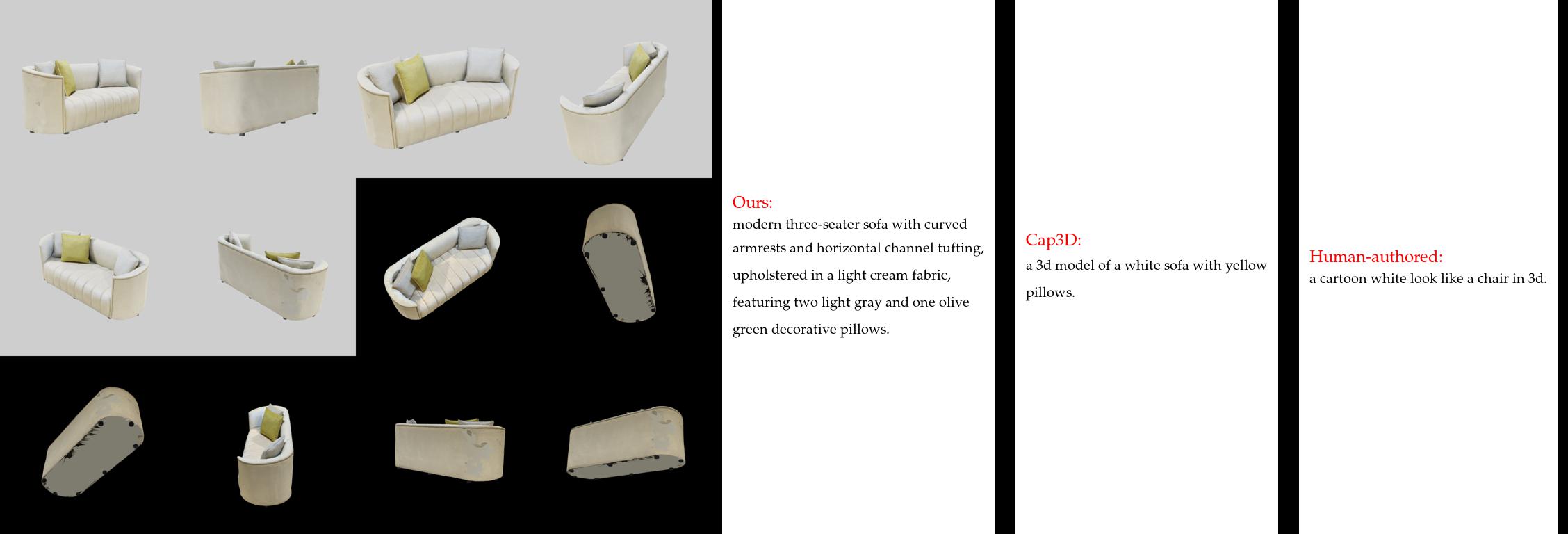}
    \caption{We compare captions through random sampling, including those generated by our method, by Cap3D, and those authored by humans.}
    \label{fig:appen_main_cmp_20}
\end{figure}

\begin{figure}
    \centering
    \includegraphics[width=\textwidth]{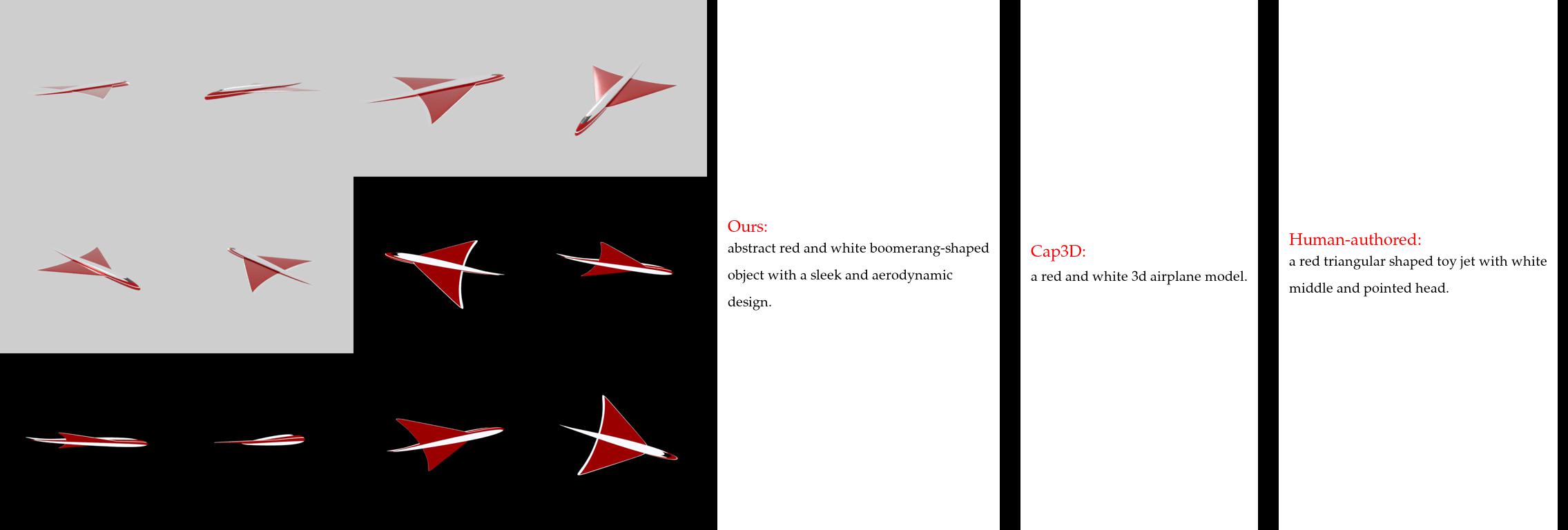}
    \caption{We compare captions through random sampling, including those generated by our method, by Cap3D, and those authored by humans.}
    \label{fig:appen_main_cmp_21}
\end{figure}

\begin{figure}
    \centering
    \includegraphics[width=\textwidth]{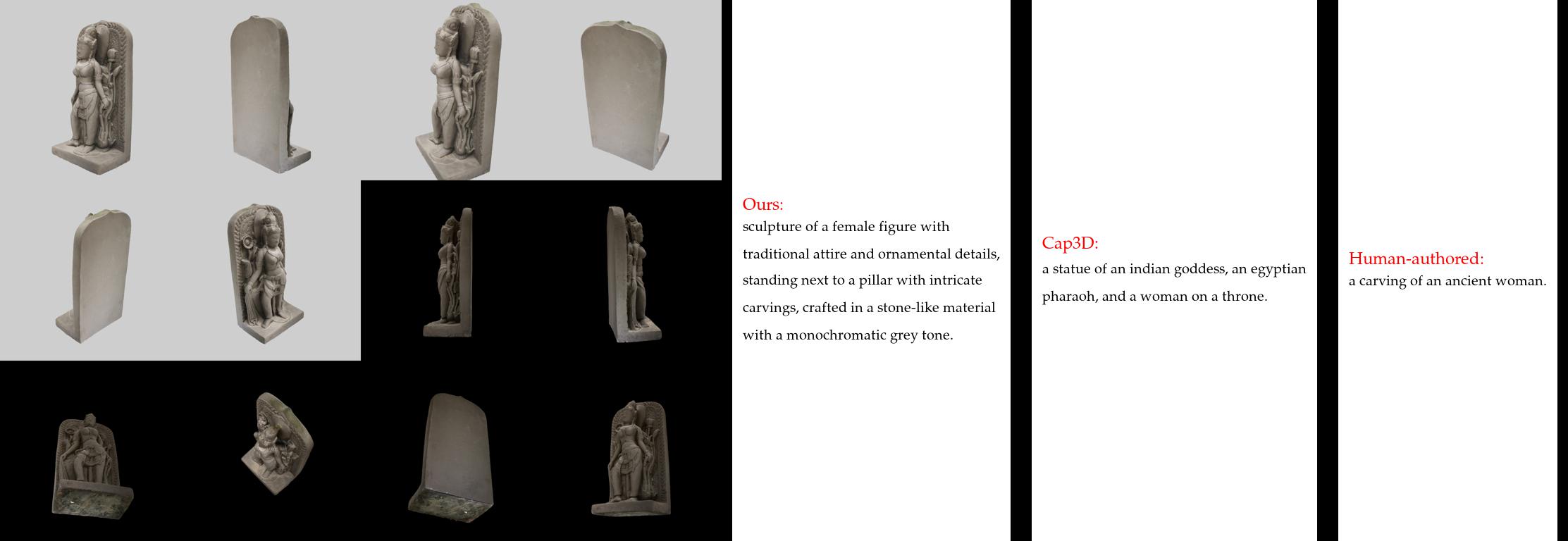}
    \caption{We compare captions through random sampling, including those generated by our method, by Cap3D, and those authored by humans.}
    \label{fig:appen_main_cmp_22}
\end{figure}

\begin{figure}
    \centering
    \includegraphics[width=\textwidth]{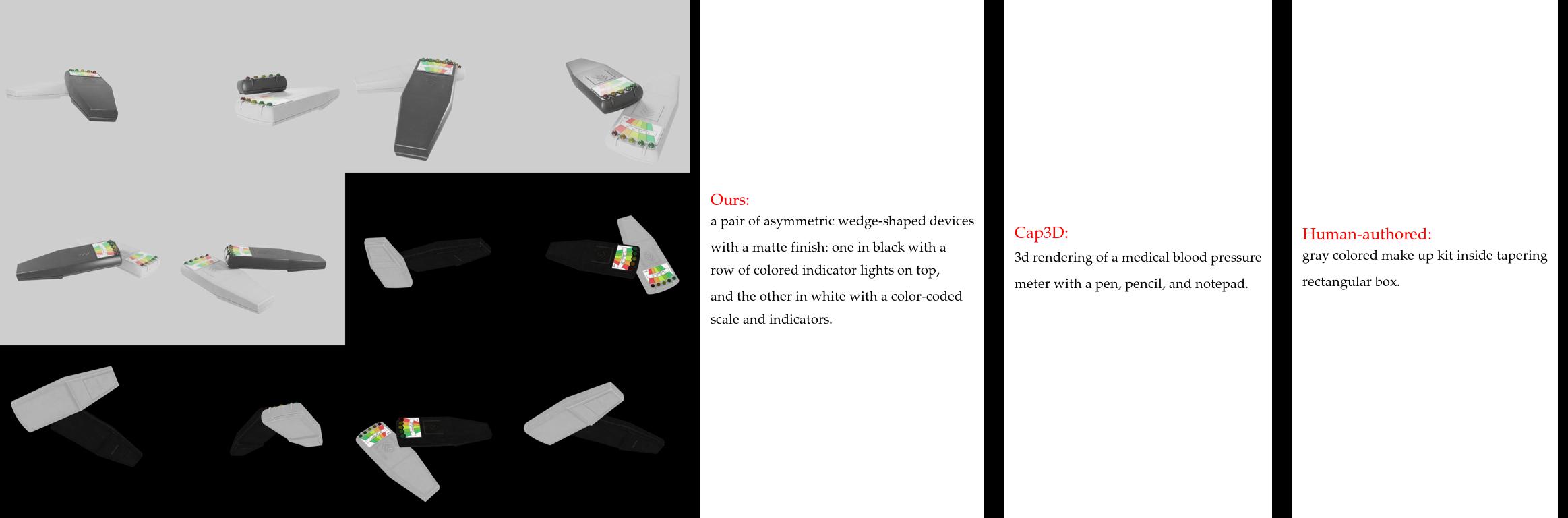}
    \caption{We compare captions through random sampling, including those generated by our method, by Cap3D, and those authored by humans.}
    \label{fig:appen_main_cmp_23}
\end{figure}

\begin{figure}
    \centering
    \includegraphics[width=\textwidth]{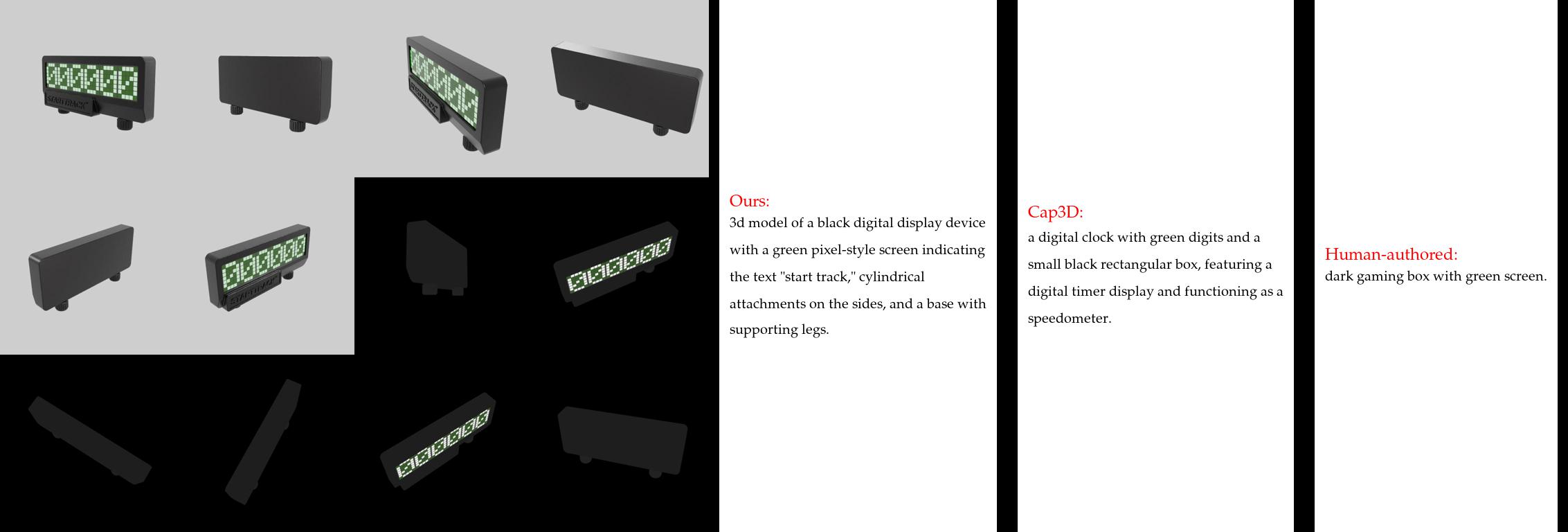}
    \caption{We compare captions through random sampling, including those generated by our method, by Cap3D, and those authored by humans.}
    \label{fig:appen_main_cmp_24}
\end{figure}

\begin{figure}
    \centering
    \includegraphics[width=\textwidth]{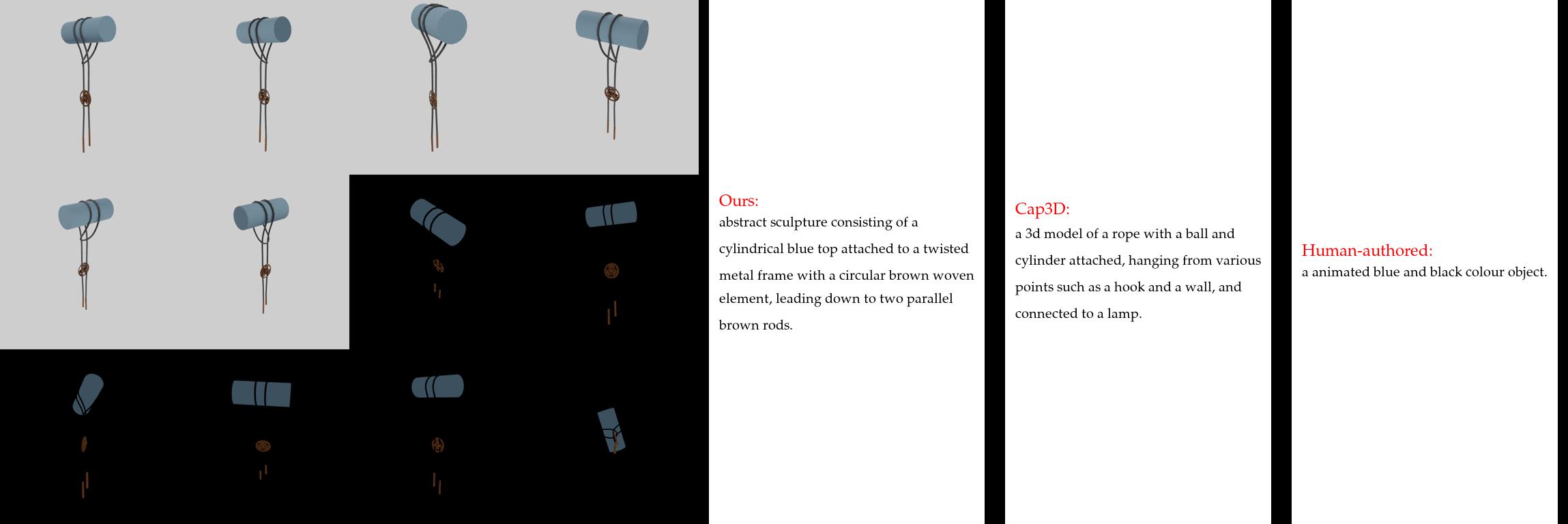}
    \caption{We compare captions through random sampling, including those generated by our method, by Cap3D, and those authored by humans.}
    \label{fig:appen_main_cmp_25}
\end{figure}

\begin{figure}
    \centering
    \includegraphics[width=\textwidth]{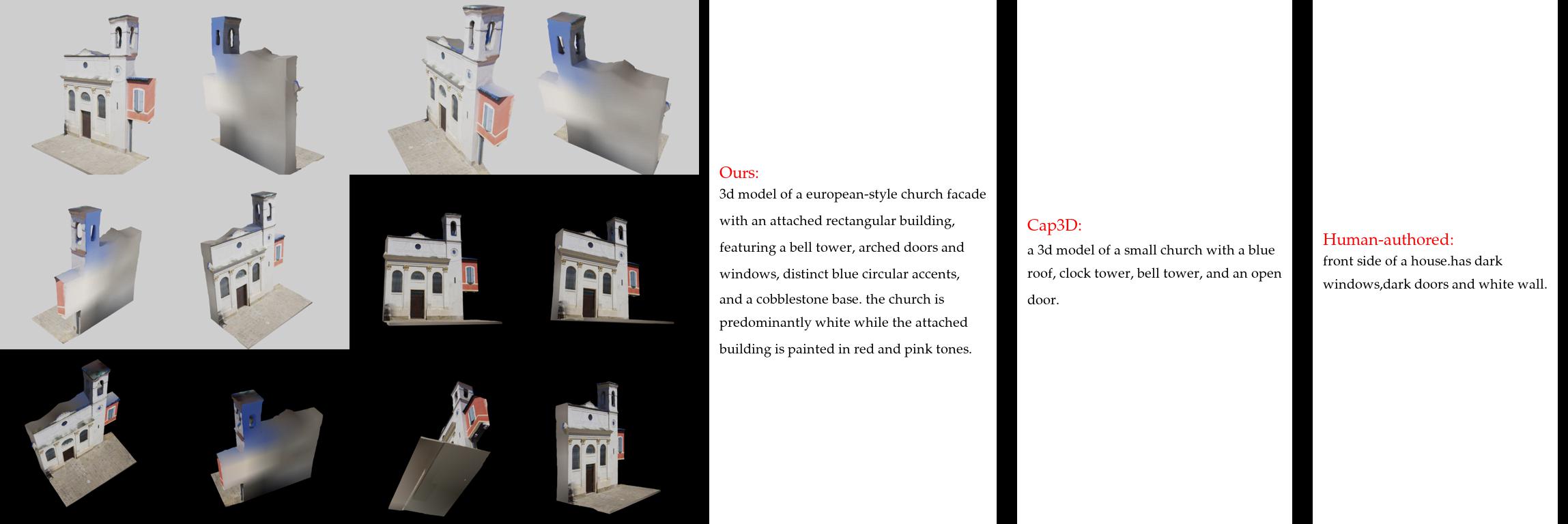}
    \caption{We compare captions through random sampling, including those generated by our method, by Cap3D, and those authored by humans.}
    \label{fig:appen_main_cmp_26}
\end{figure}

\begin{figure}
    \centering
    \includegraphics[width=\textwidth]{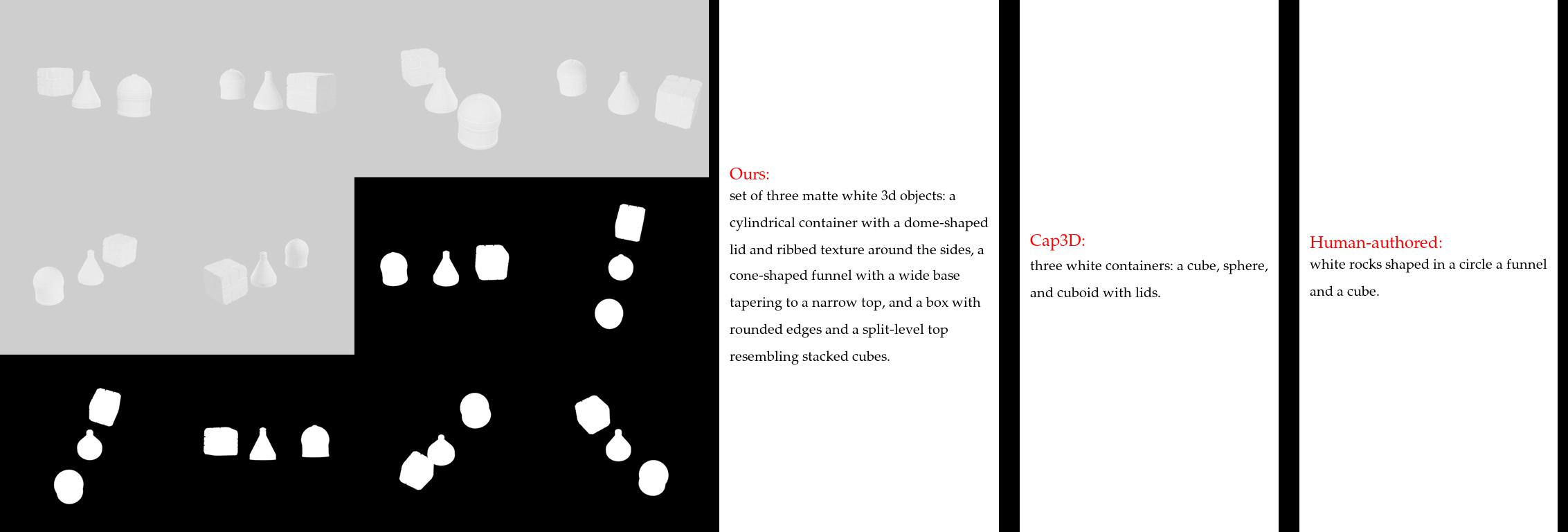}
    \caption{We compare captions through random sampling, including those generated by our method, by Cap3D, and those authored by humans.}
    \label{fig:appen_main_cmp_27}
\end{figure}

\begin{figure}
    \centering
    \includegraphics[width=\textwidth]{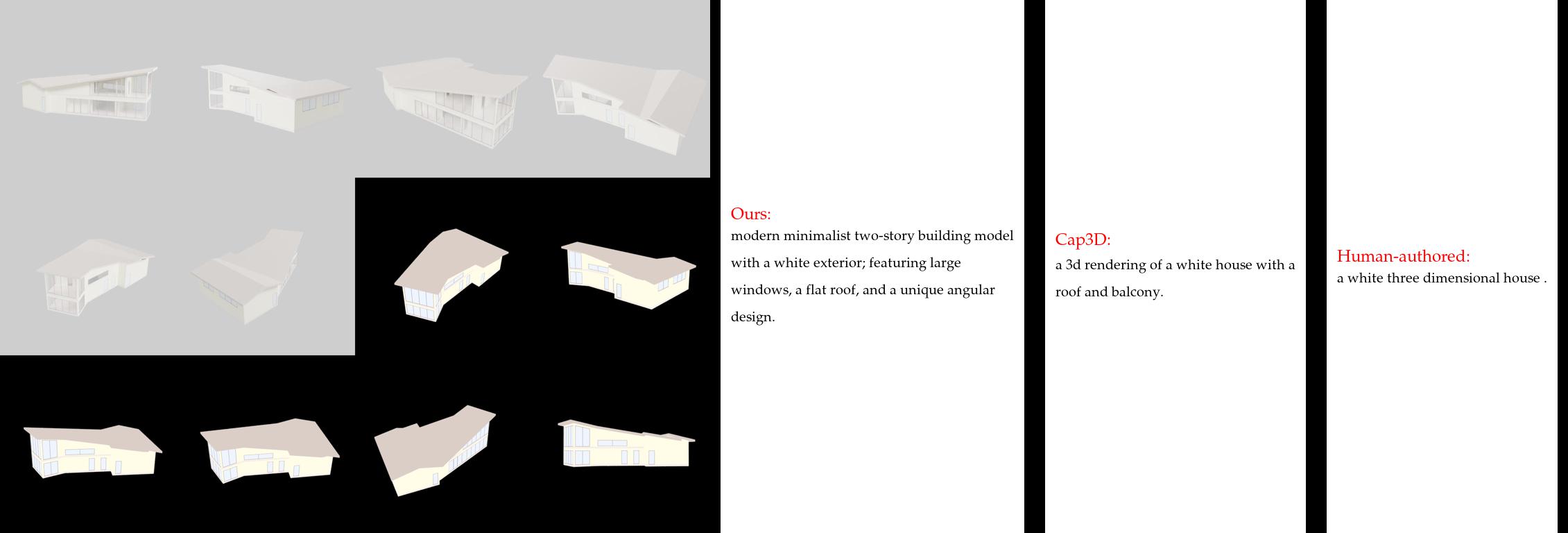}
    \caption{We compare captions through random sampling, including those generated by our method, by Cap3D, and those authored by humans.}
    \label{fig:appen_main_cmp_28}
\end{figure}

\begin{figure}
    \centering
    \includegraphics[width=\textwidth]{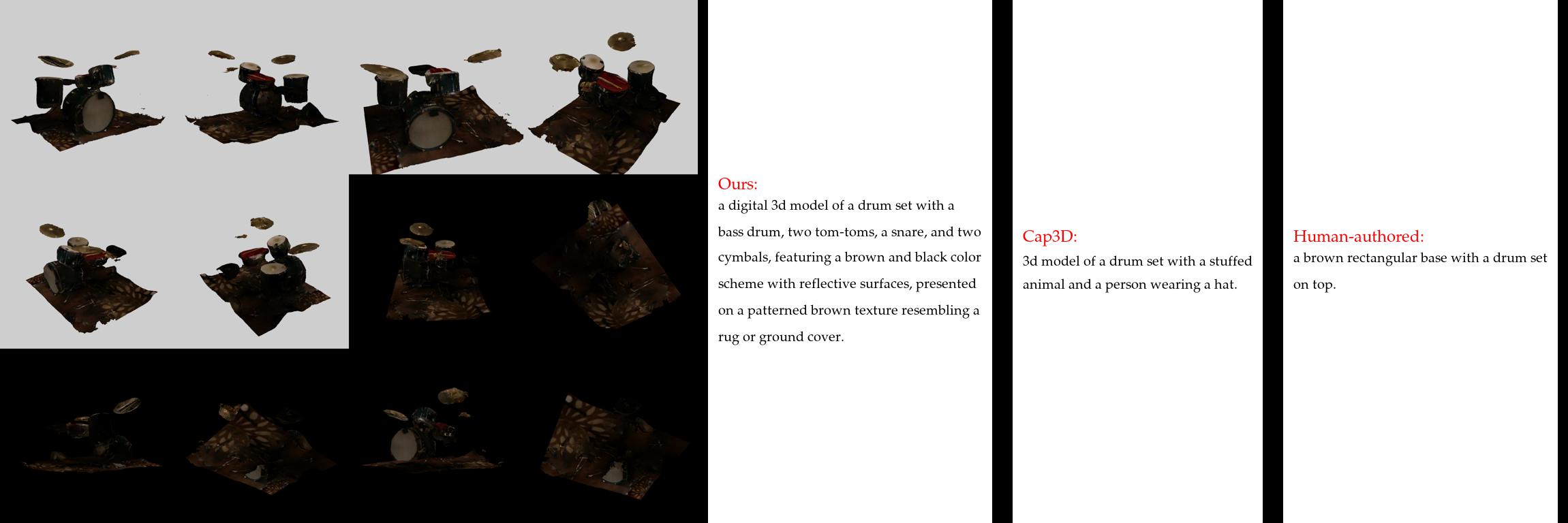}
    \caption{We compare captions through random sampling, including those generated by our method, by Cap3D, and those authored by humans.}
    \label{fig:appen_main_cmp_29}
\end{figure}

\begin{figure}
    \centering
    \includegraphics[width=\textwidth]{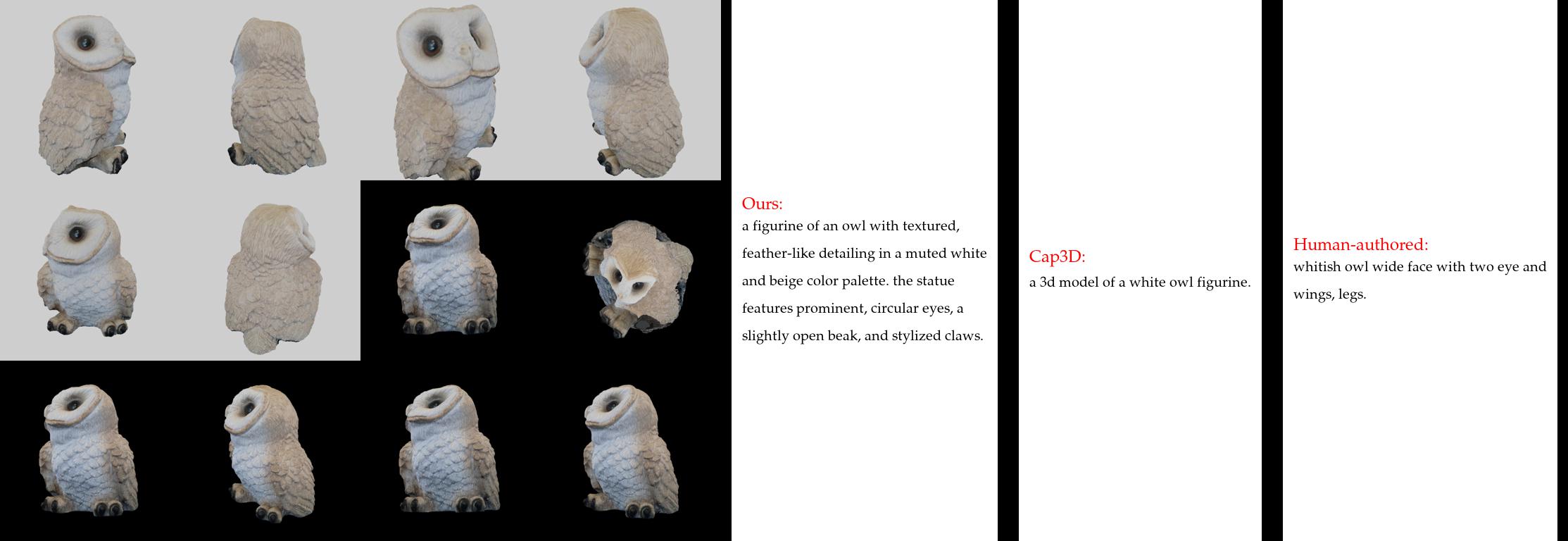}
    \caption{We compare captions through random sampling, including those generated by our method, by Cap3D, and those authored by humans.}
    \label{fig:appen_main_cmp_30}
\end{figure}

\begin{figure}
    \centering
    \includegraphics[width=\textwidth]{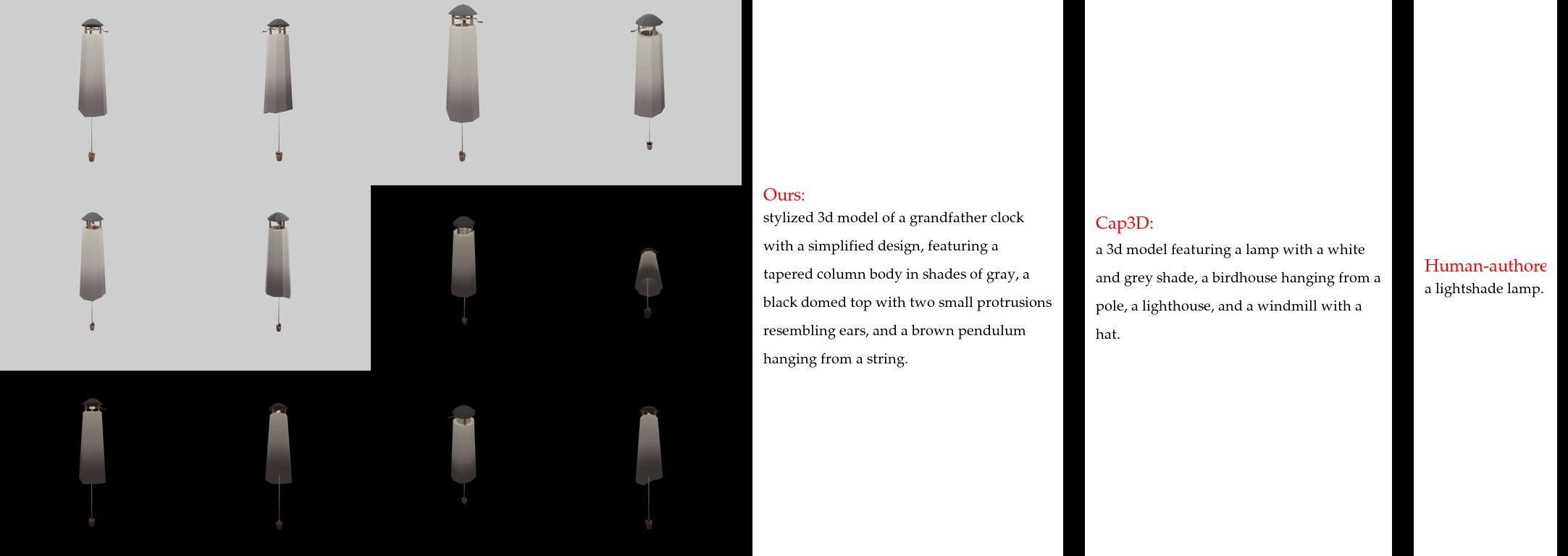}
    \caption{We compare captions through random sampling, including those generated by our method, by Cap3D, and those authored by humans.}
    \label{fig:appen_main_cmp_31}
\end{figure}

\begin{figure}
    \centering
    \includegraphics[width=\textwidth]{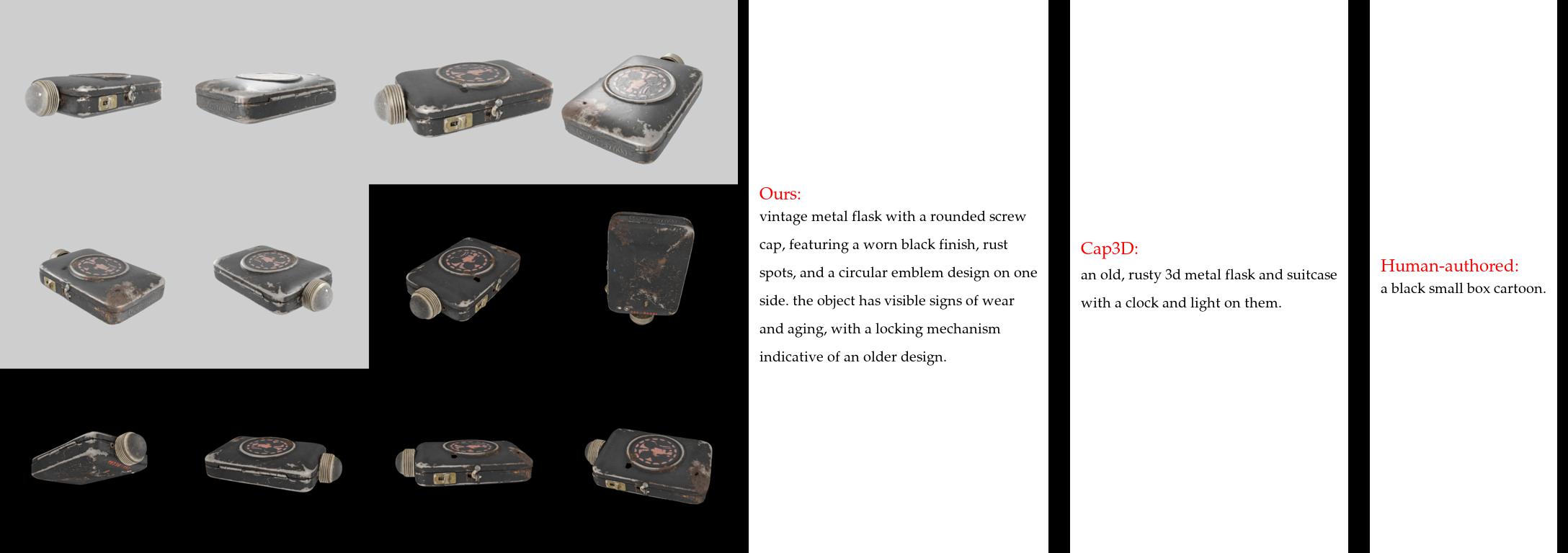}
    \caption{We compare captions through random sampling, including those generated by our method, by Cap3D, and those authored by humans.}
    \label{fig:appen_main_cmp_32}
\end{figure}

\clearpage
\hypertarget{B4}{\subsection{Captions: Ours vs. ablated variants}}
\label{appen:dataset:ablation}

We list several qualitative comparisons here to demonstrate the effectiveness of our method compared to (1) \textbf{Bottom 6-views}, we employ the 6 renderings identified as having the lowest alignment scores, as determined by our DiffuRank algorithm (refer to Alg.~\ref{alg:diffurank}); (2) \textbf{Allviews 28-views}, which involves utilizing all 28 rendered views as inputs for the GPT4-Vision; and (3) \textbf{Horizontal 6-views}, this configuration involves selecting 6 rendered views that position the camera horizontally relative to the object's default orientation, adhering to the same vertical positioning guidelines used by Cap3D. Results generally show the captions generated by our method (i.e., Top-6) contain more accurate, detailed, and less hallucinated information. 

\begin{figure}[h]
    \centering
    \includegraphics[width=\textwidth]{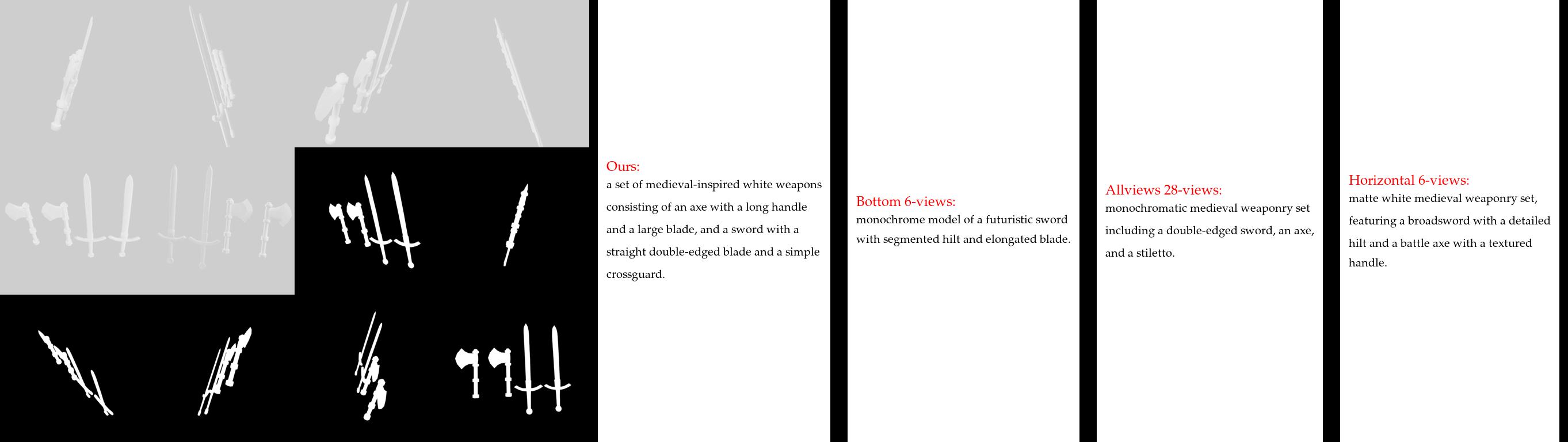}
    \caption{We evaluate captions by randomly sampling and comparing them across different methods: our approach (Top 6-views), using the bottom 6-views, utilizing all 28-views, and employing horizontal 6-views.}
    \label{fig:appen_ablate_cmp_01}
\end{figure}

\begin{figure}[h]
    \centering
    \includegraphics[width=\textwidth]{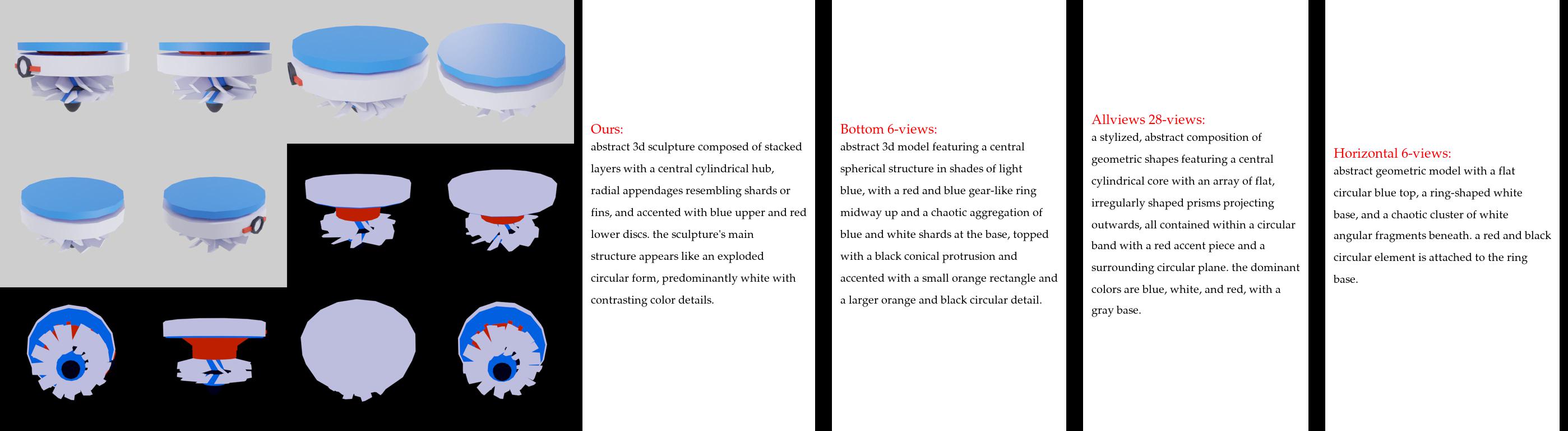}
    \caption{We evaluate captions by randomly sampling and comparing them across different methods: our approach (Top 6-views), using the bottom 6-views, utilizing all 28-views, and employing horizontal 6-views.}
    \label{fig:appen_ablate_cmp_02}
\end{figure}

\begin{figure}[h]
    \centering
    \includegraphics[width=\textwidth]{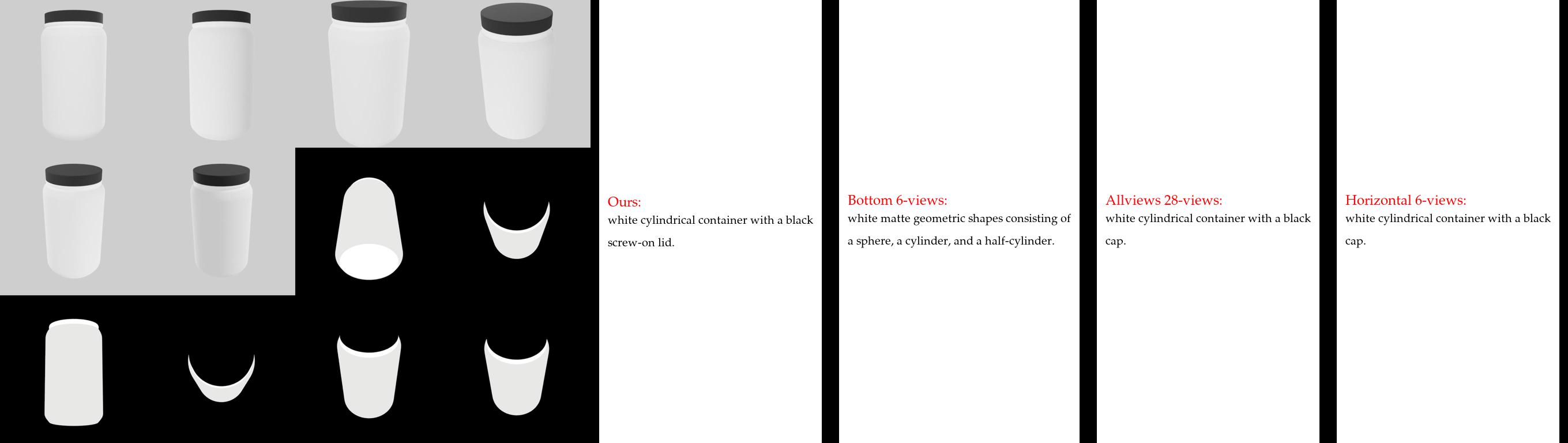}
    \caption{We evaluate captions by randomly sampling and comparing them across different methods: our approach (Top 6-views), using the bottom 6-views, utilizing all 28-views, and employing horizontal 6-views.}
    \label{fig:appen_ablate_cmp_03}
\end{figure}

\begin{figure}
    \centering
    \includegraphics[width=\textwidth]{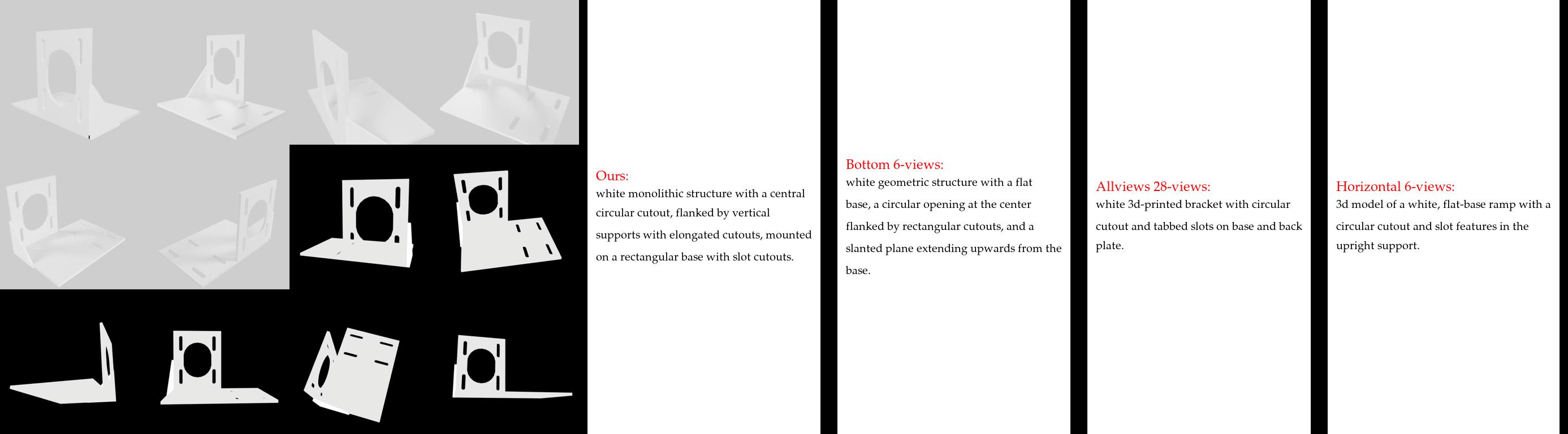}
    \caption{We evaluate captions by randomly sampling and comparing them across different methods: our approach (Top 6-views), using the bottom 6-views, utilizing all 28-views, and employing horizontal 6-views.}
    \label{fig:appen_ablate_cmp_04}
\end{figure}

\begin{figure}
    \centering
    \includegraphics[width=\textwidth]{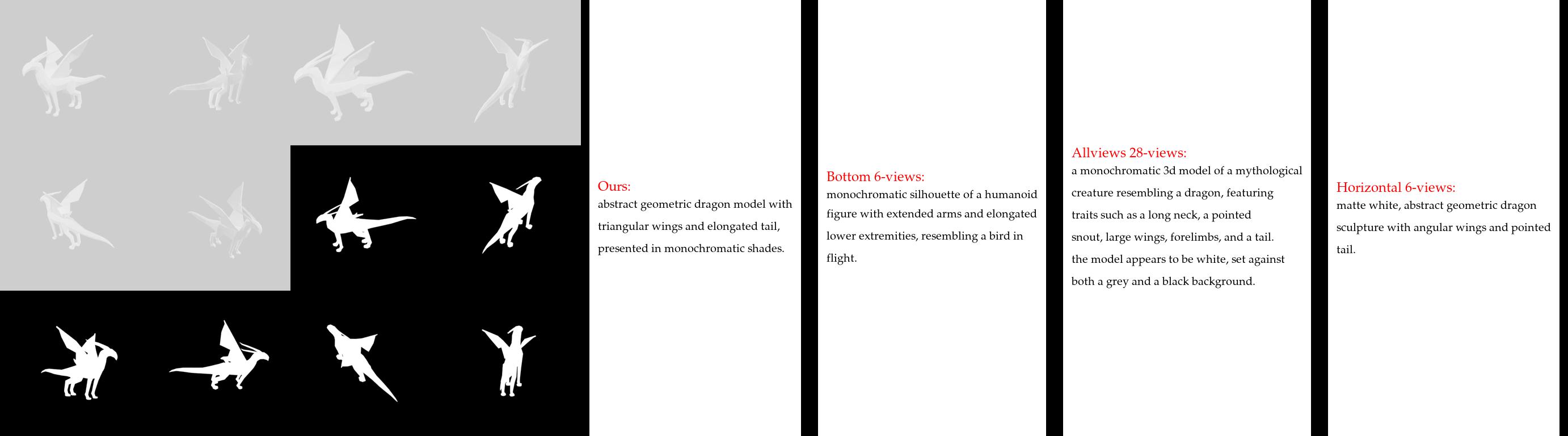}
    \caption{We evaluate captions by randomly sampling and comparing them across different methods: our approach (Top 6-views), using the bottom 6-views, utilizing all 28-views, and employing horizontal 6-views.}
    \label{fig:appen_ablate_cmp_05}
\end{figure}

\begin{figure}
    \centering
    \includegraphics[width=\textwidth]{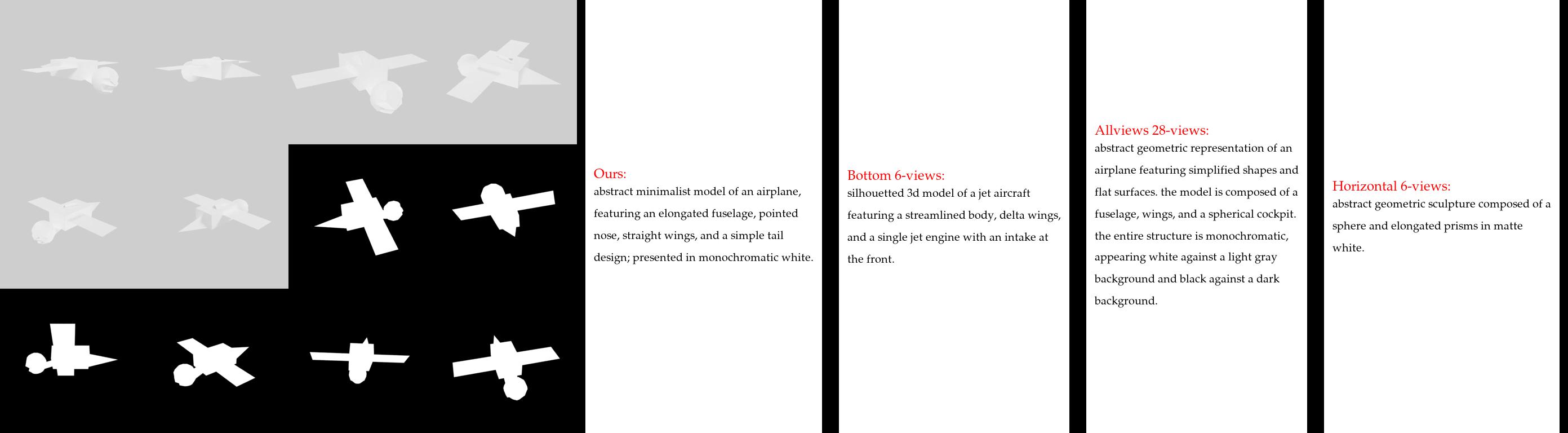}
    \caption{We evaluate captions by randomly sampling and comparing them across different methods: our approach (Top 6-views), using the bottom 6-views, utilizing all 28-views, and employing horizontal 6-views.}
    \label{fig:appen_ablate_cmp_06}
\end{figure}

\begin{figure}
    \centering
    \includegraphics[width=\textwidth]{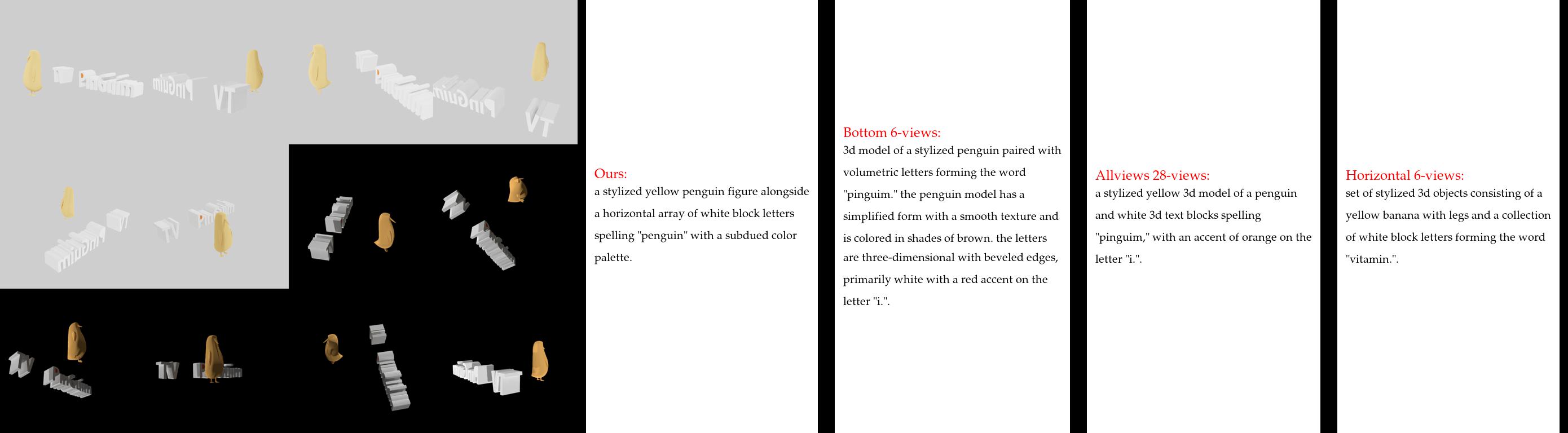}
    \caption{We evaluate captions by randomly sampling and comparing them across different methods: our approach (Top 6-views), using the bottom 6-views, utilizing all 28-views, and employing horizontal 6-views.}
    \label{fig:appen_ablate_cmp_07}
\end{figure}

\begin{figure}
    \centering
    \includegraphics[width=\textwidth]{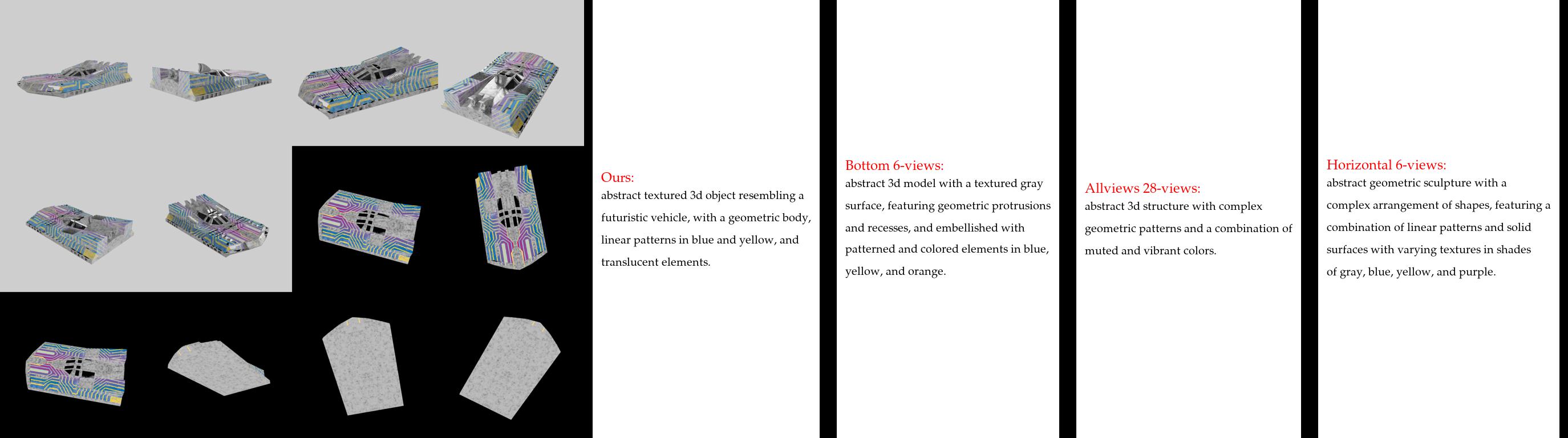}
    \caption{We evaluate captions by randomly sampling and comparing them across different methods: our approach (Top 6-views), using the bottom 6-views, utilizing all 28-views, and employing horizontal 6-views.}
    \label{fig:appen_ablate_cmp_08}
\end{figure}

\begin{figure}
    \centering
    \includegraphics[width=\textwidth]{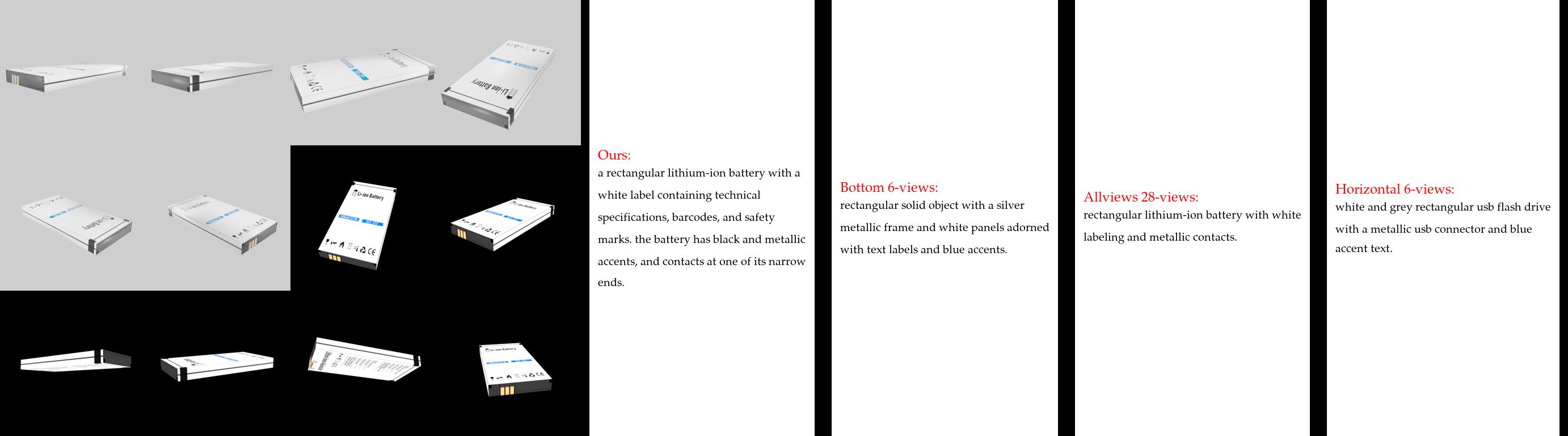}
    \caption{We evaluate captions by randomly sampling and comparing them across different methods: our approach (Top 6-views), using the bottom 6-views, utilizing all 28-views, and employing horizontal 6-views.}
    \label{fig:appen_ablate_cmp_09}
\end{figure}

\begin{figure}
    \centering
    \includegraphics[width=\textwidth]{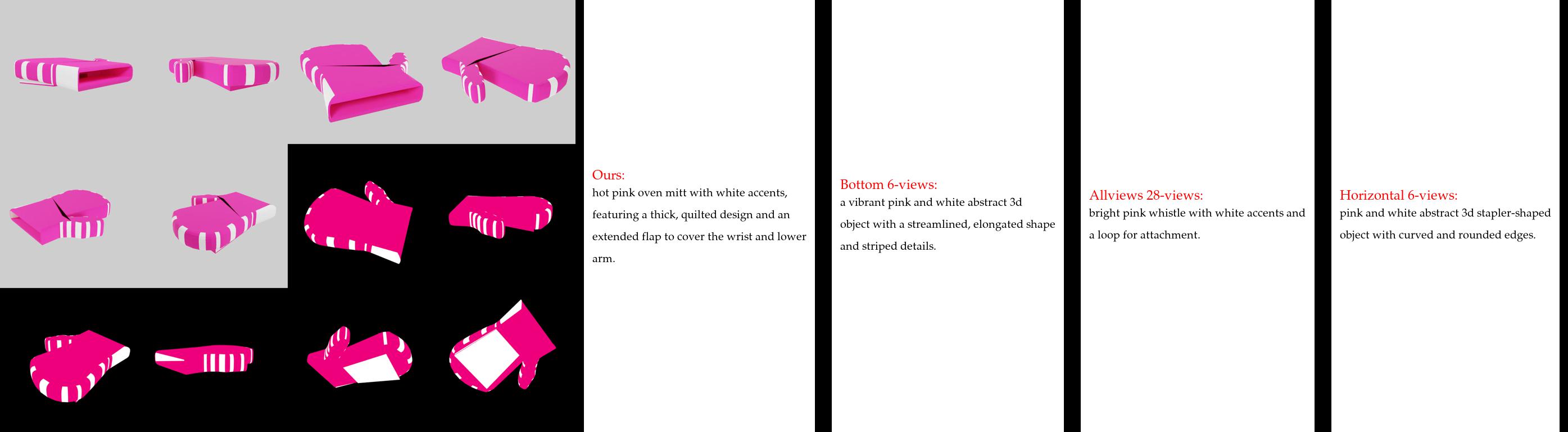}
    \caption{We evaluate captions by randomly sampling and comparing them across different methods: our approach (Top 6-views), using the bottom 6-views, utilizing all 28-views, and employing horizontal 6-views.}
    \label{fig:appen_ablate_cmp_10}
\end{figure}

\begin{figure}
    \centering
    \includegraphics[width=\textwidth]{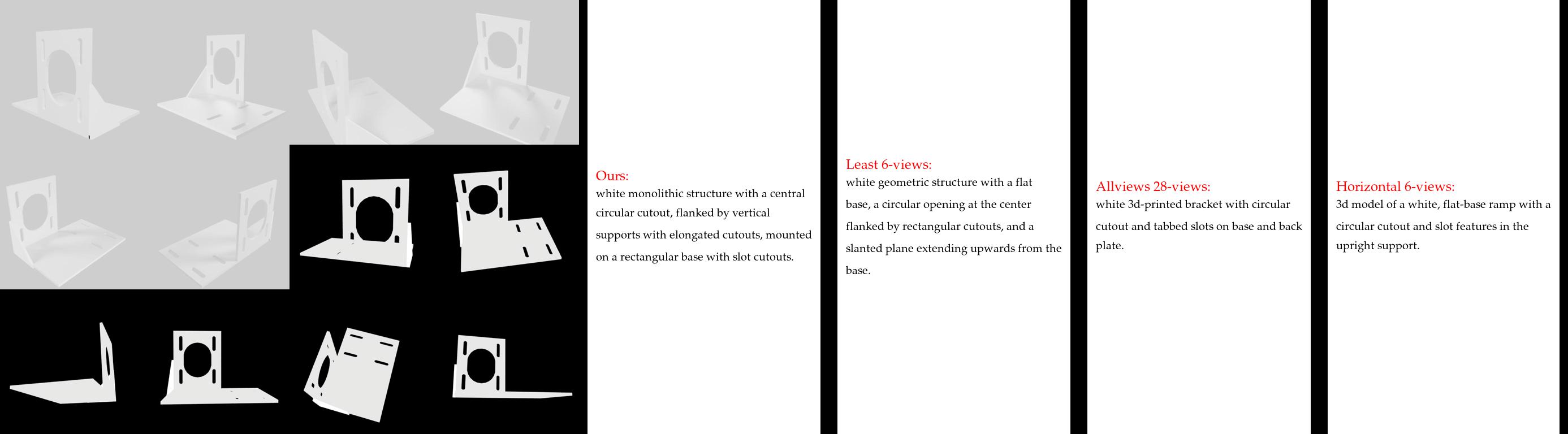}
    \caption{We evaluate captions by randomly sampling and comparing them across different methods: our approach (Top 6-views), using the bottom 6-views, utilizing all 28-views, and employing horizontal 6-views.}
    \label{fig:appen_ablate_cmp_11}
\end{figure}

\begin{figure}[h]
    \centering
    \includegraphics[width=0.85\textwidth]{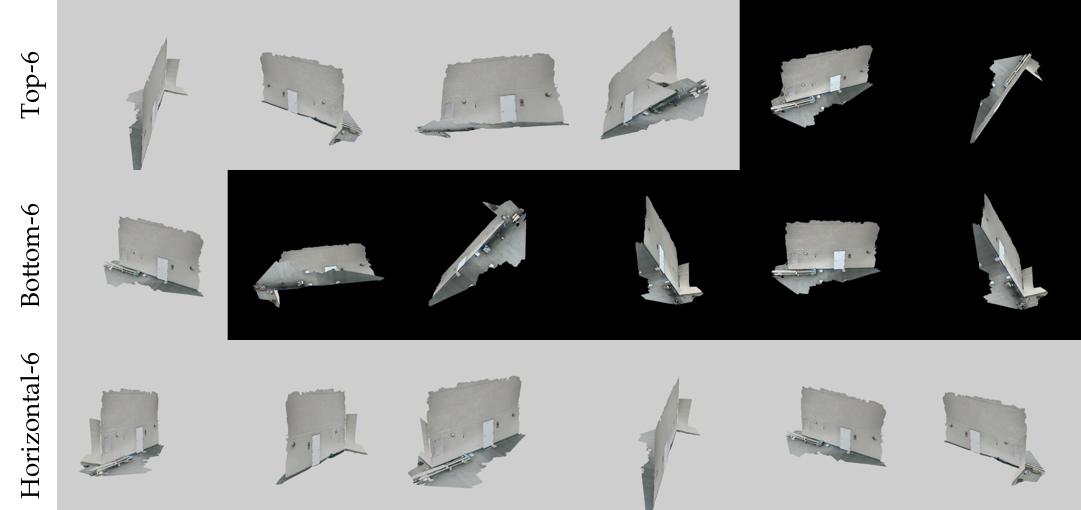}
    \caption{Randomly sampled DiffuRank comparisons. \textbf{Top-row}: Top 6-views selected by DiffuRank; \textbf{Middle-row}: Bottom 6-views selected by DiffuRank; \textbf{Bottom-row}: Horizontal 6-views. }
    \label{fig:appen_diffu_01}
\end{figure}

\begin{figure}[h]
    \centering
    \includegraphics[width=0.85\textwidth]{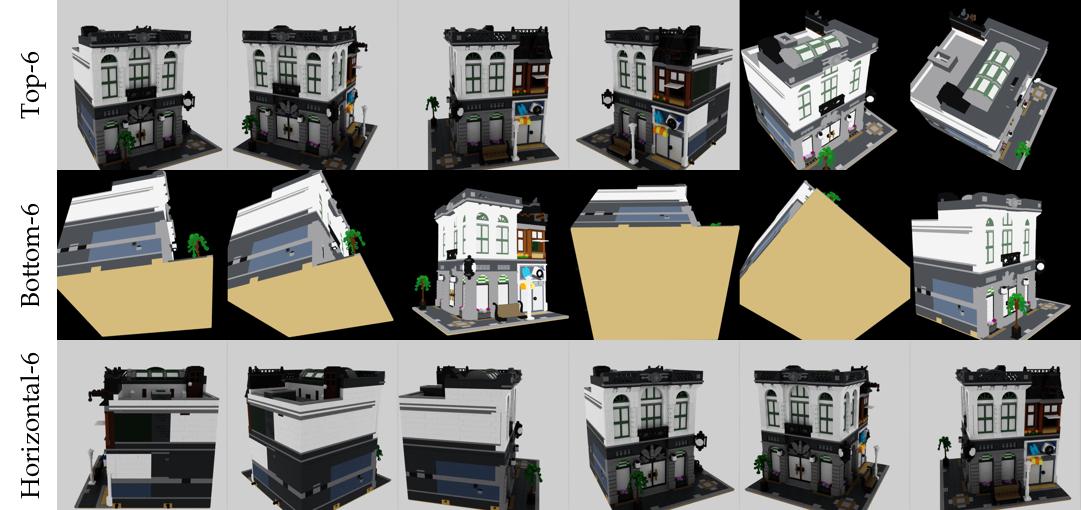}
    \caption{Randomly sampled DiffuRank comparisons. \textbf{Top-row}: Top 6-views selected by DiffuRank; \textbf{Middle-row}: Bottom 6-views selected by DiffuRank; \textbf{Bottom-row}: Horizontal 6-views. }
    \label{fig:appen_diffu_02}
\end{figure}

\begin{figure}
    \centering
    \includegraphics[width=\textwidth]{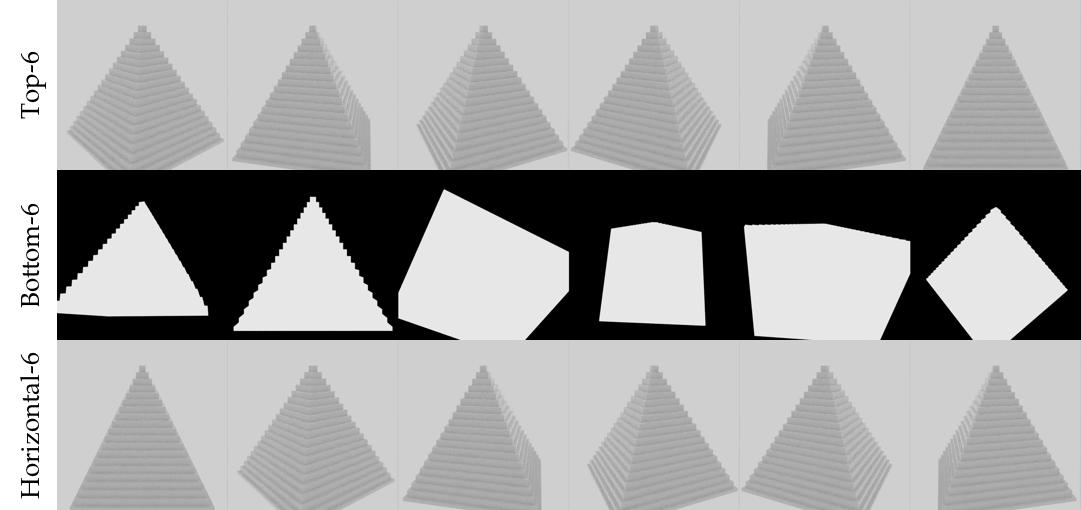}
    \caption{Randomly sampled DiffuRank comparisons. \textbf{Top-row}: Top 6-views selected by DiffuRank; \textbf{Middle-row}: Bottom 6-views selected by DiffuRank; \textbf{Bottom-row}: Horizontal 6-views. }
    \label{fig:appen_diffu_04}
\end{figure}

\begin{figure}
    \centering
    \includegraphics[width=\textwidth]{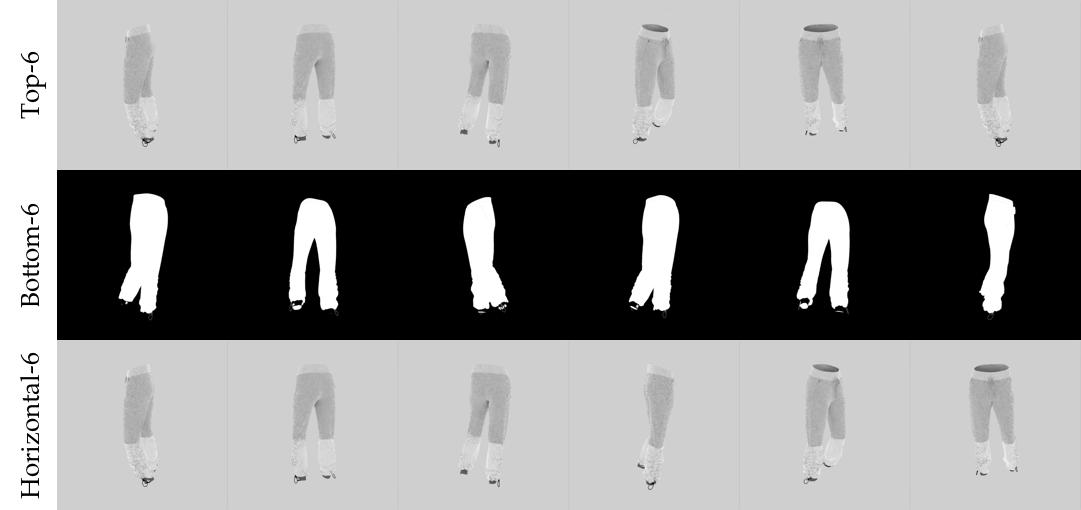}
    \caption{Randomly sampled DiffuRank comparisons. \textbf{Top-row}: Top 6-views selected by DiffuRank; \textbf{Middle-row}: Bottom 6-views selected by DiffuRank; \textbf{Bottom-row}: Horizontal 6-views. }
    \label{fig:appen_diffu_05}
\end{figure}

\begin{figure}
    \centering
    \includegraphics[width=\textwidth]{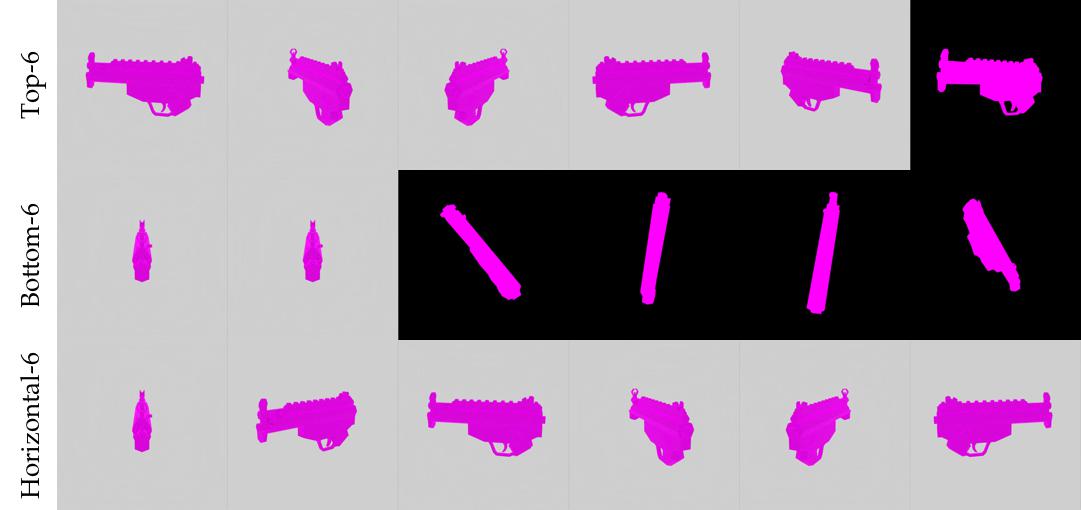}
    \caption{Randomly sampled DiffuRank comparisons. \textbf{Top-row}: Top 6-views selected by DiffuRank; \textbf{Middle-row}: Bottom 6-views selected by DiffuRank; \textbf{Bottom-row}: Horizontal 6-views. }
    \label{fig:appen_diffu_06}
\end{figure}

\begin{figure}
    \centering
    \includegraphics[width=\textwidth]{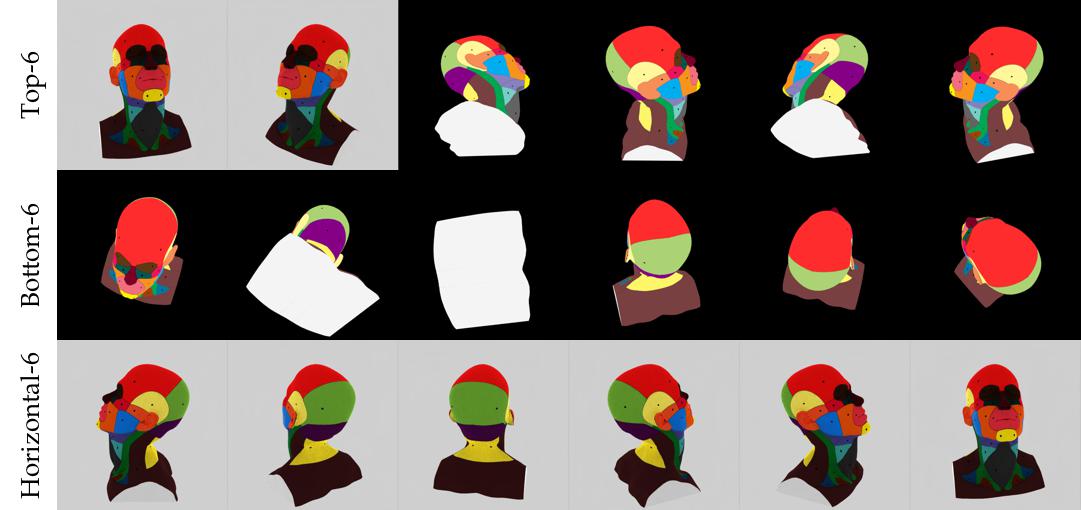}
    \caption{Randomly sampled DiffuRank comparisons. \textbf{Top-row}: Top 6-views selected by DiffuRank; \textbf{Middle-row}: Bottom 6-views selected by DiffuRank; \textbf{Bottom-row}: Horizontal 6-views. }
    \label{fig:appen_diffu_08}
\end{figure}

\begin{figure}
    \centering
    \includegraphics[width=\textwidth]{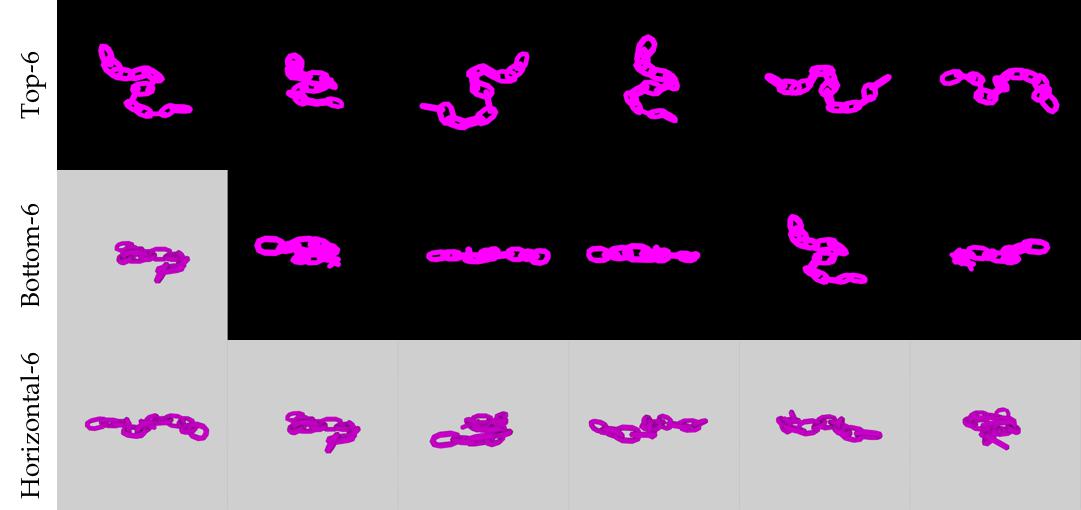}
    \caption{Randomly sampled DiffuRank comparisons. \textbf{Top-row}: Top 6-views selected by DiffuRank; \textbf{Middle-row}: Bottom 6-views selected by DiffuRank; \textbf{Bottom-row}: Horizontal 6-views. }
    \label{fig:appen_diffu_10}
\end{figure}

\begin{figure}
    \centering
    \includegraphics[width=\textwidth]{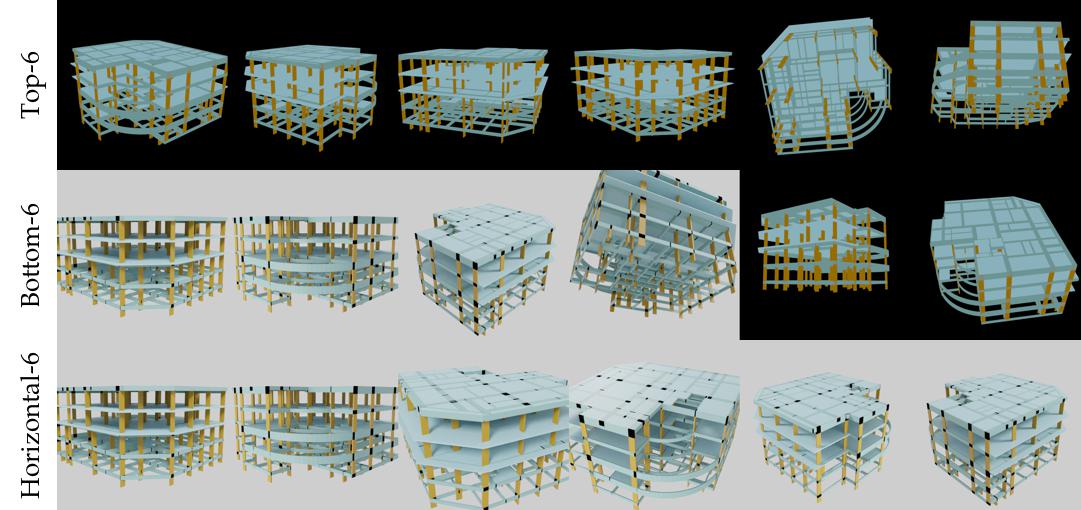}
    \caption{Randomly sampled DiffuRank comparisons. \textbf{Top-row}: Top 6-views selected by DiffuRank; \textbf{Middle-row}: Bottom 6-views selected by DiffuRank; \textbf{Bottom-row}: Horizontal 6-views. }
    \label{fig:appen_diffu_12}
\end{figure}

\begin{figure}
    \centering
    \includegraphics[width=\textwidth]{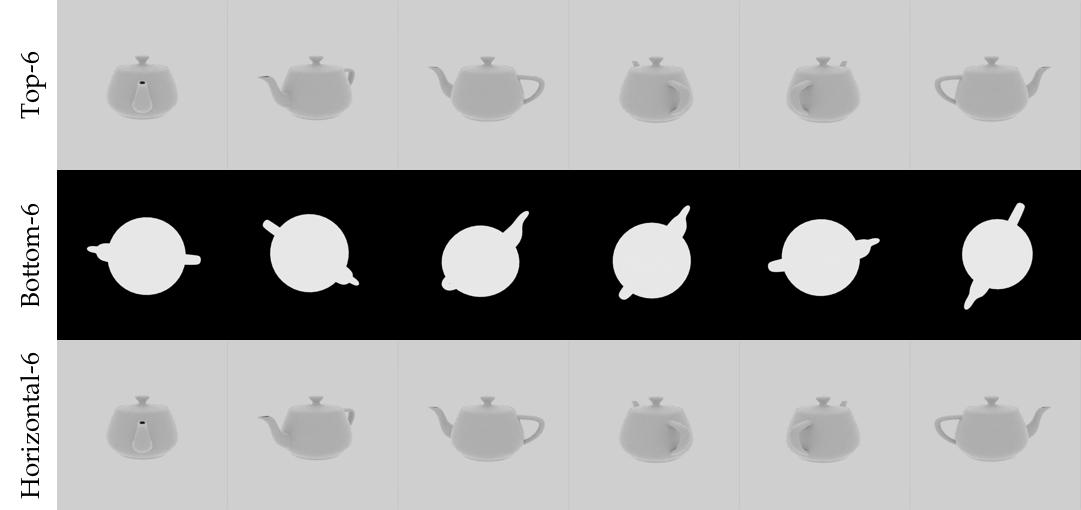}
    \caption{Randomly sampled DiffuRank comparisons. \textbf{Top-row}: Top 6-views selected by DiffuRank; \textbf{Middle-row}: Bottom 6-views selected by DiffuRank; \textbf{Bottom-row}: Horizontal 6-views. }
    \label{fig:appen_diffu_13}
\end{figure}

\begin{figure}
    \centering
    \includegraphics[width=\textwidth]{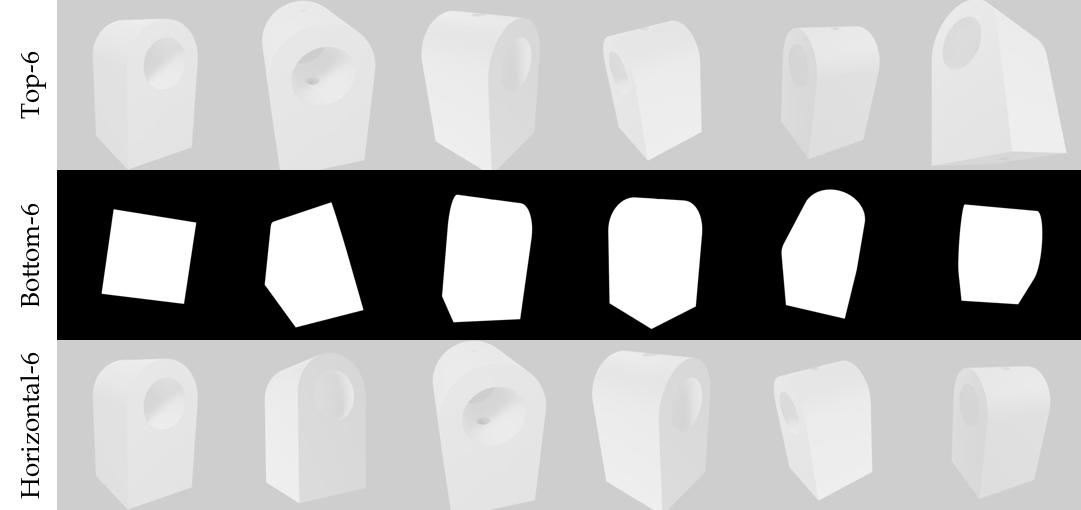}
    \caption{Randomly sampled DiffuRank comparisons. \textbf{Top-row}: Top 6-views selected by DiffuRank; \textbf{Middle-row}: Bottom 6-views selected by DiffuRank; \textbf{Bottom-row}: Horizontal 6-views. }
    \label{fig:appen_diffu_14}
\end{figure}

\begin{figure}
    \centering
    \includegraphics[width=\textwidth]{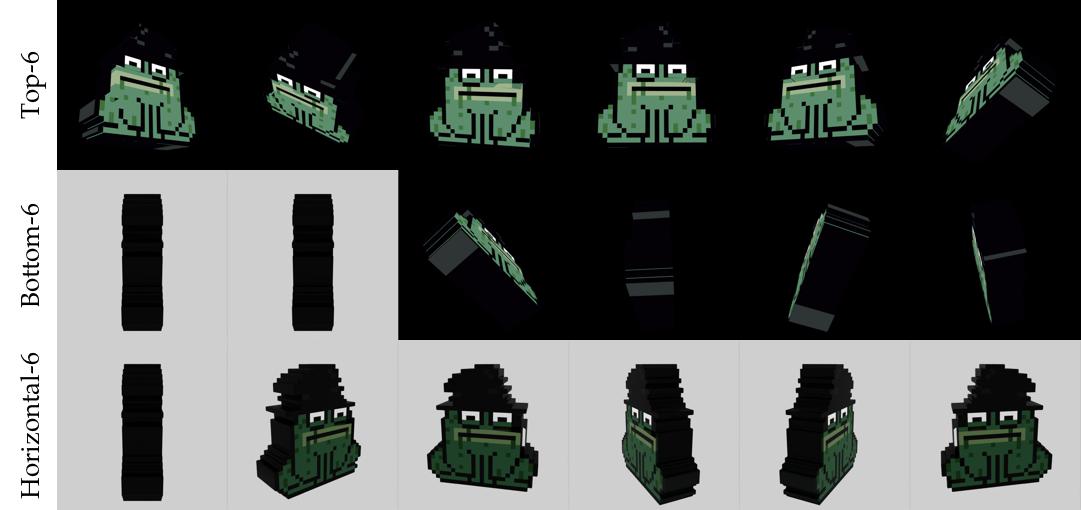}
    \caption{Randomly sampled DiffuRank comparisons. \textbf{Top-row}: Top 6-views selected by DiffuRank; \textbf{Middle-row}: Bottom 6-views selected by DiffuRank; \textbf{Bottom-row}: Horizontal 6-views. }
    \label{fig:appen_diffu_15}
\end{figure}

\begin{figure}
    \centering
    \includegraphics[width=\textwidth]{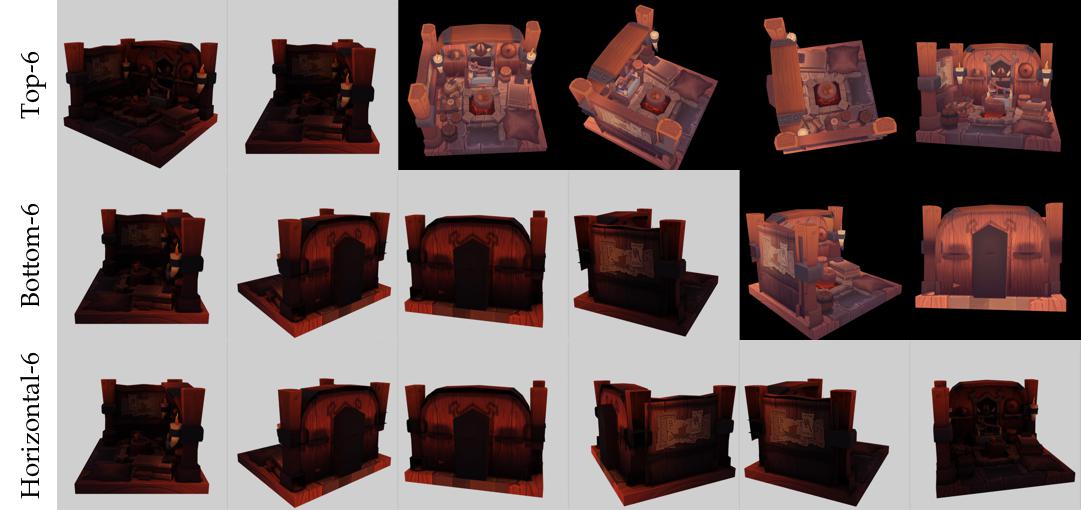}
    \caption{Randomly sampled DiffuRank comparisons. \textbf{Top-row}: Top 6-views selected by DiffuRank; \textbf{Middle-row}: Bottom 6-views selected by DiffuRank; \textbf{Bottom-row}: Horizontal 6-views. }
    \label{fig:appen_diffu_17}
\end{figure}

\begin{figure}
    \centering
    \includegraphics[width=\textwidth]{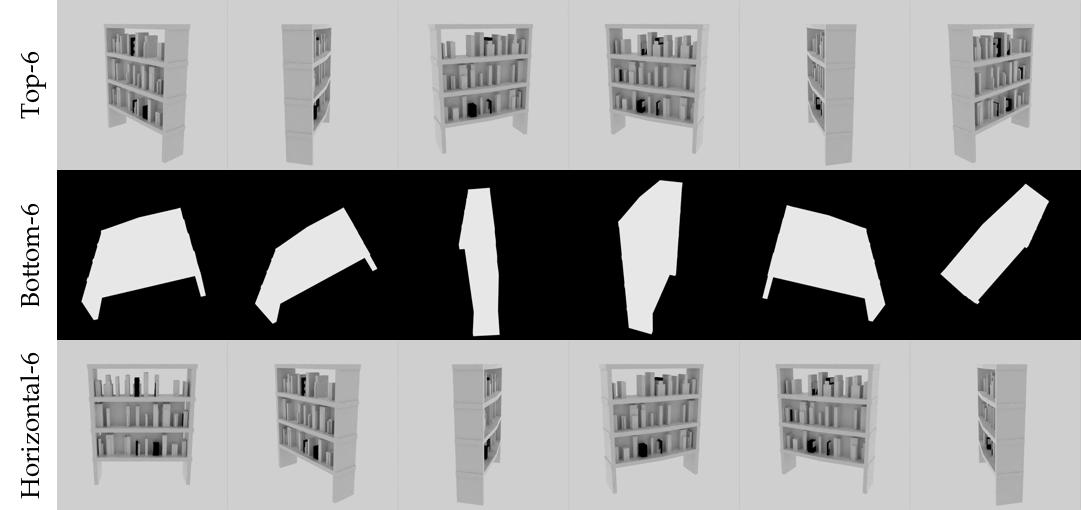}
    \caption{Randomly sampled DiffuRank comparisons. \textbf{Top-row}: Top 6-views selected by DiffuRank; \textbf{Middle-row}: Bottom 6-views selected by DiffuRank; \textbf{Bottom-row}: Horizontal 6-views. }
    \label{fig:appen_diffu_18}
\end{figure}

\begin{figure}
    \centering
    \includegraphics[width=\textwidth]{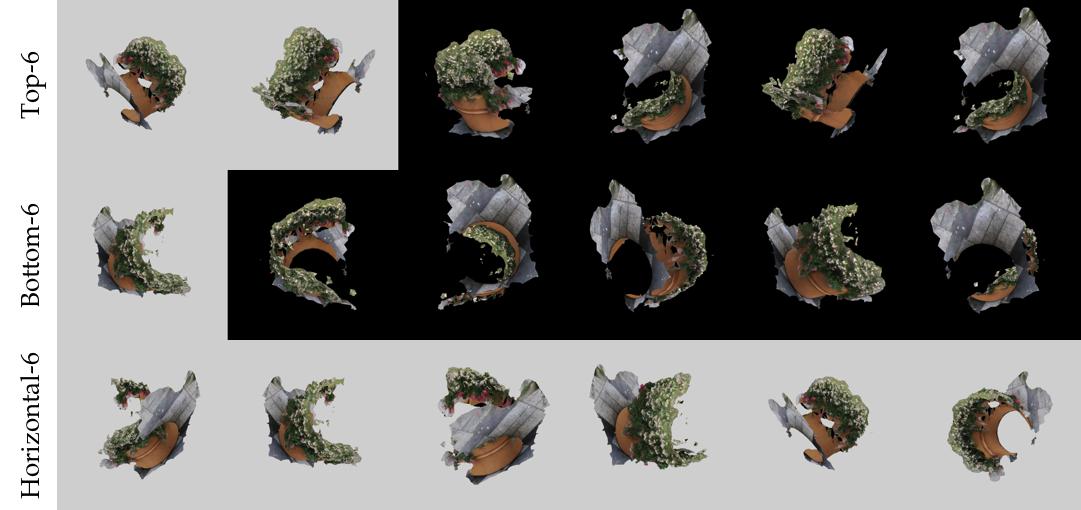}
    \caption{Randomly sampled DiffuRank comparisons. \textbf{Top-row}: Top 6-views selected by DiffuRank; \textbf{Middle-row}: Bottom 6-views selected by DiffuRank; \textbf{Bottom-row}: Horizontal 6-views. }
    \label{fig:appen_diffu_19}
\end{figure}

\clearpage
\hypertarget{B5}{\subsection{DiffuRank: Ours vs. bottom 6-views vs. horizontal 6-views}}
\label{appen:diffu:top_bottom}
This section lists several randomly sampled DiffuRank results of Top 6-views with 6 highest alignment scores (our method), bottom 6-views, and horizontal 6-views. According to the results, we can see (1) Top 6-views obviously outperforms Bottom 6-views on Figures~\ref{fig:appen_diffu_04}, \ref{fig:appen_diffu_05}, \ref{fig:appen_diffu_06}, \ref{fig:appen_diffu_08}, \ref{fig:appen_diffu_13}, \ref{fig:appen_diffu_14}, \ref{fig:appen_diffu_15}, \ref{fig:appen_diffu_18}; (2) Compared to Horizontal 6-views, DiffuRank can adaptly choose angles and types of rendering as shown in Figures~\ref{fig:appen_diffu_02}, \ref{fig:appen_diffu_17}, \ref{fig:appen_diffu_19}; (3) in some cases (Figures~\ref{fig:appen_diffu_01}, \ref{fig:appen_diffu_10}, \ref{fig:appen_diffu_12}), there are no significant difference. 

\begin{figure}[h]
    \centering
    \includegraphics[width=\textwidth]{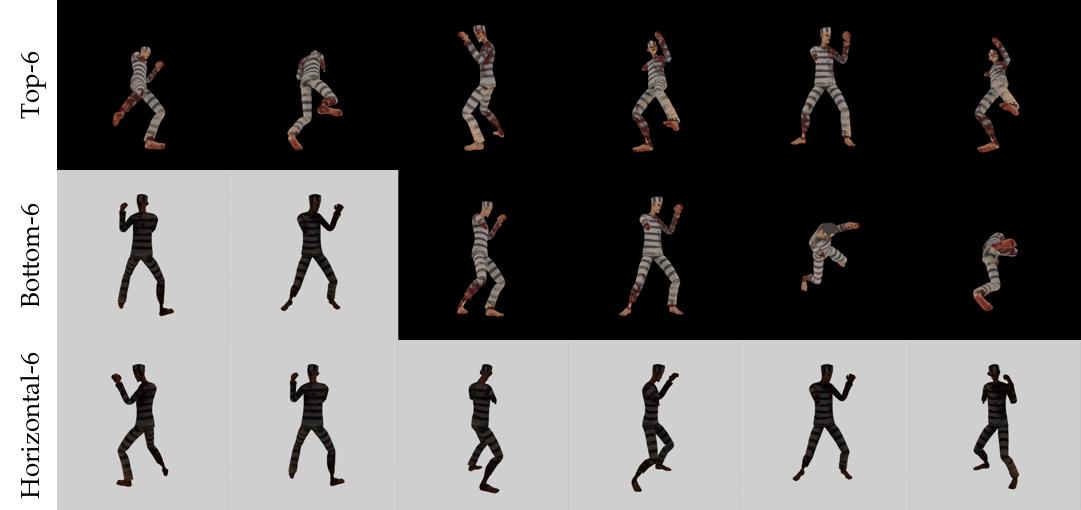}
    \caption{Randomly sampled DiffuRank comparisons. \textbf{Top-row}: Top 6-views selected by DiffuRank; \textbf{Middle-row}: Bottom 6-views selected by DiffuRank; \textbf{Bottom-row}: Horizontal 6-views. }
    \label{fig:appen_diffu_03}
\end{figure}

\begin{figure}
    \centering
    \includegraphics[width=\textwidth]{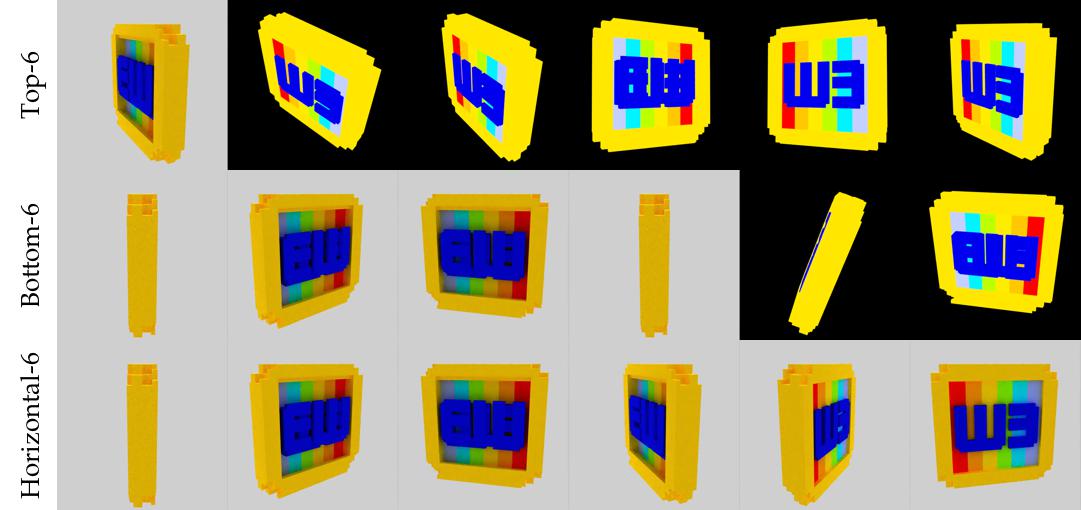}
    \caption{Randomly sampled DiffuRank comparisons. \textbf{Top-row}: Top 6-views selected by DiffuRank; \textbf{Middle-row}: Bottom 6-views selected by DiffuRank; \textbf{Bottom-row}: Horizontal 6-views. }
    \label{fig:appen_diffu_07}
\end{figure}

\begin{figure}
    \centering
    \includegraphics[width=\textwidth]{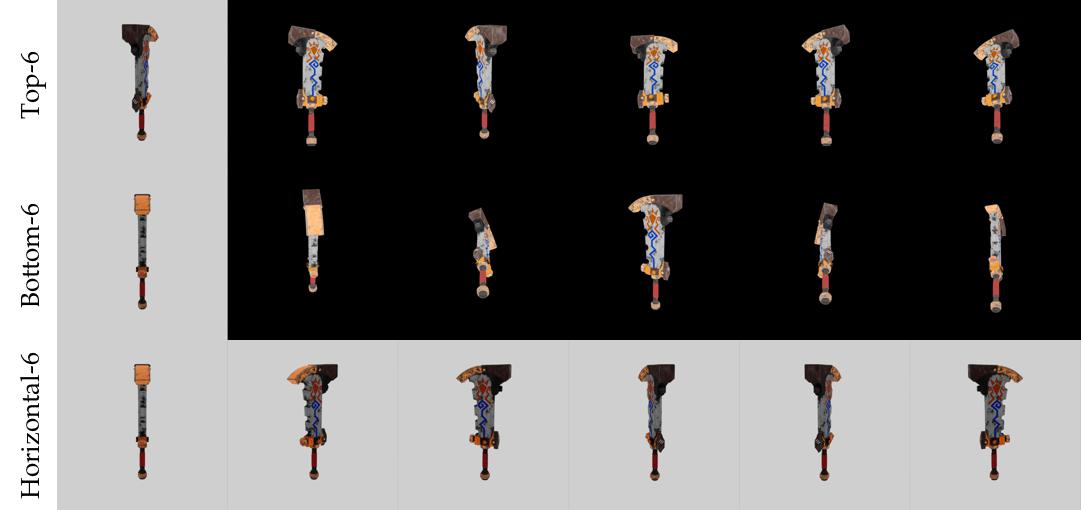}
    \caption{Randomly sampled DiffuRank comparisons. \textbf{Top-row}: Top 6-views selected by DiffuRank; \textbf{Middle-row}: Bottom 6-views selected by DiffuRank; \textbf{Bottom-row}: Horizontal 6-views. }
    \label{fig:appen_diffu_09}
\end{figure}

\begin{figure}
    \centering
    \includegraphics[width=\textwidth]{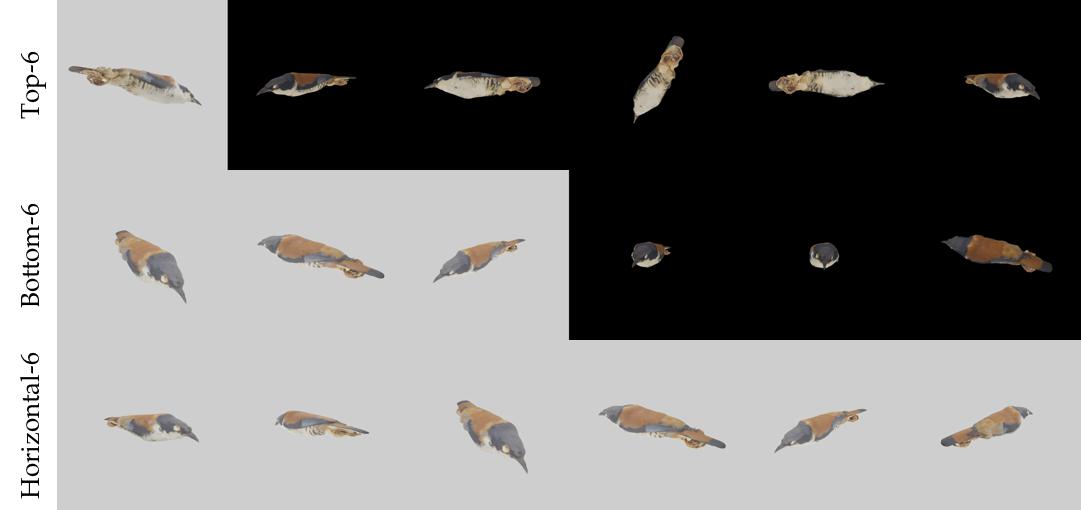}
    \caption{Randomly sampled DiffuRank comparisons. \textbf{Top-row}: Top 6-views selected by DiffuRank; \textbf{Middle-row}: Bottom 6-views selected by DiffuRank; \textbf{Bottom-row}: Horizontal 6-views. }
    \label{fig:appen_diffu_11}
\end{figure}

\begin{figure}
    \centering
    \includegraphics[width=\textwidth]{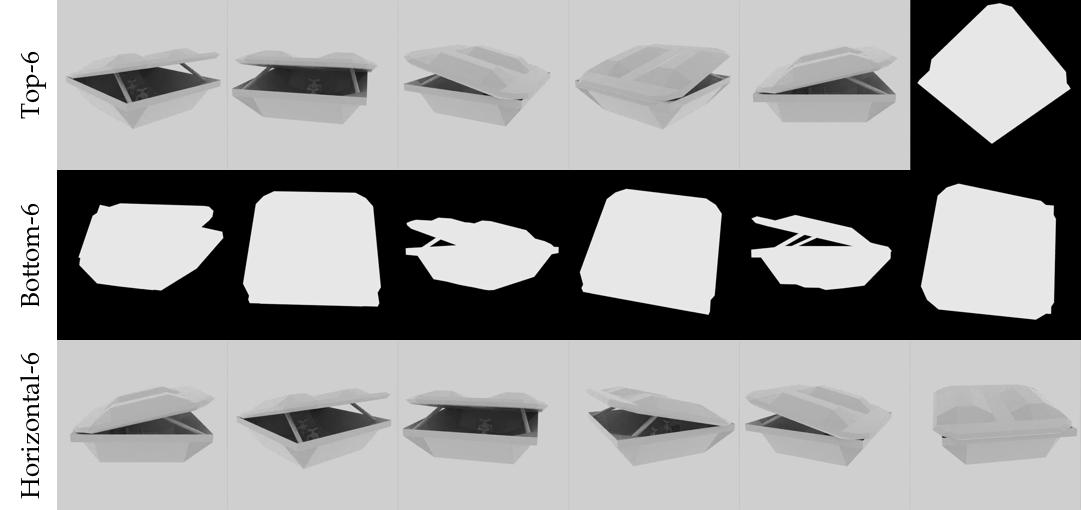}
    \caption{Randomly sampled DiffuRank comparisons. \textbf{Top-row}: Top 6-views selected by DiffuRank; \textbf{Middle-row}: Bottom 6-views selected by DiffuRank; \textbf{Bottom-row}: Horizontal 6-views. }
    \label{fig:appen_diffu_16}
\end{figure}

\begin{figure}
    \centering
    \includegraphics[width=\textwidth]{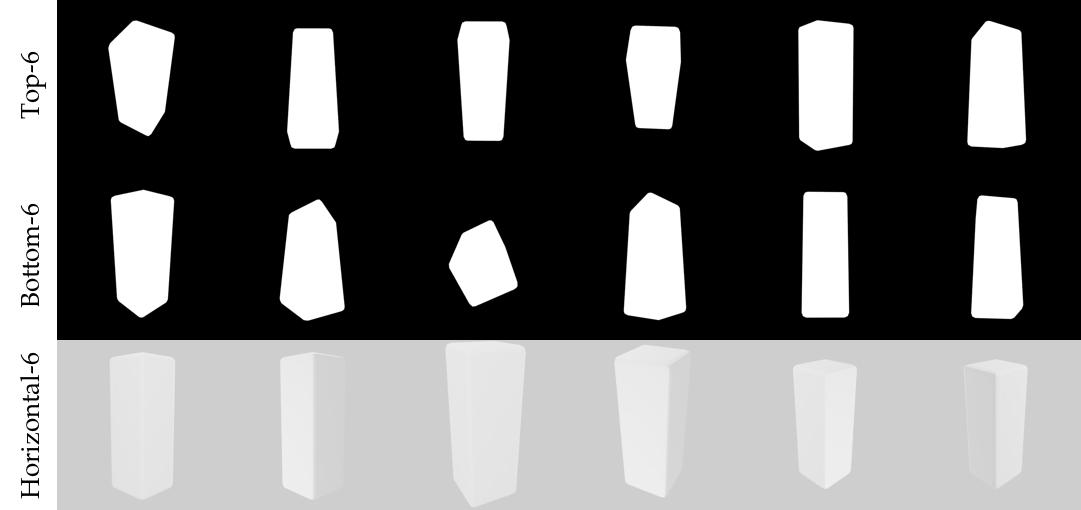}
    \caption{Randomly sampled DiffuRank comparisons. \textbf{Top-row}: Top 6-views selected by DiffuRank; \textbf{Middle-row}: Bottom 6-views selected by DiffuRank; \textbf{Bottom-row}: Horizontal 6-views. }
    \label{fig:appen_diffu_20}
\end{figure}

\begin{figure}
    \centering
    \includegraphics[width=\textwidth]{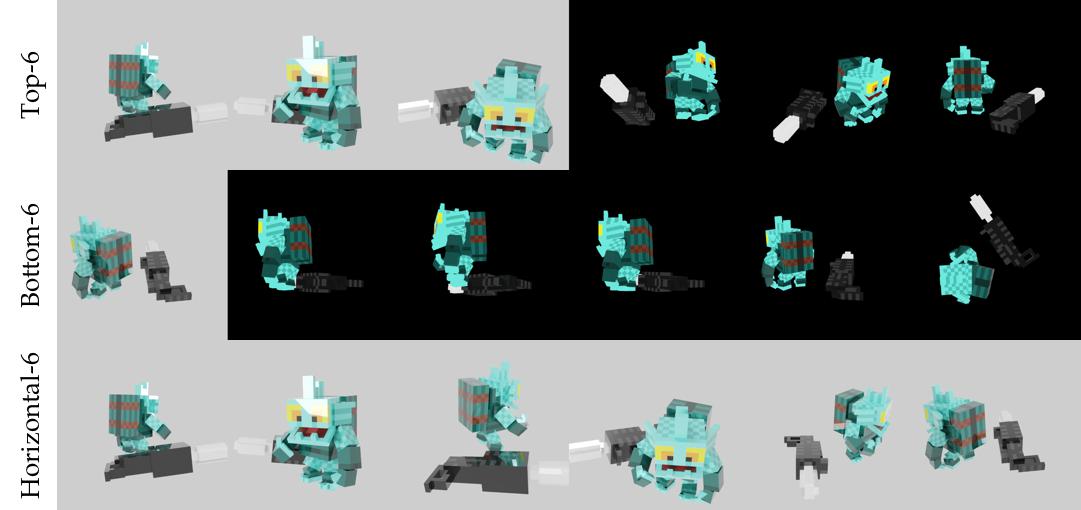}
    \caption{Randomly sampled DiffuRank comparisons. \textbf{Top-row}: Top 6-views selected by DiffuRank; \textbf{Middle-row}: Bottom 6-views selected by DiffuRank; \textbf{Bottom-row}: Horizontal 6-views. }
    \label{fig:appen_diffu_21}
\end{figure}

\begin{figure}
    \centering
    \includegraphics[width=\textwidth]{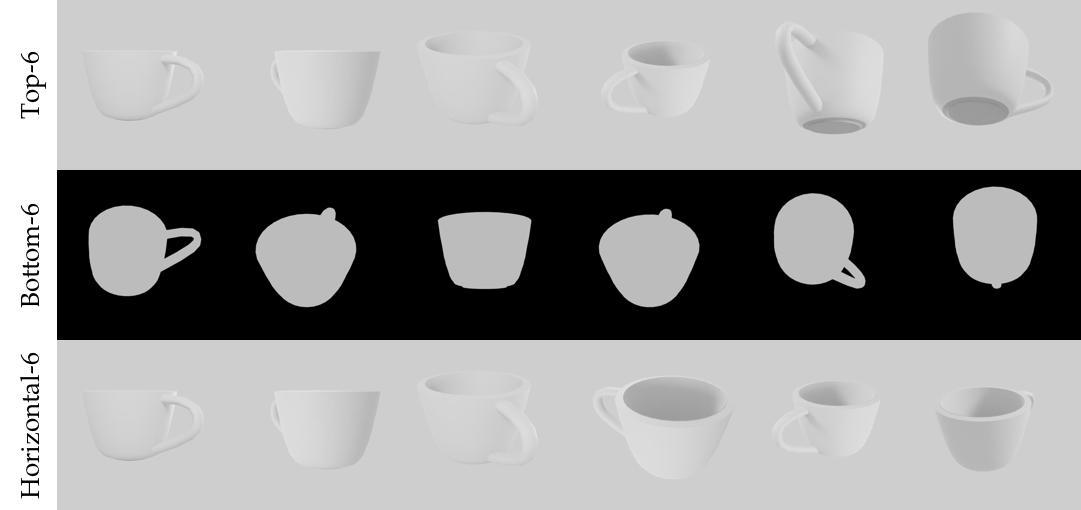}
    \caption{Randomly sampled DiffuRank comparisons. \textbf{Top-row}: Top 6-views selected by DiffuRank; \textbf{Middle-row}: Bottom 6-views selected by DiffuRank; \textbf{Bottom-row}: Horizontal 6-views. }
    \label{fig:appen_diffu_22}
\end{figure}

\begin{figure}[h]
    \centering
    \includegraphics[width=\textwidth]{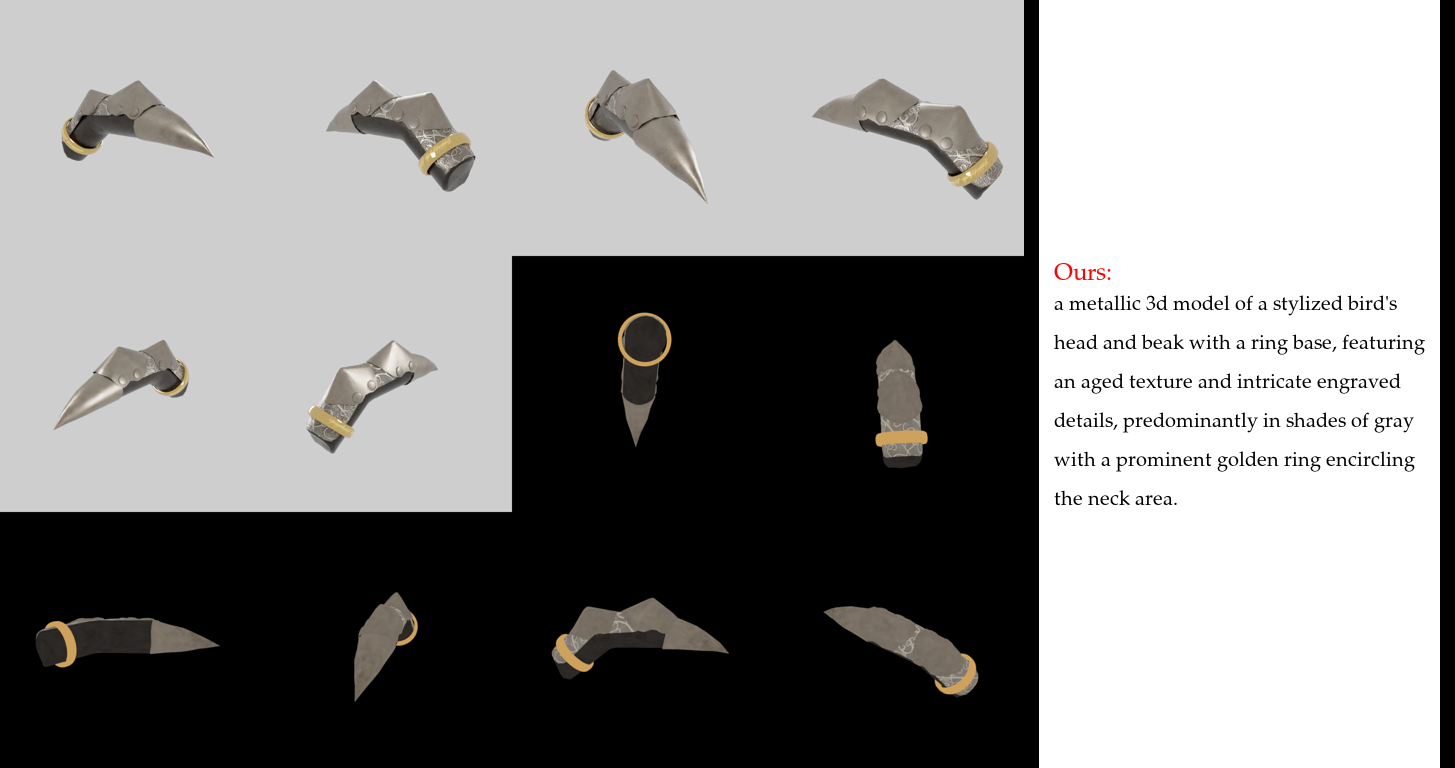}
    \caption{Failure cases: hallucination.}
    \label{fig:failure_3}
    \end{figure}

\begin{figure}[h]
    \centering
    \includegraphics[width=\textwidth]{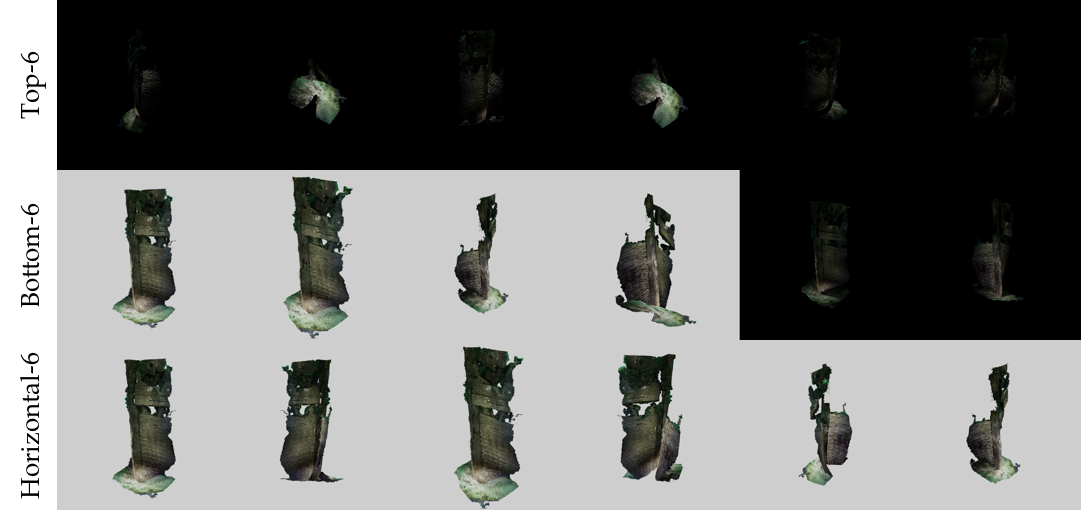}
    \caption{Failure cases: BLIP2 captioning fails or alignment compute not accurate.}
    \label{fig:failure_1}
    \end{figure}

\begin{figure}[h]
    \centering
    \includegraphics[width=\textwidth]{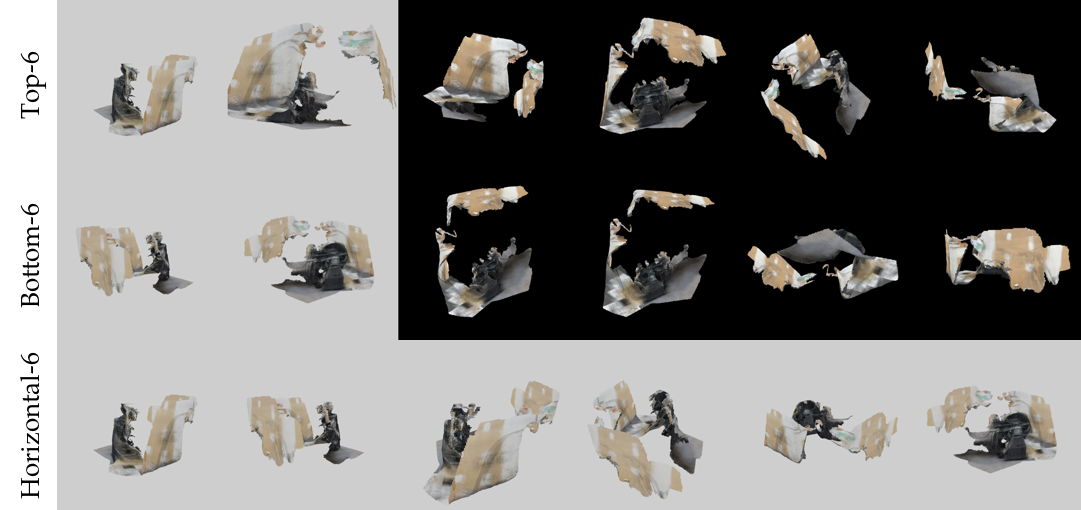}
    \caption{Failure cases: for some scene renderings, our framework fails to capture meaningful characteristics. }
    \label{fig:failure_2}
    \end{figure}

\clearpage
\hypertarget{B6}{\subsection{Failure cases}}
\label{appen:dataset:failure}
We have observed three types of failure cases: (1) DiffuRank fails due to BLIP2 captioning fails or alignment compute not accurate. As shown in Figure~\ref{fig:failure_1}, where BLIP2 captions contain a lot of ``a tree in the dark". Since our DiffuRank needs the initial captioning results to compute alignment scores, BLIP2 captioning fails will cause rendering selection poorly and further cause final caption inaccurate. This could be solved via stronger captioning model, such as GPT4-Vision. Also, as mentioned in Future work (Appendix~\ref{sec:future_limitation}), with better captions, we can fine-tune stronger Text-to-3D models, which help to obtain more accurate alignment scores. (2) sometimes, our captioning method fails to capture small object. One example is in Figure~\ref{fig:cmp_1}, where there is a small black person above the rock, while the caption fails to describe it. Also, it may contain hallucinations with small chances (according to our eyeballs over 10k captions) as shown in Figure~\ref{fig:failure_3}. (3) for some scene renderings, the model failed to capture meaningful characteristics for Figure~\ref{fig:failure_2} with caption ``Abstract 3D composition with fragmented, textured surfaces in shades of beige, white, and charcoal". However, human may also not distinguish this kind of renderings. 

\hypertarget{B7}{\subsection{Human evaluation details}}
\label{appen:human_evaluation}

\begin{figure}[h]
    \centering
    \includegraphics[width=\textwidth]{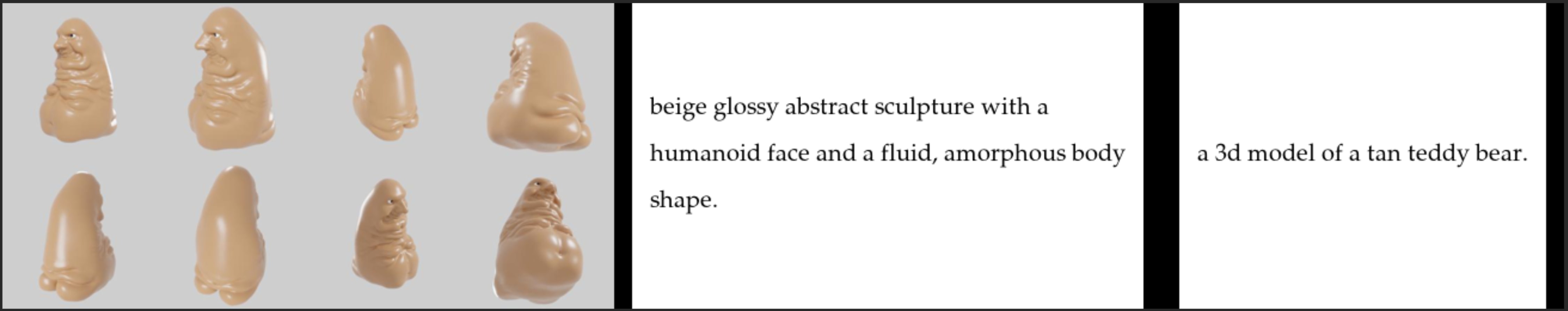}
    \caption{Example hive case. Caption are from ours and Cap3D.}
    \label{fig:hive_1}
\end{figure}

We utilize the \textcolor{purple}{\href{https://thehive.ai/}{Hive}} platform for conducting crowdsourced A/B testing. In this process, participants are presented with an image accompanied by two different captions as shown in Figure~\ref{fig:hive_1}. They are asked to judge which caption is more suitable based on a 5-point scale, where a score of 3 indicates neither caption is preferred over the other. Scores of 1 and 2 suggest a preference for the left caption, with 1 indicating a strong preference and 2 a moderate preference. The sequence in which the captions are presented (left or right) is varied randomly in each case.  

Participants receive guidelines on how to perform this task, including examples that set the standard for quality. We have two distinct types of tasks as shown in Table~\ref{table:caption_evaluation}: quality and hallucination. For quality tasks, workers are advised to focus first on the accuracy of their choices, followed by the level of detail provided in terms of type, structure, and appearance. For hallucination tasks, workers are advised to focus on if the caption contain hallucination or false information. 

We totally hired 46 workers from Hive without access to their personally identifiable information. They are paid approximately \$35 per 1k tasks for our caption evaluation tasks. The entire procedure was carried out in compliance with the ECCV ethics guidelines.

The platform automatically excludes workers who fail to meet the required standards on essential test examples set by us. However, our review revealed that some workers managed to meet the criteria for these essential examples but engaged in deceitful practices for the rest. The prevalent forms of deceit included consistently choosing the same option (always choose left or right) or selecting captions based on their length, either the shortest or the longest. Consequently, we conducted a thorough examination of all workers and excluded those found to be engaging in these deceptive practices, also disregarding their evaluations. 

\hypertarget{C}{\section{Text-to-3D: more details \& results}}
\label{appen:text_to_3d}
In this section, we provide a detailed examination of our Text-to-3D experiments, along with a comprehensive set of qualitative comparisons. It is important to note that employing captions generated by our method typically enhances the performance of Shap·E pre-trained models, a trend that is clearly supported by the data presented in Table~\ref{tab:text_3d}. However, when we fine-tune the Shap·E pre-trained model using Cap3D, we observe a decline in performance across all CLIP-based metrics.

\hypertarget{C1}{\subsection{Setting}}
We adopted the same fine-tune strategy used in Cap3D~\cite{luo2023scalable} for fair comparisons. We employed the AdamW optimizer alongside the CosineAnnealingLR scheduler, setting the initial learning rate at $1e-5$ for fine-tuning both the Point·E and Shap·E models. The batch sizes were set to $64$ for Shap·E and $256$ for Point·E. For training epochs, we set the training epoch which would cost approximately three days. The training was performed on four A40 GPUs. 

The evaluation times, measured in seconds per iteration and inclusive of rendering, are as follows:
\begin{itemize}
    \item For Point·E, the total time is 37 seconds, with 28 seconds dedicated to text-to-3D processing and 9 seconds to rendering.
    \item Shap·E (stf) requires 16 seconds in total for both text-to-3D processing and rendering.
    \item Shap·E (NeRF) takes significantly longer, with a total of 193 seconds for both text-to-3D processing and rendering.
\end{itemize}

\hypertarget{C2}{\subsection{Qualitative comparisons}}
\begin{figure}[h]
    \centering
    \includegraphics[width=0.7\textwidth]{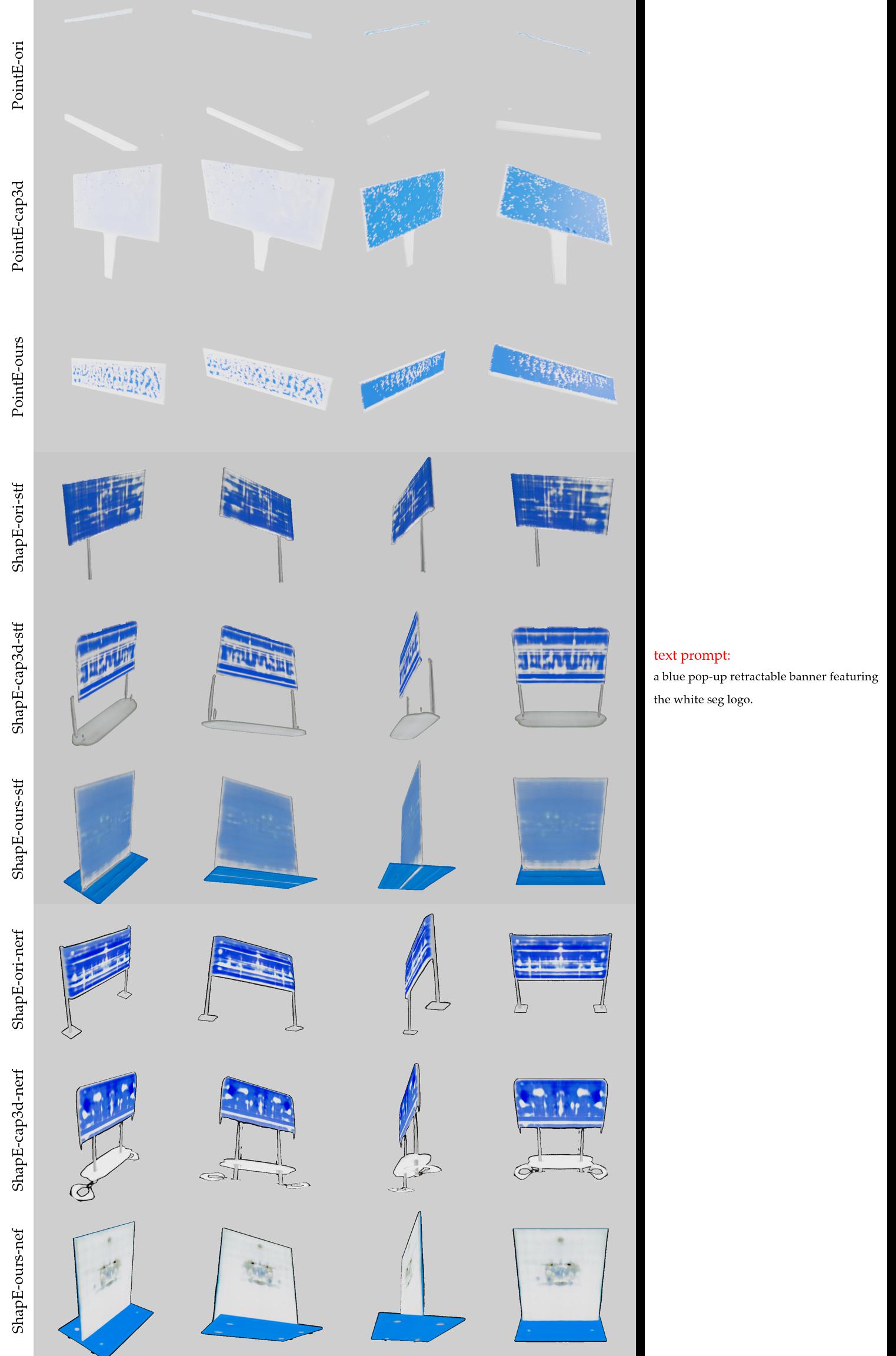}
    \caption{Randomly sampled Text-to-3D results.}
    \label{fig:appen_text23d_01}
\end{figure}

\begin{figure}[h]
    \centering
    \includegraphics[width=\textwidth]{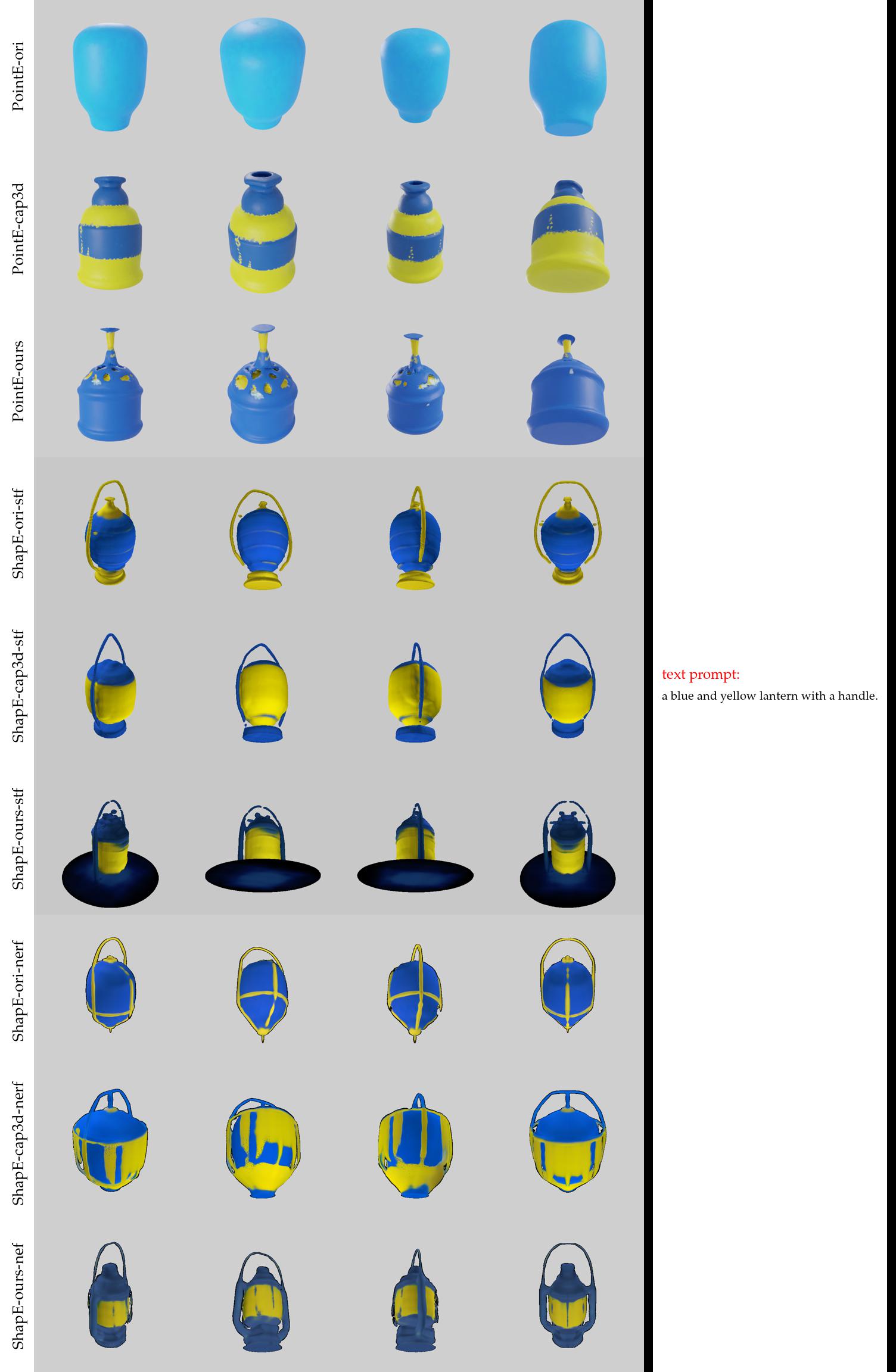}
    \caption{Randomly sampled Text-to-3D results.}
    \label{fig:appen_text23d_02}
\end{figure}

\begin{figure}
    \centering
    \includegraphics[width=\textwidth]{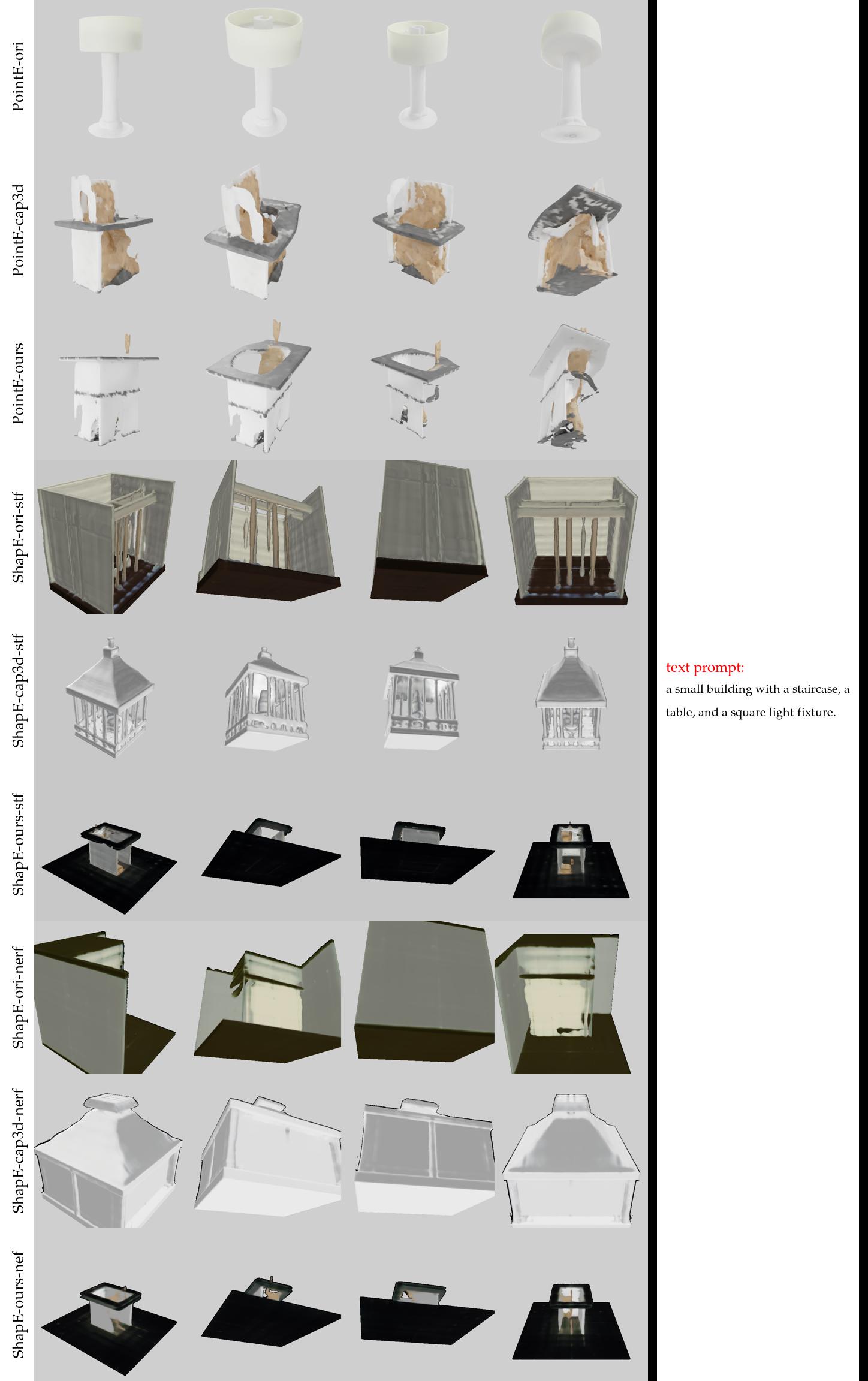}
    \caption{Randomly sampled Text-to-3D results.}
    \label{fig:appen_text23d_03}
\end{figure}

\begin{figure}
    \centering
    \includegraphics[width=\textwidth]{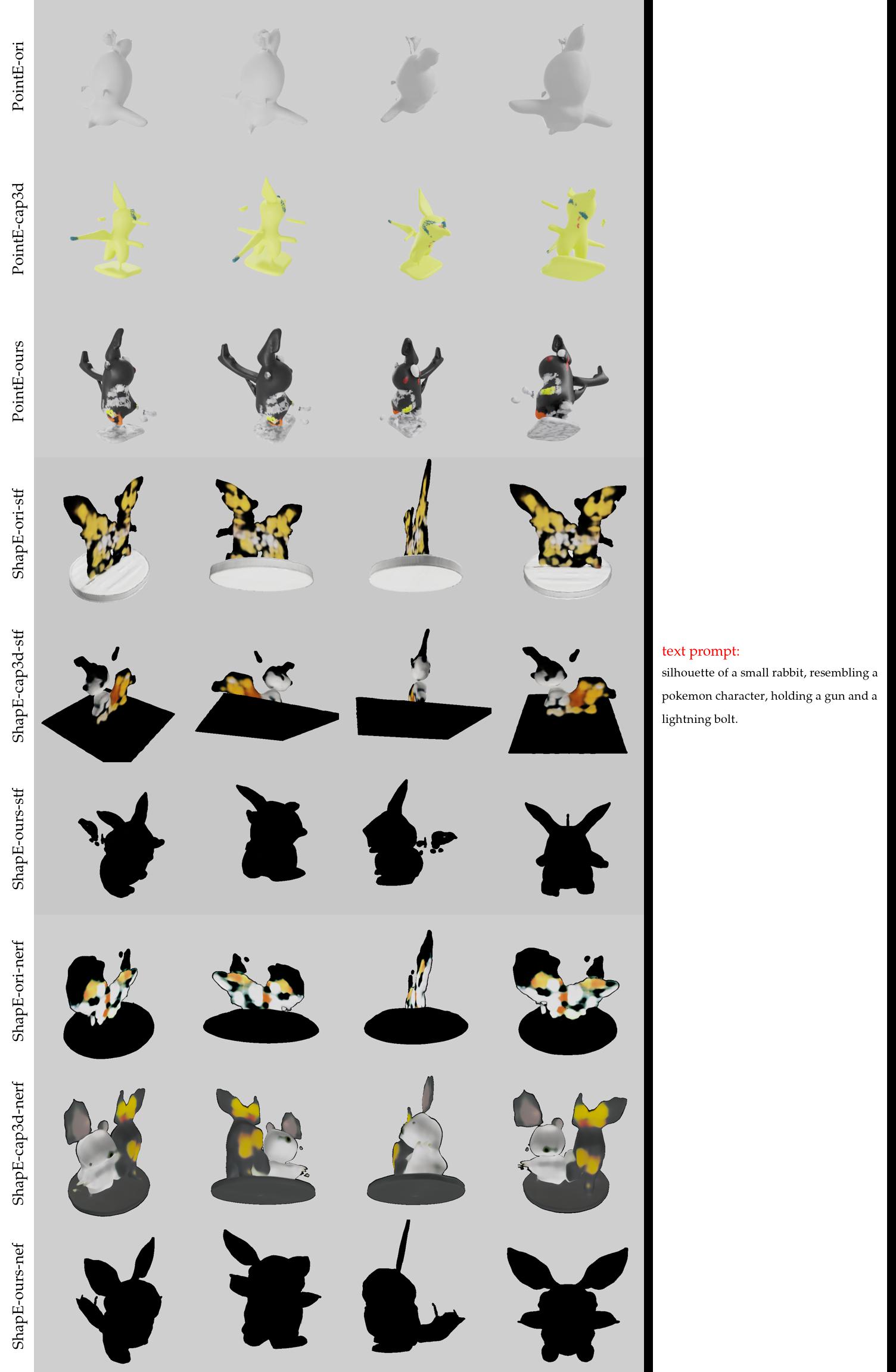}
    \caption{Randomly sampled Text-to-3D results.}
    \label{fig:appen_text23d_04}
\end{figure}

\begin{figure}
    \centering
    \includegraphics[width=\textwidth]{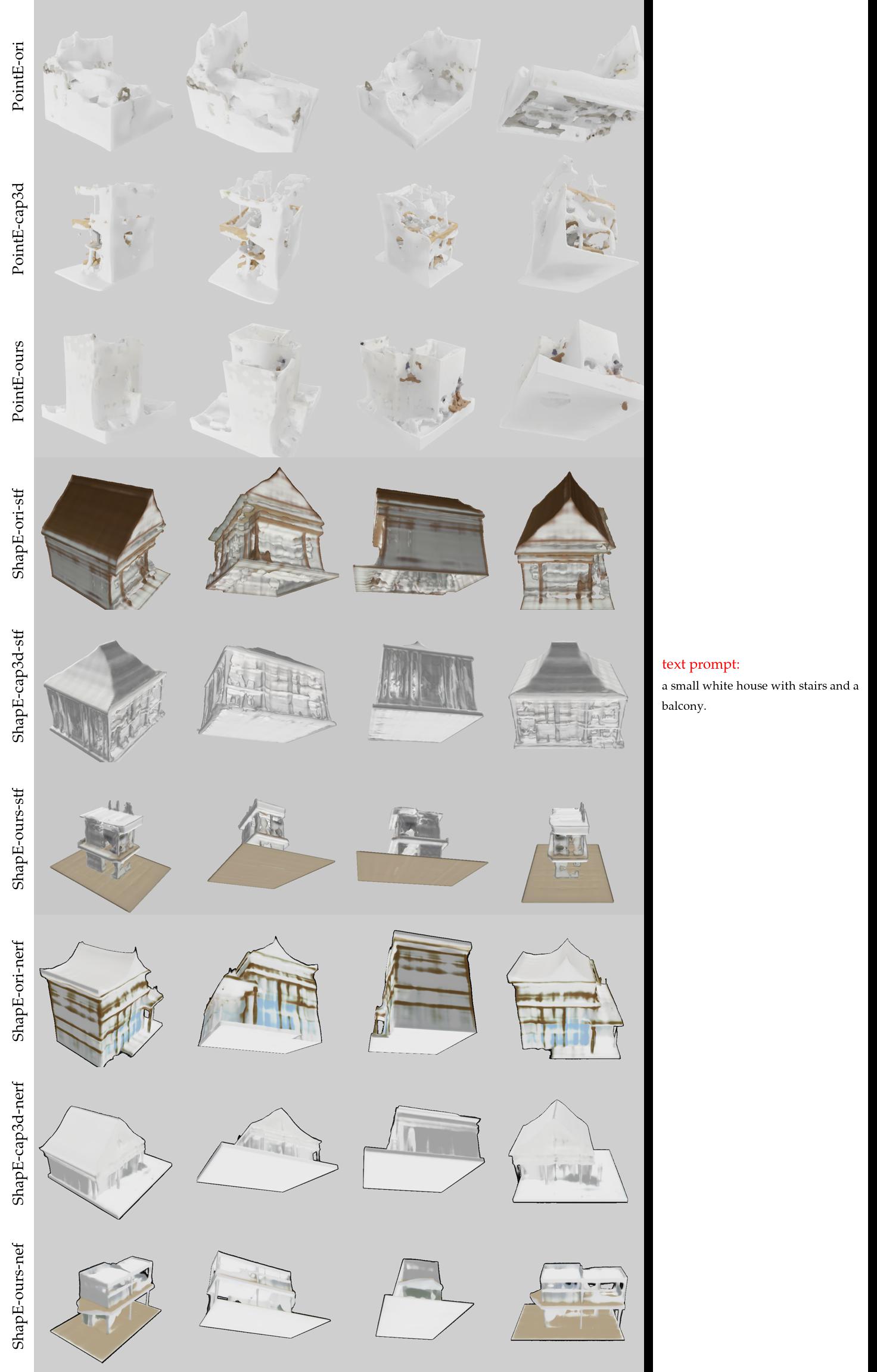}
    \caption{Randomly sampled Text-to-3D results.}
    \label{fig:appen_text23d_05}
\end{figure}

\begin{figure}
    \centering
    \includegraphics[width=\textwidth]{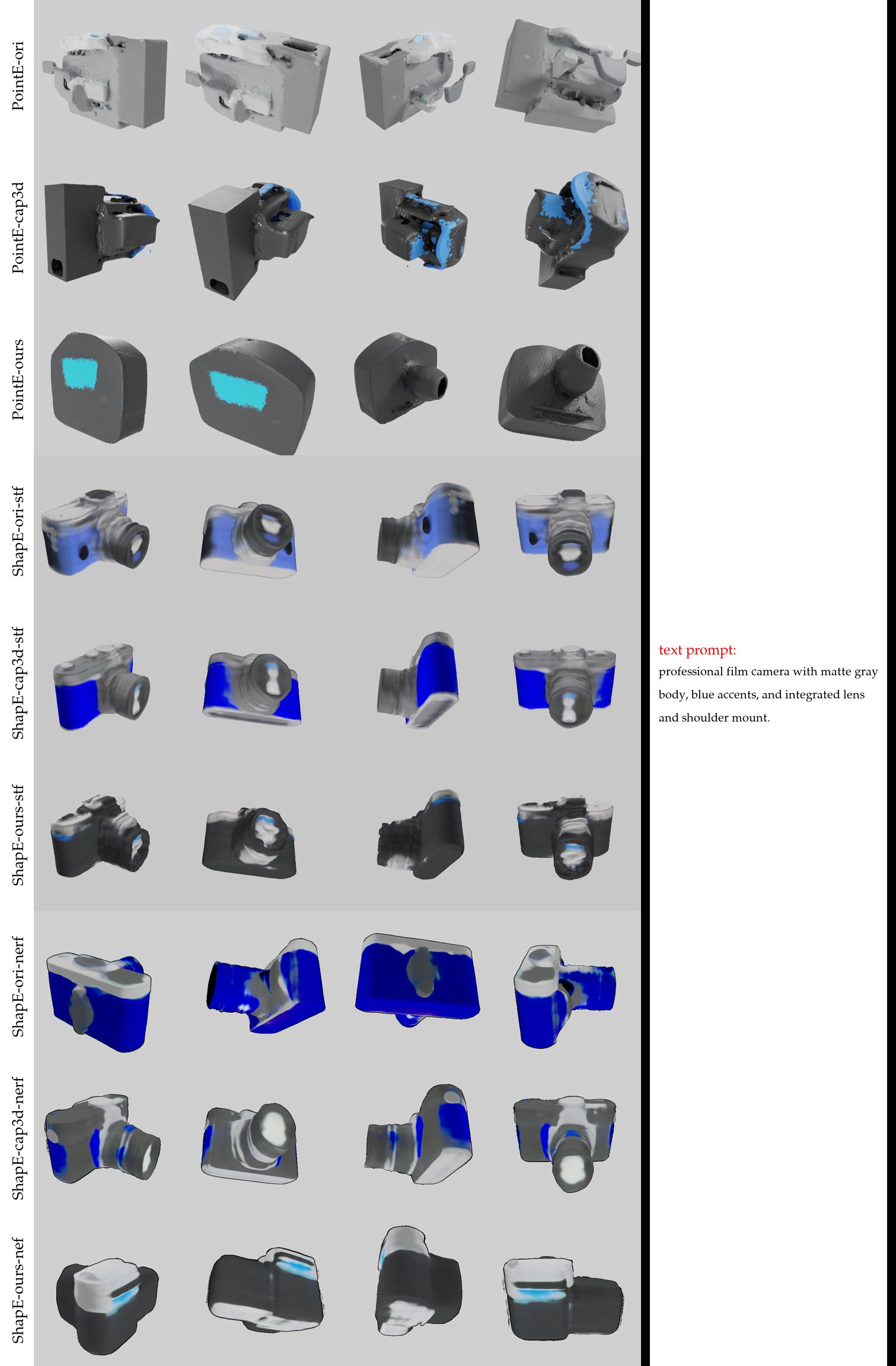}
    \caption{Randomly sampled Text-to-3D results.}
    \label{fig:appen_text23d_06}
\end{figure}

\clearpage
\hypertarget{D}{\section{DiffuRank on VQA}}
\label{appen:vqa}
Algorithm~\ref{alg:diffurank_vqa} demonstrates the DiffuRank approach to the task of 2D Visual Question Answering. Initially, the process involves converting the question and each potential answer/option into a coherent statement. As shown in Figure~\ref{fig:vqa}, we convert Question: ``Is the school bus driving towards or away from the camera?"  and options ``(a) Towards the camera (b) Away from the camera" into statements (1) ``The school bus is driving towards the camera and statement" and (2) ``The school bus is driving away from the camera". Another example shows converting Question: ``Is there a shadow on the flower?" and options ``(a) Yes (b) No,(a)" into statements (1) ``There is a shadow on the flower." and (2) ``There is not a shadow
on the flower."

This conversion is accomplished through the utilization of GPT-4 in our implementation. Subsequently, we determine the alignment scores by evaluating the correspondence between each generated statement and the provided 2D image. The statement that exhibits the highest alignment score, along with its associated option, is then selected as the definitive answer. 

Different from Algorithm~\ref{alg:diffurank}, our objective here is computed over noise difference, the way adopted in our used stable-diffusion models~\cite{rombach2022high}.

\begin{algorithm}[h]
\caption{DiffuRank for modeling the alignments between 2D images and answers for VQA tasks}
\label{alg:diffurank_vqa}
\begin{algorithmic}
\REQUIRE Given a Visual Question Answering (VQA) task, which consists of images $\mathcal{O}$, a question $q$, and multiple options ${o}_{i}$, and a pre-trained text-to-2D model $D{\text{text-to-2D}}$

\STATE \# 1. Turn question $q$ and multiple options $\{o\}_{i=1,\cdots,M}$ into multiple corresponding statements $\{s\}_{i=1,\cdots,M}$;

\STATE \# 2. Compute average alignment scores
\FOR{each statement $s_i$}
    \FOR{$k \gets 1$ to $\text{num\_samples}$}
        \STATE Sample timestamp $t_k \sim \text{Uniform}(0, 1)$.
        \STATE Sample noise $\epsilon_k \sim \mathcal{N}(0, I)$.
        \STATE Compute noised input $\mathcal{O}_{t_k} = \sqrt{\bar{\alpha}_{t_k}} \mathcal{O}_0 + \sqrt{1-\bar{\alpha}_{t_k}} \epsilon_k$.
        \STATE Compute loss $\mathcal{L}_{s_i,k}=\|D_{\text{text-to-3D}}(\mathcal{O}_{t_k} |s_i)  - \epsilon_k$.
    \ENDFOR
    \STATE Compute average loss for each statement $s_i$, $Cor({s_i},\mathcal{O}) = -\mathbb{E}_{k} \mathcal{L}_{s_i, k}$.
\ENDFOR
\RETURN Top-1($\{Cor(s_i, \mathcal{O})\}_{i=1,\cdots, M}$)
\end{algorithmic}
\end{algorithm}

\clearpage
\hypertarget{E}{\section{Future Work \& Limitations}}
\label{sec:future_limitation}

\textbf{Future Work:} DiffuRank leverages a pre-trained text-to-3D diffusion model for rendering view ranking, enhancing 3D object captioning. Improved captioning enables the refinement of the diffusion model, creating a feedback loop that cyclically utilizes the model for data generation and employs this data to fortify the model further. Besides, due to our limited computational resources and funding, it is not feasible to encompass all Objaverse-XL objects, presenting an opportunity for industrial entities.

\textbf{Limitations:} During our subtitling process, we use DiffuRank to select 6 rendered views out of 28 views. This process requires us to render more views, generate captions, and perform inference using a pre-trained text-to-3D diffusion model to compute alignment scores. All of the steps take calculation and time.

As highlighted in the related work (Section~\ref{sec:related_work}), DiffuRank faces challenges with speed, requiring multiple samplings for each option and necessitating forward model processing for all options. Our process for a single 3D object involves 28 rendered views, 5 captions per view, and performing sampling 5 times ($num_{sample}$ in Alg.~\ref{alg:diffurank}), resulting in a total of 700 inference operations. While parallel processing (large batch size) can mitigate delays, the system's performance is inherently slow. We show a VQA extension in Section~\ref{sec:exp:diffurank} as it only has two options. But, generally, DiffuRank's design is not optimal for tasks requiring numerous options, such as classification and image-text retrieval. 

Our discussion around broader impact is listed in Appendix~\ref{appen:broader_impact}. Some of the failure cases and analysis are included in Appendix~\ref{appen:dataset:failure}.
\end{document}